\documentclass[11pt, a4paper, logo, onecolumn]{deepmind}

\pdftrailerid{redacted}

\makeatletter
\renewcommand\bibentry[1]{\nocite{#1}{\frenchspacing\@nameuse{BR@r@#1\@extra@b@citeb}}}
\makeatother

\usepackage{todonotes}

\usepackage{kantlipsum, lipsum}
\usepackage{dsfont}
\usepackage{float}
\usepackage{dm-colors}
\usepackage{amsmath}
\usepackage{subcaption}
\usepackage{siunitx}
\usepackage{gensymb}
\usepackage{chngcntr}
\usepackage{multicol}
\usepackage{epigraph} 

\newcommand*{\defeq}{\mathrel{\vcenter{\baselineskip0.5ex \lineskiplimit0pt
                     \hbox{\scriptsize.}\hbox{\scriptsize.}}}
                     =}

\usepackage[authoryear, sort&compress, round]{natbib} 
\graphicspath{{figures/}}
\title{Learning Robust Real-Time Cultural Transmission without Human Data}
\correspondingauthor{Edward Hughes (\href{mailto:edwardhughes@deepmind.com}{edwardhughes@deepmind.com}), Richard Everett (\href{mailto:reverett@deepmind.com}{reverett@deepmind.com}).}

\reportnumber{} 
\renewcommand{\today}{} 

\usepackage{caption}
\author[*]{Cultural General Intelligence Team}

\affil[*]{See Section \ref{authors} for Authors \& Contributions}

\begin{abstract}

Cultural transmission is the domain-general social skill that allows agents to acquire and use information from each other in real-time with high fidelity and recall. In humans, it is the inheritance process that powers cumulative cultural evolution, expanding our skills, tools and knowledge across generations. We provide a method for generating zero-shot, high recall cultural transmission in artificially intelligent agents. Our agents succeed at real-time cultural transmission from humans in novel contexts without using any pre-collected human data.  We identify a surprisingly simple set of ingredients sufficient for generating cultural transmission and develop an evaluation methodology for rigorously assessing it. This paves the way for cultural evolution as an algorithm for developing artificial general intelligence. 

\end{abstract}

\begin{document}

\begin{figure}[ht]
    \centering
    
    \begin{subfigure}{0.24\textwidth}
        \includegraphics[width=\textwidth]{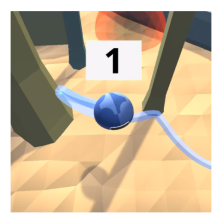}
        \caption{\centering}
        \label{fig:goal-cycle-overview-a}
    \end{subfigure}
    \hfill
    \begin{subfigure}{0.24\textwidth}
        \includegraphics[width=\textwidth]{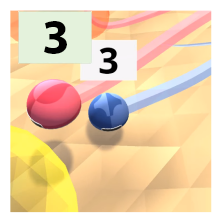}
        \caption{\centering}
        \label{fig:goal-cycle-overview-b}
    \end{subfigure}
    \hfill
    \begin{subfigure}{0.24\textwidth}
        \includegraphics[width=\textwidth]{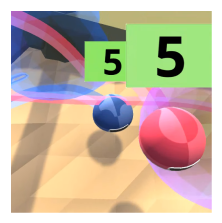}
        \caption{\centering}
        \label{fig:goal-cycle-overview-c}
    \end{subfigure}
    \hfill
    \begin{subfigure}{0.24\textwidth}
        \includegraphics[width=\textwidth]{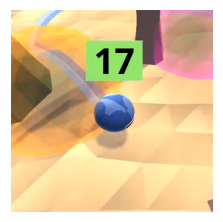}
        \caption{\centering}
        \label{fig:goal-cycle-overview-d}
    \end{subfigure}
    
    \caption{Freeze-frames from a single episode of test-time evaluation, in chronological order from left to right. (a) Our cultural transmission agent (blue avatar) is spawned in a held-out task; (b) the agent finds a human (red avatar); (c) the agent follows the human on a rewarding path through goals while navigating terrain and obstacles; (d) the agent recalls and reproduces the demonstrated path after the human has dropped out. The numbers above the avatars indicate cumulative score across the episode. Videos of this agent are available on the \href{https://sites.google.com/view/dm-cgi}{website accompanying this paper}.}
    \label{fig:goal-cycle-overview}
\end{figure}

\maketitle

\newcommand{\expect}[2]{\mathds{E}_{{#1}} \left[ {#2} \right]}
\newcommand{\myvec}[1]{\boldsymbol{#1}}
\newcommand{\myvecsym}[1]{\boldsymbol{#1}}
\newcommand{\vx}{\myvec{x}}
\newcommand{\vy}{\myvec{y}}
\newcommand{\vz}{\myvec{z}}
\newcommand{\vtheta}{\myvecsym{\theta}}

\leavevmode\newpage
\section{Introduction}
\label{sec:introduction}

Intelligence can be defined as the ability to acquire new knowledge, skills, and behaviours efficiently across a wide range of contexts \citep{chollet2019measure}. Human intelligence is especially dependent on our ability to acquire knowledge efficiently from other humans. This knowledge is collectively known as culture, and the transfer of knowledge from one individual to another is known as \emph{cultural transmission}. Cultural transmission is a form of social learning, learning assisted by contact with other agents. It is specialised for the inheritance of culture \citep{heyes2018cognitive} via high fidelity, consistent recall, and generalisation to previously unseen situations. We refer to these properties as \textit{robustness} \citep{marcus2020next}. Robust cultural transmission is ubiquitous in everyday human social interaction, particularly in novel contexts: copying a new recipe as seen on TV, following the leader on a guided tour, showing a colleague how the printer works, and so on. 

We seek to generate an artificially intelligent agent capable of robust real-time cultural transmission from human co-players in a rich 3D physical simulation. The motivations for this are threefold. First, since cultural transmission is an ever-present feature of human social behaviour, it is a skill that an artificially intelligent agent should possess to facilitate beneficial human-AI interaction. Second, cultural transmission is the inheritance process that underlies the (Darwinian) evolution of culture \citep{blackmore2000meme},\footnote{Cultural evolution, like all evolution by natural selection, is the inevitable consequence of three processes: variation, selection and inheritance.} arguably the fastest known intelligence-generating process \citep{henrich2015secret}. A convincing demonstration of cultural transmission would pave the way for cultural evolution as an AI-generating algorithm \citep{clune2019ai, leibo2019autocurricula}. Third, rich 3D physical simulations with sparse reward pose hard exploration problems for artificial agents, yet human behaviour in this setting is often highly informative and cheap in small quantities. Cultural transmission provides an efficient means of structured exploration in behaviour space.

We focus on a particular form of cultural transmission, known in the psychology and neuroscience literature as observational learning \citep{bandura1977social} or imitation \citep{heyes2016homo}. In this field, imitation is defined to be the copying of body movement. It is a particularly impressive form of cultural transmission because it requires solving the challenging ``correspondence problem'' \citep{nehaniv2002correspondence}, instantaneously translating a sensory observation of another agent's motor behaviour into a motor reproduction of that behaviour oneself. In humans, imitation provides the kind of high-fidelity cultural transmission needed for the cultural evolution of tools and technology \citep{de1903laws, meltzoff1988imitation}. Borrowing the language of cognitive science, we want to know how the skill of cultural transmission can develop during artificial agent ontogeny, or more colloquially, childhood \citep{tomasello2016ontogeny}.

Our artificial agent is parameterised by a neural network and we use deep reinforcement learning (RL) to train the weights. The resulting neural network (with fixed weights) is capable of zero-shot, high-recall cultural transmission within a ``test'' episode on a wide range of previously unseen tasks. Our approach contrasts with prior methods in which the training itself is a process of cultural transmission, namely imitation learning \citep{argall2009survey, hussein2017imitation, edwards2019imitating, torabi2019recent} and policy distillation \citep{hinton2015distilling, rusu2016policy}. These methods are highly effective on individual tasks, but they are not few-shot learners, require privileged access to datasets of human demonstrations or a target policy, and struggle to generalise to held-out tasks \citep{ren2020generalization}. Closer to our setting, \cite{borsa2017observational,woodward2020learning,ndousse2020learning} use RL to generate test-time social learning, but these agents lack the ability of within-episode recall and only generalise across a limited task space. 

\clearpage
The central novelty in this work is the application of agent-environment co-adaptation \citep{wang2019paired, oelt2021open} to generate an agent capable of robust real-time cultural transmission. To this end, we introduce a new open-ended reinforcement learning environment, \textit{GoalCycle3D}. This environment offers a diverse task space by virtue of procedural generation, 3D rigid-body physics, and continuous first-person sensorimotor control. Here, we explore more challenging exploration problems and generalisation challenges than in previous literature. We introduce a rigorous \textit{cultural transmission metric} and transplant a two-option paradigm from cognitive science \citep{Dawson1965ObservationalLI,whiten1998imitation, aplin2015experimentally} to make causal inference about information transfer from one individual to another. This puts us on a firm footing from which to establish state-of-the-art generalisation and recall capabilities.

Via careful ablations, we identify a minimal sufficient ``starter kit'' of training ingredients required for cultural transmission to emerge, namely function approximation, memory (M), the presence of an expert co-player (E), expert dropout (D), attentional bias towards the expert (AL), and automatic domain randomisation (ADR). We refer to this collection by the acronym MEDAL-ADR. Individually, these components aren't complex, but together they generate a powerful agent. We analyse the capabilities and limitations of our agent's cultural transmission abilities on three axes inspired by the cognitive science of imitation, namely recall, generalisation, and fidelity. We find that cultural transmission generalises outside the training distribution, and that agents recall demonstrations within a single episode long after the expert has departed. Introspecting the agent's "brain", we find strikingly interpretable neurons responsible for encoding social information and goal states. 

\clearpage
\section{Task space}
\label{sec:environment}

To train and evaluate our agents, we introduce a 3D physical simulated task space called \textit{GoalCycle3D}, inspired by the GoalCycle environment of \cite{ndousse2020learning}. This task space allows us to explore cultural transmission of navigational skills: a natural starting point given the importance of navigation in human and animal culture \citep{bond2020wayfinding, jesmer2018ungulate, palacin2011cultural, mueller2013social} and the prior focus on navigation problems in AI \citep{mirowski2016learning, mirowski2018learning, banino2018vector, alonso2020deep}. 

Each task contains procedurally-generated terrain, obstacles, and goal spheres, with parameters randomly sampled on task creation. Each agent is independently rewarded for visiting goals in a particular cyclic order, also randomly sampled on task creation. The correct order is not provided to the agent, so an agent must deduce the rewarding order either by experimentation or via cultural transmission from an expert. Our task space presents navigational challenges of open-ended complexity, parameterised by world size, obstacle density, terrain bumpiness and number of goals. 

\begin{figure}[t]
\centering
\includegraphics[width=1.0\linewidth]{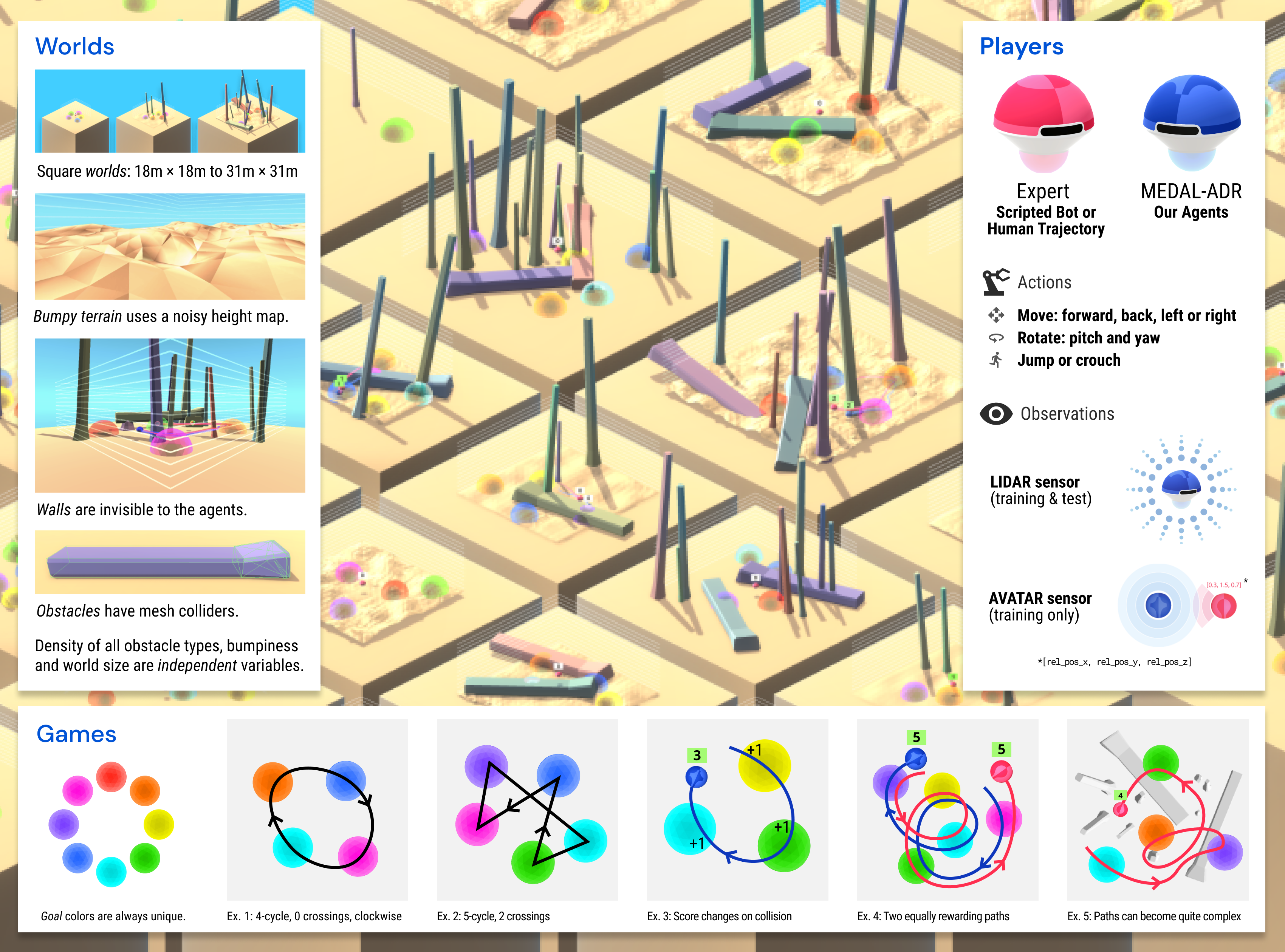}
\caption{\small The GoalCycle3D task space at a glance.}
\label{fig:env-at-a-glance}
\end{figure}

Similar to \cite{oelt2021open}, we decompose an agent's task as the direct product of a world, a game and a set of co-players. The world comprises the size and shape of the terrain and locations of objects. The game defines the correct order(s) of goals and the reward dynamics for each player. Co-players are other interactive policies in the world, consuming observations and producing actions. In the remainder of this section, we provide an overview of the world space, game space and player interface, summarised in Figure \ref{fig:env-at-a-glance}. For detailed information, see Appendix \ref{app:task_space}. 
\subsection{World space}\label{sec:world}

Each world is built using the Unity game engine \citep{juliani2018unity, ward2020using}. It is a three-dimensional simulated physical space in which avatars are situated. An avatar is an embodiment of a player in the virtual world and is capable of perception and movement. Each world is perfectly square, of size between $16 \times 16 \SI{}{\meter\squared}$ and $37 \times 37\SI{}{\meter\squared}$. The bumpy terrain is procedurally generated using Libnoise \citep{bevinslibnoise}, parameterised by frequency and amplitude. The playable terrain is bounded by an impermeable barrier, invisible to players but visible to human observers. 

Within the playable terrain are procedurally-generated horizontal and vertical obstacles, parameterised by density. The obstacles create navigational and perception challenges for players. In empty worlds, the path between two goals is often immediately visible, and typically a straight line. In worlds with obstacles, visibility can be highly restricted, since obstacles block vision and the LIDAR sensors used by our agents (see Section \ref{sec:player-interface}). Moreover, players may need to take complex paths to reach a goal, including jumping or crouching, requiring continued actions for many steps. Figures \ref{fig:worlds-same-seed} and \ref{fig:worlds-three-seeds} illustrate some representative worlds using different parameters and seeds.

\subsection{Game space}\label{sec:game}

Players are positively rewarded for visiting \textit{goal} spheres in particular cyclic orders. In each possible order, every goal appears exactly once. Therefore, the set of distinct orders may be conveniently expressed as cyclic permutations of maximum length. To construct a game, given a number of goals $n$, an order $\sigma$ is sampled uniformly at random. The positively rewarding orders for the game are then fixed to be $\{\sigma, \sigma^{-1}\}$ where $\sigma^{-1}$ is the opposite direction of the order $\sigma$. An agent has a chance $\frac{2}{(n-1)!}$ of selecting a correct order at random at the start of each episode. In all our training and evaluation we use $n \geq 4$, so one is always more likely to guess incorrectly. Note that $\sigma$ and $\sigma^{-1}$ are equally difficult, equally rewarding options, provided that the world is not chiral.\footnote{That is to say, assuming there is no difference between the world and its mirror image from above with respect to performing a given trajectory.} 

Players receive a reward of $+1$ for entering a goal in the correct order, given the previous goals entered. The first goal entered in an episode always confers a reward of $+1$. If a player enters an incorrect goal, they receive a reward of $-1$ and must now continue as if this were the first goal they had entered. If a player re-enters the last goal they left, they receive a reward of $0$. The optimal policy is to divine a correct order, by experimentation or observation of an expert, and then visit the spheres in this cyclic order for the rest of the episode. At the start of each episode, goals are placed randomly in the world, subject to the constraint that they do not overlap, achieved via rejection sampling. Figure \ref{fig:goal-cycle-mechs} illustrates some of the game mechanics. 

\begin{figure}[htb]
\centering
\includegraphics[width=1.0\linewidth]{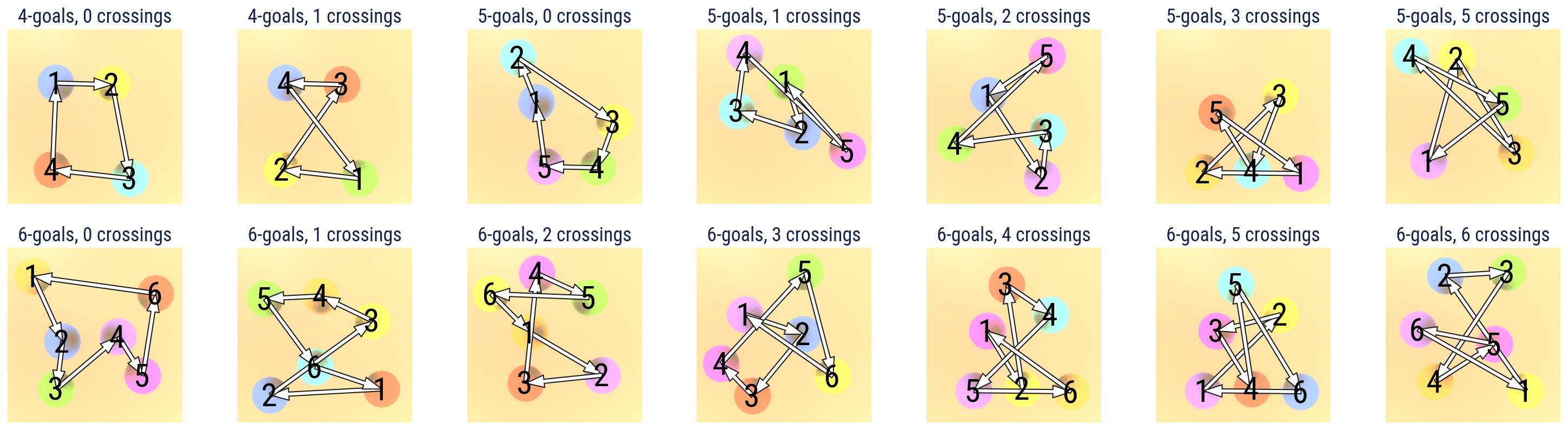}
\caption{An overview of all $4$- $5$-, and $6$-cycle topologies, characterised by the number of crossings in their shortest paths.}
\label{fig:game-crossings}
\end{figure}

We can classify paths between goals according to their topology. Distinct topologies are characterised by the number of self-intersections on the path, which we refer to as ``crossings''. Different topologies present players with different challenges, since the relative position of the next goal with respect to the others is altered. Moreover, topologies with a higher number of crossings tend to have longer paths. During training, we use rejection sampling to achieve a uniform distribution over topologies. Figure \ref{fig:game-crossings} depicts all possible topologies in games with 4, 5 and 6 goals, which we refer to as $4$-, $5$-, and $6$-cycle tasks.  

\subsection{Player interface}
\label{sec:player-interface}

\begin{figure}[ht]
\centering
\includegraphics[width=1\linewidth]{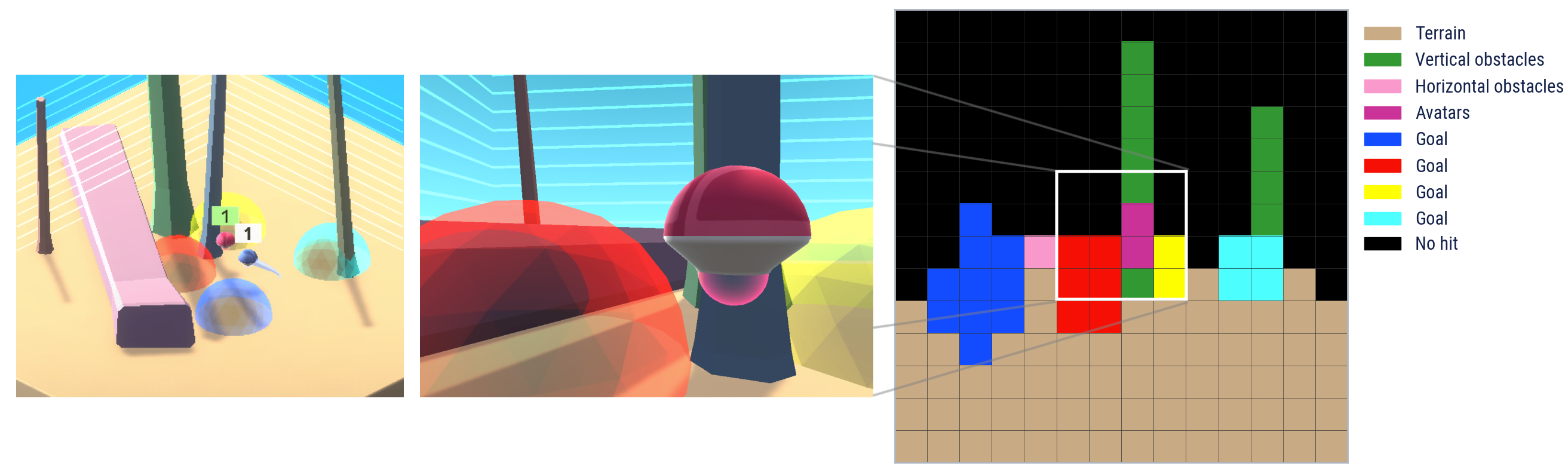}
\caption{(left) A third-person camera view at an instant in time. (centre) The human player RGB observation for the blue avatar. (right) A visualisation of the agent LIDAR observation for the same avatar. Each ray also reports the distance to the hit point, which is not shown in this rendering. The human player observation corresponds approximately to the highlighted central $4\times4$ array of the agent LIDAR observation.}
\label{fig:lidar-visualisation}
\end{figure}

Human players observe the environment through an egocentric first-person camera with a resolution of $640\times480$ pixels. Agents observe the environment through a LIDAR sensor. The LIDAR sensor performs raycasts uniformly distributed in polar coordinates, with azimuth ranging from $0\degree$ to $360\degree$, altitude ranging from $-90\degree$ to $+90\degree$, and a grid of $14\times 14$ rays. Each ray returns a one-hot encoding of the object with which the ray has intersected (vertical obstacle, horizontal obstacle, goal, avatar, terrain), its distance and, for goals only, its RGB color. The ray only returns the first object with which it collides. Therefore, for agents, all objects are opaque, including goals, and the boundary of the world is invisible. In addition, each agent is equipped with an AVATAR sensor, which outputs the $3$-dimensional relative distance of the nearest co-player in Cartesian coordinates in the frame of reference of the avatar.\footnote{This is used as a regression target during training but not passed as an input to the agent's neural network, so is not required at test time.} Figure \ref{fig:lidar-visualisation} compares the human RGB and agent LIDAR observations.

The action space is $5$-dimensional and continuous, with each action dimension taking values in $[-1,1]$. The five dimensions represent moving forwards and backwards, moving left and right, rotating left and right, rotating up and down, and jumping and crouching. Players may take any combination of actions simultaneously. Humans interact using a discretised action set, with each action having values in $\{-1,0,+1\}$, mapped to keyboard inputs. Movement dynamics are subject to inertia: players continue to move in their current direction, albeit at a diminishing rate, even when not sending any movement actions corresponding to that same direction.

Avatars can be controlled by a Unity scripted player, which we dub an \textit{expert bot}. Expert bots are ``oracles'', receiving privileged information about the correct order of goals to traverse, navigating using the Unity NavMesh \citep{vantoll2012}, and jumping and crouching when colliding with horizontal obstacles. These movement patterns are simple heuristics, so the expert bots are not guaranteed to find the most efficient trajectory from one goal to the next. 

\subsection{Probe tasks}\label{sec:probe_tasks}
\begin{figure}[h]
    \centering
    \begin{subfigure}{0.48\textwidth}
        \includegraphics[width=\textwidth]{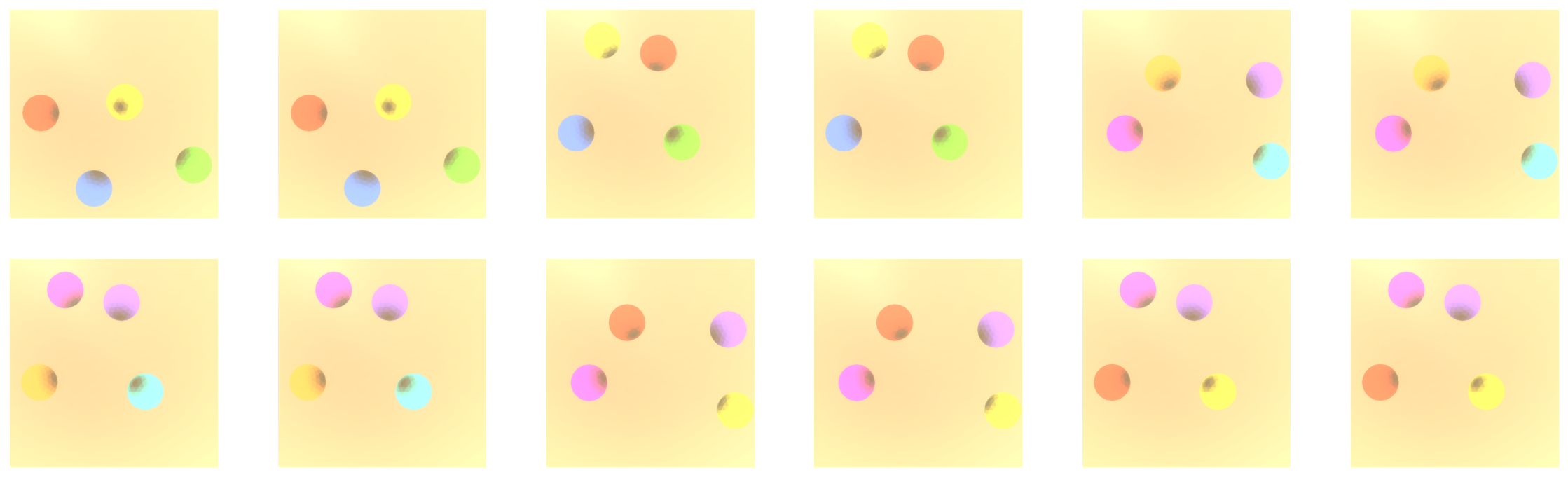}
        \caption{\centering Empty world, 4-goal games.}
        \label{fig:probes_simple_4}
    \end{subfigure}
    \hfill
        \begin{subfigure}{0.48\textwidth}
        \includegraphics[width=\textwidth]{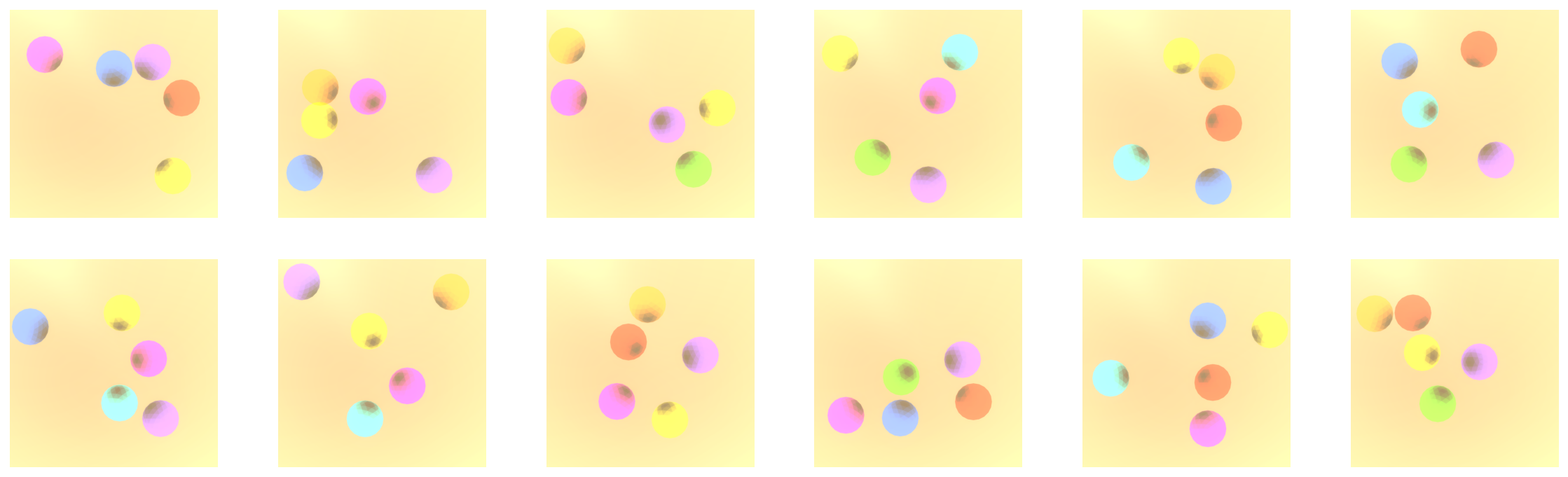}
        \caption{\centering Empty world, 5-goal games.}
        \label{fig:probes_simple_5}
    \end{subfigure}
    \begin{subfigure}{0.75\textwidth}
        \includegraphics[width=\textwidth]{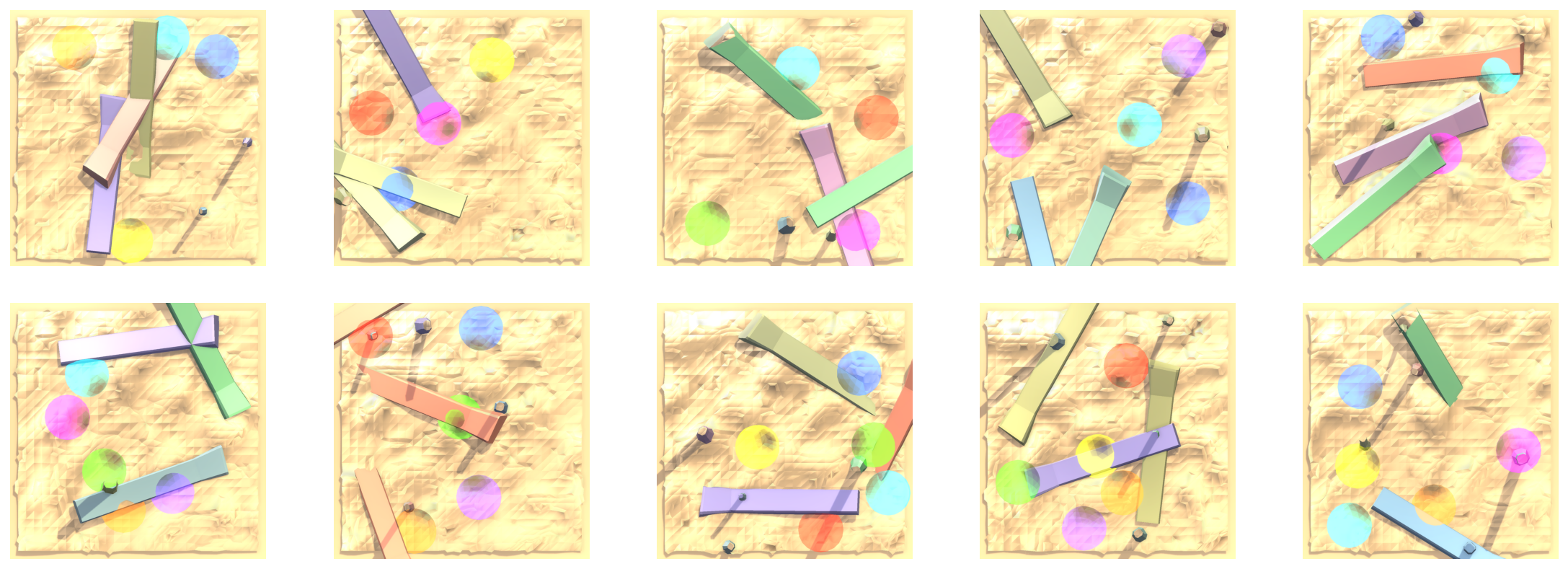}
        \caption{\centering Complex world, 4- and 5-goal games.}
        \label{fig:probes_complex}
    \end{subfigure}
    \caption{The worlds and games used as probe tasks. These cover a representative range of crossings and colour combinations. The complex world probes require clear examples of jumping behaviours and navigation around vertical obstacles. The human movement pattern in all probes is always goal-directed and near-optimal, but clearly different from a scripted bot, taking some time to get situated in the first few seconds and not taking an identical path on repeated cycles, for instance. See \href{https://sites.google.com/view/dm-cgi\#h.itlel1jrvsnn}{example videos}.
    }
    \label{fig:probe_tasks}
\end{figure}

To better understand and compare the performance of our agents under specific conditions, we test them throughout training on a set of ``probe tasks''. These tasks are hand picked and held out from the training distribution (i.e. they are not used to update any neural network weights). The worlds and games used in our probe tasks are shown in Figure~\ref{fig:probe_tasks}.

Importantly, these tasks are not chosen based on agent performance. Instead, they are chosen to represent a wide space of possible worlds, games, and co-players. For example, we desire cultural transmission both in worlds devoid of any obstacles and in worlds that are densely covered. Consequently, we included both in our set of probe tasks. We save checkpoints of agents throughout training at regular intervals and evaluate each checkpoint on the probe tasks. This yields a held-out measure of cultural transmission at different points during training, and is a consistent measure to compare across independent training runs.

While we seek to generate agents capable of robust real-time cultural transmission from human co-players, it is infeasible to measure this during training at the scale necessary for conducting effective research. Therefore we create ``human proxy'' expert co-players for use in our probe tasks as follows. A member of the team plays each task alone and with privileged information about the optimal cycle. For each task, we record the trajectory taken by the human as they demonstrate the correct path. We then replay this trajectory to generate an expert co-player in probe tasks, disabling the human proxy's collision mesh to ensure that the human trajectory cannot be interfered with by the agent under test. 

\clearpage
\section{Learning algorithm}\label{background}

Our hypothesis is that robust cultural transmission is a ``cognitive gadget'' \citep{heyes2018cognitive}. We show that cultural transmission can be reinforcement learned purely from individual reward \citep{SILVER2021103535}. We require only that the environment contains knowledgeable co-players and that the agent has a minimal sufficient number of representational and experiential biases (see Section \ref{sec:training}). From the perspective of an agent capable of cultural transmission from an expert, the policy of the expert must be inferred online using only the transition dynamics of the environment. In other words, the agent must solve the correspondence problem.  

If our agent learns a cultural transmission policy, it is by observing an expert agent in the world and correlating improved expected return with the ability to reproduce the other agent's behaviour. If that cultural transmission policy is robust, then reinforcement learning (RL) must have favoured imitation with high fidelity, generalising across a range of contexts, and recalling transmitted behaviours. As a basis for our claims, we use a state-of-the-art continuous control deep reinforcement learning algorithm, maximum a posteriori policy optimisation (MPO) \citep{abdolmaleki2018maximum}. For details of the RL formalism and MPO algorithm, see Appendix \ref{app:learning_algorithm}. 

A crucial challenge in RL is the balance between exploration and exploitation. Exploration involves discovering new parts of the state-action space and their implications for value; exploitation involves using learned information about valuable states and actions to gain reward \citep{kearns2002near}. Exploration is challenging when reward is sparse or deceptive and the state-action space is large. There may be many steps required to discover an improved strategy which forego the reward offered by exploitation. In such ``hard exploration'' problems, RL is prone to falling into local optima. We show how an independent agent can learn to model its environment, and particularly others therein, to solve held-out hard exploration problems.

Reinforcement learning of cultural transmission cannot occur in fully-observed tabular settings where the environment reward is unaffected by the behaviour of the expert. The proof of this is straightforward. For an agent to learn cultural transmission, they must be rewarded in states where the behaviour of an expert co-player is salient. However, in a fully observed setting, these states also contain perfect information about the environmental features. By assumption these features determine the agent's reward independently of the behaviour of the expert. If the setting is tabular, there is no aliasing of states, so the behaviour of the expert is irrelevant from the perspective of the Q-function. 

In our setting we violate the assumptions of this lemma in two ways: by operating in a partially observable setting, and by operating in a rich 3D physical world that necessitates function approximation. Under these conditions, with appropriate domain randomisation and attention biases, we show that reinforcement learning on environment reward alone is sufficient to learn a cultural transmission policy. The corresponding neural network aliases states in such a way that it can both infer hidden information from the expert and reuse that information after the expert has dropped out. 
\clearpage
\section{Methods}
\label{sec:training}

Here we describe the minimal sufficient list of ingredients for the emergence of robust real-time cultural transmission. Individually, none of the ingredients are complex. Yet, from these simple, general building blocks arises an agent with powerful generalisation and recall. 

\begin{figure}[htb]
    \centering
    \includegraphics[width=0.7\linewidth]{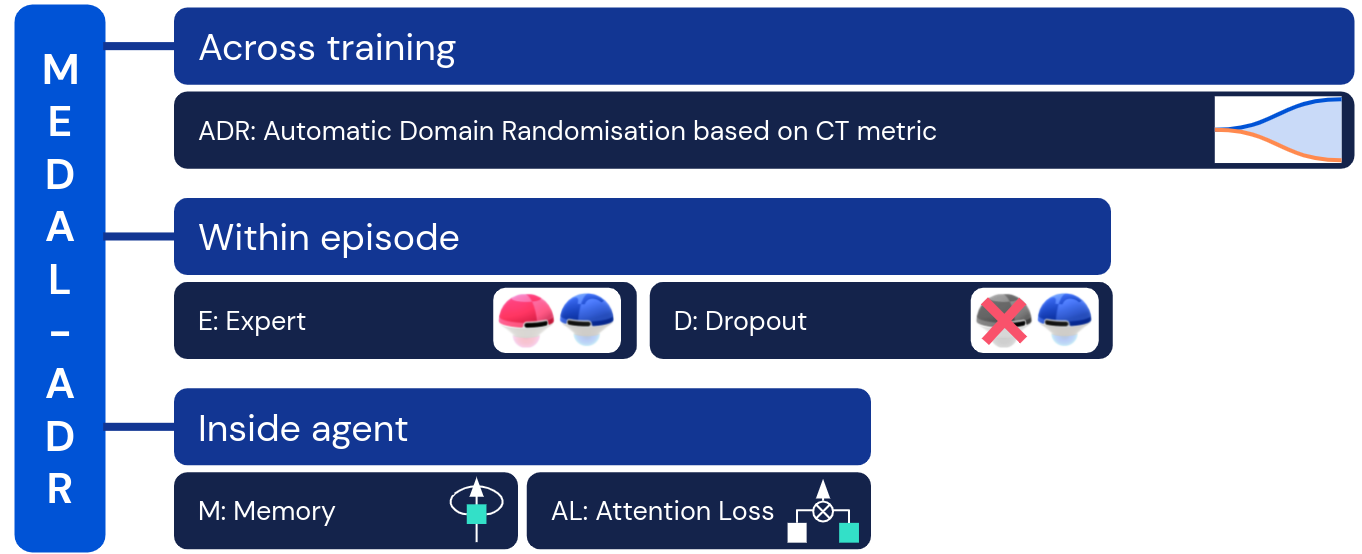}
    \caption{The minimal sufficient ingredients that comprise our methods, grouped by the timescale on which they operate.}
    \label{fig:method-overview}
\end{figure}

\subsection{Cultural Transmission (CT) metric}\label{sec:ctmetric}

To rigorously evaluate the cultural transmission capability of a fixed neural network, we must first define precisely what this means. Na\"ively, the agent must improve its performance upon witnessing an expert demonstration, and maintain that improvement within the same episode once the demonstrator has departed. But this is not quite a sufficient condition, for two reasons. 

First, what seems like test-time cultural transmission might actually be cultural transmission during training, leading to memorisation (and subsequent deployment) of fixed navigation routes. To address this, we evaluate cultural transmission in held-out test tasks and with human expert demonstrators \citep{cobbe2020leveraging, anand2021procedural, justesen2018illuminating}, similar to the familiar train-test dataset split in supervised learning \citep{xu2018splitting}. Capturing this intuition, we define a scalar \textit{cultural transmission metric} below. 

Second, what looks like test-time cultural transmission may in fact merely be independent behaviour, conditioned on the presence of a co-player, but not the information they bear. To address this, we borrow the ``two-option paradigm'', standard in studies of animal imitation \citep{Dawson1965ObservationalLI,whiten1998imitation, aplin2015experimentally}. Every one of our tasks contains two different, equally difficult, and equally rewarding options, of which the expert demonstrates only one. We find that our agent consistently displays the same option as the expert, both when together with the expert and after the expert has dropped out. This provides evidence of test-time cultural transmission (see Section \ref{sec:fidelity}). Crucially, there is no option bias during training or testing: in any given task we sample the expert's direction uniformly at random from the $2$ optimal possibilities. 

Let $E$ be the total score achieved by the expert in an episode of a given task. Let $A_{\text{full}}$ be the score of an agent with the expert present for the full episode. Let $A_{\text{solo}}$ be the score of the same agent without the expert. Finally, let $A_{\text{half}}$ be the score of the agent with the expert present from the start to halfway into the episode. These tasks are implemented by using the expert dropout types `No', `Full', and `Half', described in Section~\ref{sec:expert_dropout}. We define the cultural transmission (CT) metric as

\begin{equation}\label{eqn:ctmetric}
    \textrm{CT} \defeq \frac{1}{2}\frac{A_{\text{full}} - A_{\text{solo}}}{E} + 
    \frac{1}{2}\frac{A_{\text{half}} - A_{\text{solo}}}{E} \, .
\end{equation}

We identify several important values for CT. A completely independent agent doesn't use any information from the expert. Therefore it has a value of CT near $0$, since $A_{\text{full}}$, $A_{\text{solo}}$, and $A_{\text{half}}$ are all similar. A fully expert-dependent agent has a value of CT near $0.75$. This is because $A_{\text{full}} \approx E$, $A_{\text{solo}}\approx 0$ and $A_{\text{half}}\approx 0.5 E$, making the first term close to $0.5$ and the second term close to $0.25$. An agent that follows perfectly when the expert is present, but continues to achieve high scores once the expert is absent has a value of CT near $1$. This is the desired behaviour of an agent from a cultural transmission perspective, since the knowledge about how to solve the task was transmitted to, retained by and reproduced by the agent. Note that CT does not distinguish well between perfect following and the combination of imperfect following and imperfect recall. This motivates the more detailed analysis of our agent's capabilities in Section \ref{sec:evaluation}.

\subsection{Agent architecture and Memory (M)}

\begin{figure}[htb]
    \centering
    \includegraphics[width=0.5\linewidth]{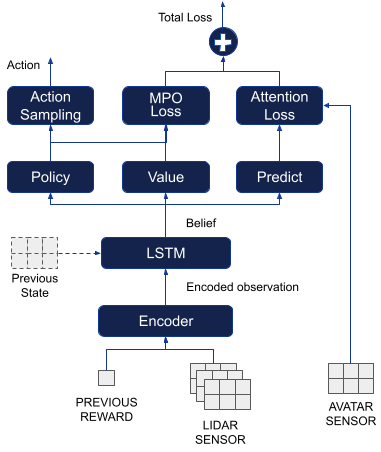}
    \caption{Agent architecture, including training losses.}
    \label{fig:agent-architecture}
\end{figure}

Our agent architecture is depicted in Figure \ref{fig:agent-architecture}. The encoded observation is fed to a single-layer recurrent neural network (RNN) with an LSTM core \citep{hochreiter1997lstm} of size $512$. This RNN is unrolled for $800$ steps during training. The output of the LSTM, which we refer to as the \textit{belief}, is passed to a policy, value and auxiliary prediction head. The policy and value heads together implement the MPO algorithm, while the prediction head implements the attention loss described in Section \ref{sec:auxloss}. The AVATAR sensor observation is used as a prediction target for this loss. Our agent is trained using a large scale distributed training framework, described in Appendix \ref{sec:training-framework}. Further details of the hyperparameters used for learning are reported in Appendices
\ref{sec:hyperparameters}.

\subsection{Expert Dropout (ED)}\label{sec:expert_dropout}

Cultural transmission requires the acquisition of new behaviours from others. For an agent to robustly demonstrate cultural transmission, it is not sufficient to imitate instantaneously; the agent must also internalise this information, and later recall it in order to display the transmitted behaviour. We introduce \textit{expert dropout} as a mechanism both to test for and to train for this recall ability. 

At each timestep in an episode, the expert is rendered visible or hidden from the agent. Given a difficult exploration task which the agent cannot solve through solo exploration, we can measure its ability to recall information gathered through cultural transmission by observing whether or not it is able to solve the task during the contiguous steps in which the expert is hidden. During training, we apply expert dropout in an automatic curriculum to encourage this recall ability, as described in Section \ref{sec:adr}.

Mathematically, we formulate expert dropout as follows. Let $e_t \in \mathbb{Z}_2$ be the \emph{state} of expert dropout at timestep $t$. State $0$ corresponds to the expert being hidden at time $t$, by which we mean it will not be detected in any player's observation. State $1$ corresponds to the expert being visible at time $t$. An expert dropout scheme is characterised by the state transition functions $e_{t+1} = f(e_t, t)$. We define the following schemes:
\begin{multicols}{2}
\paragraph{No dropout} $f(e_t, t) = 1$ for all $t$.
\columnbreak
\paragraph{Full dropout} $f(e_t, t) = 0$ for all $t$.
\end{multicols}
\paragraph{Half dropout} For episodes of length $N$ timesteps, 
\begin{equation}
f(e_t, t) = 
\begin{cases}
1 & t \leq \lfloor N/2 \rfloor \\
0 & t > \lfloor N/2 \rfloor \\
\end{cases}.
\end{equation}

\paragraph{Probabilistic dropout} Given transition probability $p \in [0, 1]$,
\begin{equation}
    f(e_t, t) = 
    \begin{cases}
    e_t + 1 \mod 2& \text{with probability $p$} \\
    e_t & \text{with probability $1 - p$} \\
    \end{cases}.
\end{equation}

\subsection{Attention Loss (AL)}\label{sec:auxloss}

To use social information, agents need to notice that there are other players in the world that have similar abilities and intentions as themselves \citep{ndousse2020learning, DBLP:journals/corr/abs-1802-07740}. Agents observe the environment without receiving other players' explicit actions or observations, which we view as privileged information. Therefore, we propose an \textit{attention loss} which encourages the agent's belief to represent information about the current relative position of other players in the world. We use ``attention'' here in the biological sense, identifying what is important, in particular, that agents should pay attention to their co-players. Similar to previous work (e.g. \citet{DBLP:journals/corr/abs-1906-01470}), we use a privileged AVATAR sensor as a prediction target, but not as an input to the neural network, so it is not required at test time. 

Starting from the belief, we concatenate the agent's current action, pass this through two MLP layers of size $32$ and $64$ with \textit{relu} activations, and finally predict the egocentric relative position of other players in the world at the current timestep. The objective is to minimise the $\ell^1$ distance between the ground truth and predicted relative positions. The attention loss is set to zero when the agent is alone (for instance, when the expert has dropped out). For more details, including empirical determination of attention loss hyperparameters, see Appendix \ref{app:bonus-attention-loss}.

\subsection{Automatic Domain Randomisation (ADR)}\label{sec:adr}

An important ingredient for the development of cultural transmission in agents is the ability to train over a diverse set of tasks. Without diversity, an agent can simply learn to memorise an optimal route. It will pay no attention to an expert at test-time and its behaviour will not transfer to distinct, held-out tasks. This diverse set of tasks must be adapted according to the current ability of the agent. In humans, this corresponds to Vygotsky's concept of a dynamic ``Zone of Proximal Development'' (ZPD). \citep{vygotsky1980mind}. This is defined to be the difference between a child's ``actual development level as determined by independent problem solving'' and ``potential development as determined through problem solving under adult guidance or in collaboration with more capable peers''. We also refer to the ZPD by the more colloquial term ``Goldilocks zone'', one where the difficulty of the task is not too easy nor too hard, but just right for the agent.

We use Automatic Domain Randomisation (ADR) \citep{openai2019} to maintain task diversity in the Goldilocks zone for learning a test-time cultural transmission ability. To apply ADR, each task must be parameterised by a set of $d$ parameters, denoted by $\lambda \in \mathbb{R}^d$. In \textit{GoalCycle3D}, these parameters may be related to the world, such as terrain size or tree density, the game, such as number of goals, or the co-players, such as bot speed.

Each set of task parameters $\lambda$ are drawn from a distribution $P_{\phi}(\Lambda)$ over the $(d-1)$-dimensional standard simplex, parameterised by a vector $\phi$. We use a product of uniform distributions with $2d$ parameters and joint cumulative density function
\begin{equation}
P_{\phi}(\lambda) = \prod_{i=1}^d \frac{1}{\phi_i^L - \phi_i^H} \, ,
\end{equation}
defined over the standard simplex given by 
\begin{equation}
\left\{
\lambda: \lambda_i \in [\phi_i^L,\, \phi_i^H] \text{ for } 
i \in \{1,\dots,d\}\text{, }
\lambda \in \mathbb{R}^d
\right
\} \, .
\end{equation}

Roughly speaking, the simplex boundaries $\phi_{i}^L$ or $\phi_{i}^H$ are expanded if the training cultural transmission metric exceeds an upper threshold and contracted if the training cultural transmission metric drops below a lower threshold. This maintains the task distribution in the Goldilocks zone for learning cultural transmission. For more details, see Appendix \ref{sec:app_adr}.
\clearpage
\section{Training and evaluation}
\label{sec:results}

Our agent initialisation and procedural world generation algorithm use seeds for their pseudorandom number generators. We first describe in detail results for individual representative seeds, then provide mean and variance statistics across $10$ seeds in Section \ref{sec:ablation-results}. We distinguish between the \textit{training} CT metric, measured on tasks sampled from the training task distribution, and the \textit{evaluation} CT metric, measured on held-out probe tasks defined in Section \ref{sec:probe_tasks}. In both cases, we compute a scalar CT metric from the per-task CT metrics by taking a uniform average. For further results, see Appendix \ref{app:training}.

\subsection{Training without ADR}

\begin{figure}[h]
    \centering
    \includegraphics[width=0.8\textwidth]{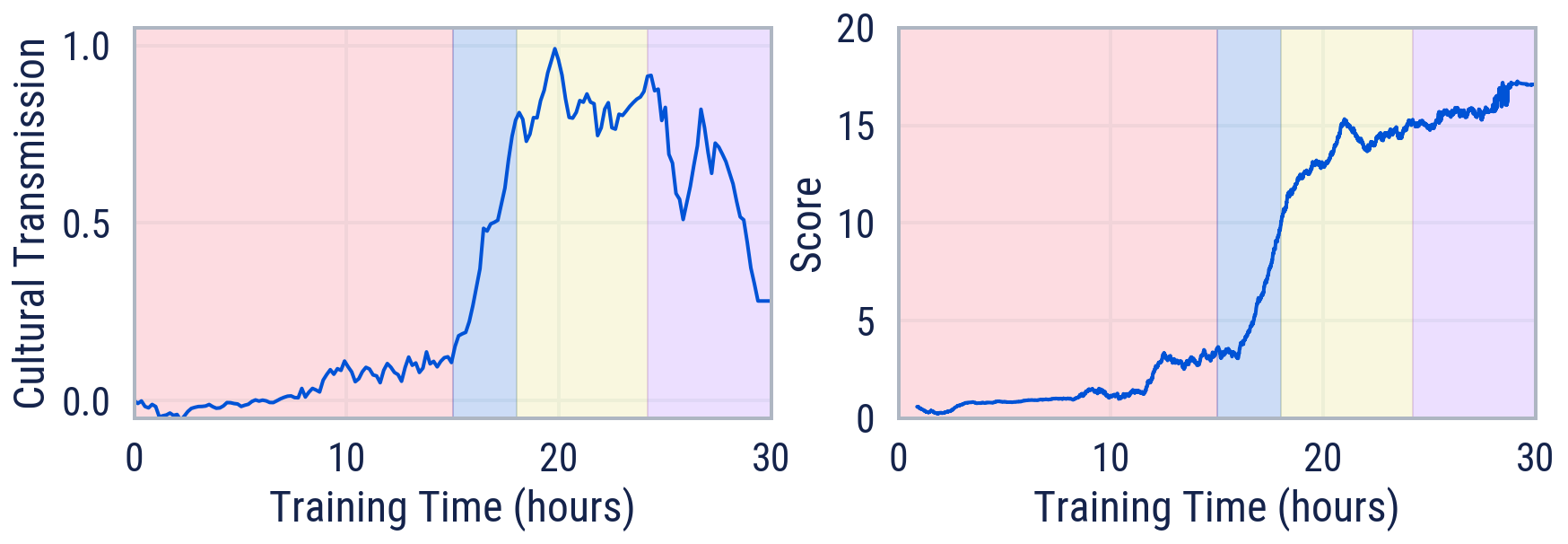}
    \caption{Training cultural transmission metric and agent score for training without ADR on 4-goal, $16 \times 16$ world with no obstacles and flat terrain. Colours indicate four distinct phases of agent behaviour: (1) (red) startup and exploration, (2) (blue) learning to follow, (3) (yellow) learning to remember, (4) (purple) becoming independent from expert.}
    \label{fig:adr-none}
\end{figure}

In this experiment, we do not control any task parameter using ADR. We study a simple task consisting of a 4-goal game in a $16 \times 16$ world with no obstacles and flat terrain. Figure~\ref{fig:adr-none} shows the training cultural transmission metric and the agent score as training progresses. 

The training cultural transmission metric shows four distinct phases over the training run, each corresponding to a distinct social learning behaviour of the agent. In phase 1 (red), the agent starts to familiarise itself with the task, learns representations, locomotion, and explores, without much improvement in score. In phase 2 (blue), with sufficient experience and representations shaped by the attention loss, the agent learns its first social learning skill of following the expert bot to solve the task. The training cultural transmission metric increases to $0.75$, which suggests pure following. 

In phase 3 (yellow), the agent learns the more advanced social learning skill that we call cultural transmission. It remembers the rewarding cycle while the expert bot is present and retrieves that information to continue to solve the task when the bot is absent. This is evident in a training cultural transmission metric approaching $1$ and a continued increase in agent score. 

Lastly, in phase 4 (purple), the agent is able to solve the task independently of the expert bot. The training cultural transmission metric falls back towards $0$ while the score continues to increase. Together, this suggests that the agent is able to achieve high scores with or without the bot. It has learned an ``experimentation'' behaviour, which involves using hypothesis-testing to infer the correct cycle without reference to the bot, followed by exploiting that correct cycle more efficiently than the bot does, resulting in a continuing increase in score. We do not see this experimentation behaviour emerge in the absence of prior social learning abilities: social learning creates the right prior for experimentation to emerge. Example \href{https://sites.google.com/view/dm-cgi\#h.m5zueafwx4p}{videos} corroborate our observations.

\subsection{Training with single-parameter ADR}
\label{sec:adr-single-param}
\begin{figure}[h]
    \centering
     \begin{subfigure}[b]{0.32\textwidth}
         \centering
         \includegraphics[width=\textwidth]{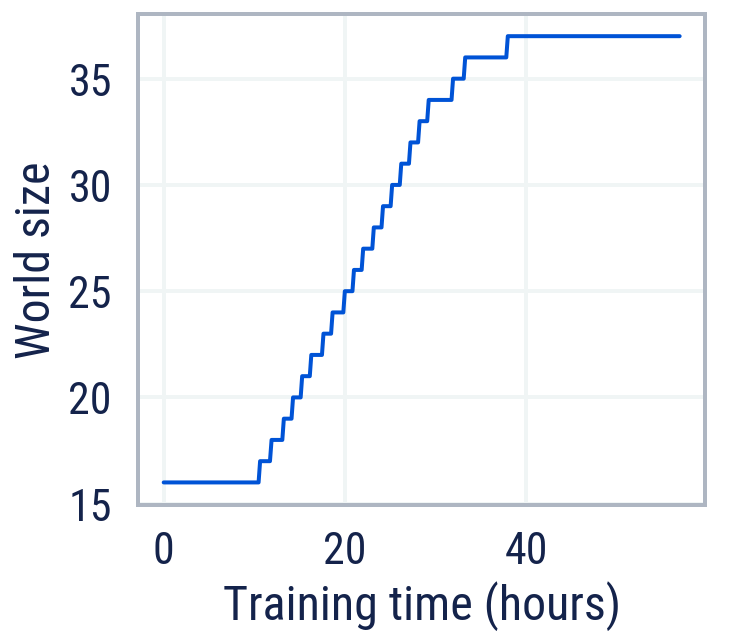}
         \label{fig:adr:world_size:upper_boundary}
     \end{subfigure}
     \hfill
     \begin{subfigure}[b]{0.32\textwidth}
         \centering
         \includegraphics[width=\textwidth]{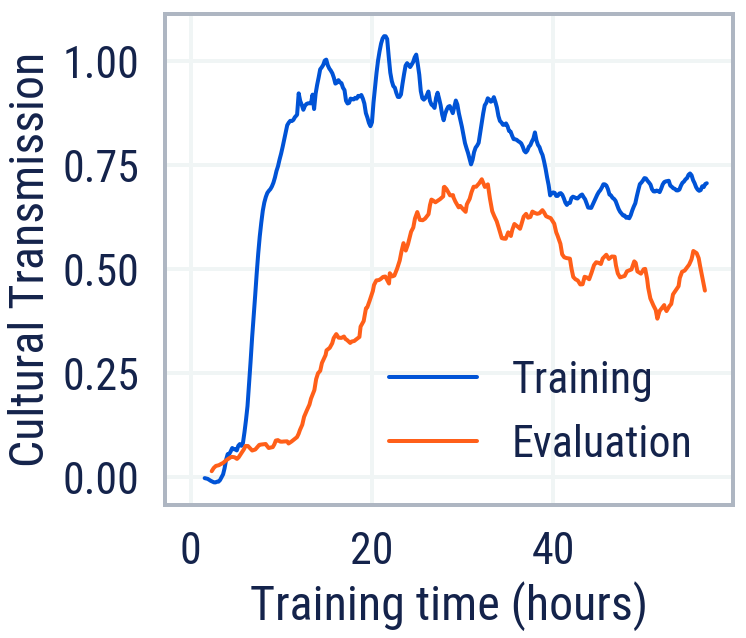}
         \label{fig:adr:world_size:ct}
     \end{subfigure}
    \hfill
     \begin{subfigure}[b]{0.32\textwidth}
         \centering
         \includegraphics[width=\textwidth]{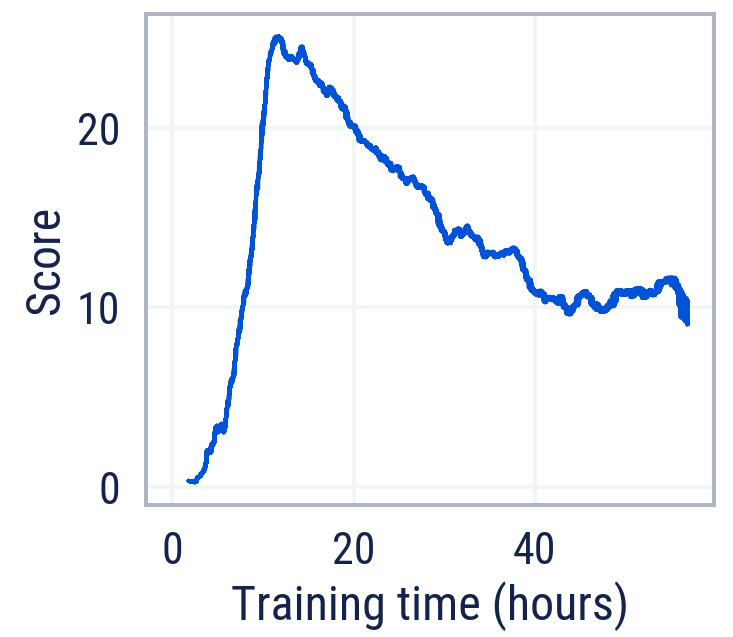}
         \label{fig:adr:world_size:score}
     \end{subfigure}
     \caption{Training curves for training with ADR controlling the world size parameter. (left) The upper boundary of the world size randomisation range, the lower boundary is fixed at $16$. (centre) The cultural transmission metric as measured from training environments and held-out probe tasks. (right) The average avatar score over the training run. Note that the decreasing score does not indicate lower performance, since the maximum achievable score decreases with increasing world size.
     }
     \label{fig:adr-world-size}
\end{figure}

In this experiment, we use ADR to increase task complexity appropriately to maintain the Goldilocks zone for the learning of cultural transmission, controlling the world size of the training tasks. Figure~\ref{fig:adr-world-size} shows the training curves for the experiment.

The difference between Figures~\ref{fig:adr-none} and~\ref{fig:adr-world-size} is clear. In Figure~\ref{fig:adr-world-size}, the training cultural transmission increases to near $1$ and remains above $0.7$ for the duration of the experiment without dropping towards $0$. This shows the effect of applying ADR to a single parameter. At the start of training, before the world size upper boundary started to increase, all worlds are of size $16 \times 16$. This is the same as in the previous experiment, and we see the same progression in the training cultural transmission metric through phases 1 to 3 as in Figure~\ref{fig:adr-none}. However, as the training cultural transmission metric increases above a threshold ($0.8$ in this experiment), the upper boundary begins to increase in steps of $1$. This continues until the maximum world size of $37 \times 37$ is reached. The progression of novel tasks prevents the collapse of the CT metric towards $0$: our agent no longer enters phase 4.

\subsection{Training with multi-parameter ADR} \label{sec:full-adr}
In this experiment, we train our best cultural transmission agent by using ADR to control the set of task parameters in Table~\ref{tab:adr-parameters}. The parameters include changes to the world that make navigating the terrain more difficult (world size, bumpiness), additional skills to be learned (vertical and horizontal obstacles), and changes to the expert behaviour (bot speed, dropout transition probability). Together, these parameters interact to create tasks that increase in complexity and decrease in expert reliability.

\begin{figure}[h]
    \centering
    \includegraphics[width=\textwidth]{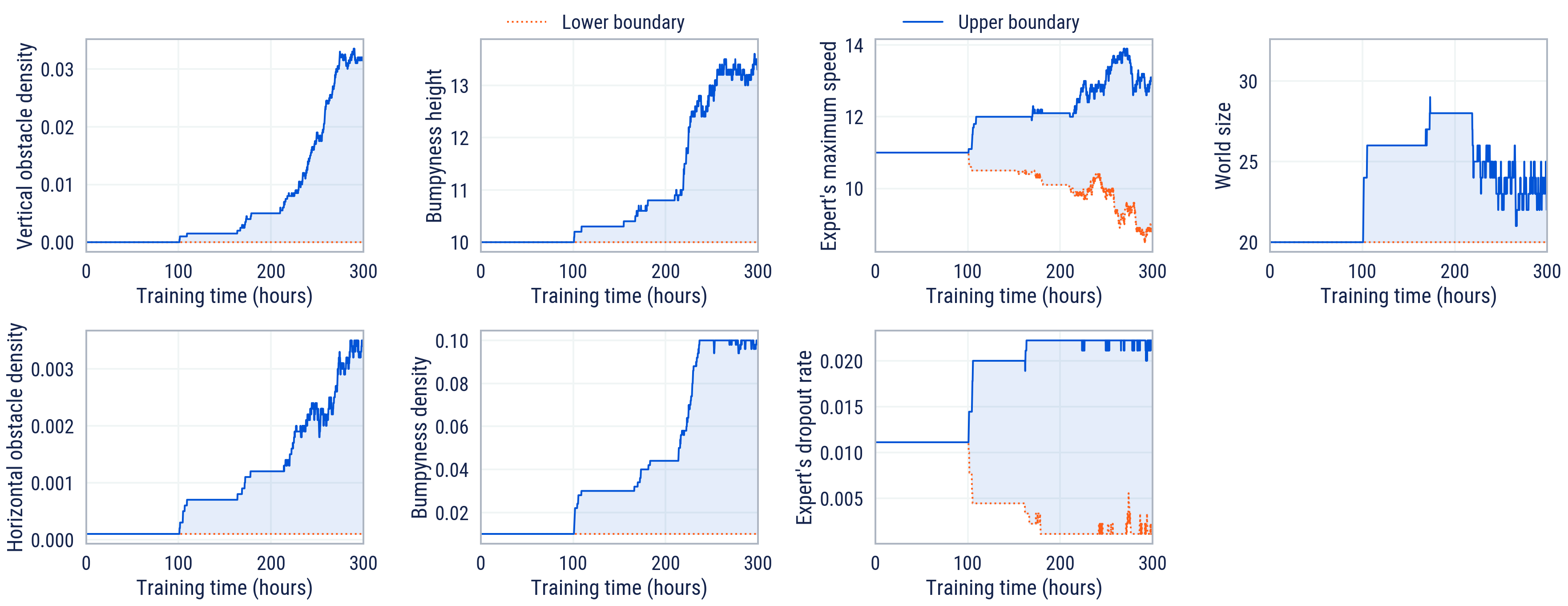}
    \caption{Expansion of parameter randomisation ranges over time, as controlled by ADR. See Figure \ref{fig:adr-full-training-local-ct} for training CT metrics used to decide whether to expand or contract range boundaries. Note a sharp reduction and prolonged instability in world size upper boundary after 200 hours of training. This is likely due to the lack of coordination between parameter boundaries under ADR.}
    \label{fig:adr-full-training}
\end{figure}

Figure~\ref{fig:adr-full-training} shows the expansion of the randomisation ranges for all parameters for the duration of the experiment. The training cultural transmission metric is maintained between the boundary update thresholds $0.75$ and $0.85$. We see an initial start-up phase of around 100 hours when social learning first emerges in a small, simple set of tasks. Once the training cultural transmission metric exceeds $0.75$, all randomisation ranges began to expand. Different parameters expand at different times, indicating when the agent has mastered different skills such as jumping over horizontal obstacles or navigating bumpy terrain. For intuition about the meaning of the parameter values, see \href{https://sites.google.com/view/dm-cgi\#h.vwuspybn66tu}{example videos} at different times during training.

\begin{figure}[h]
    \centering
    \includegraphics[width=0.9\textwidth]{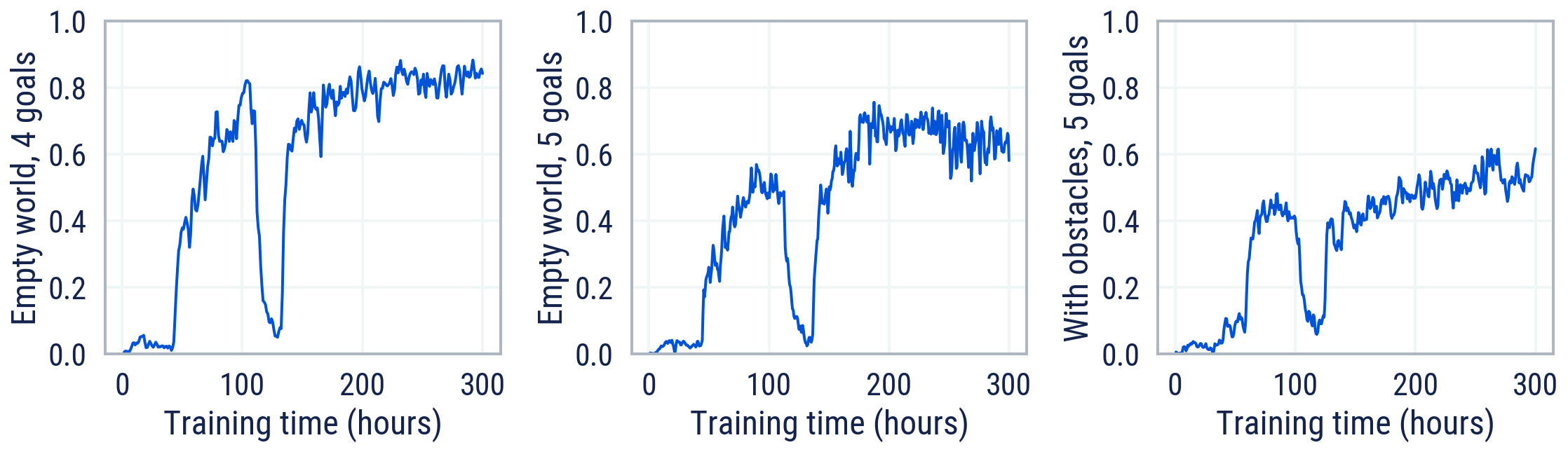}
    \caption{Evaluation cultural transmission metric in three sets of probe tasks.}
    \label{fig:adr-full-training-eval-ct}
\end{figure}

Figure~\ref{fig:adr-full-training-eval-ct} shows the evaluation cultural transmission metric over time. Note that all probe tasks are strictly out-of-distribution: during training the world size never expands to $32 \times 32$, and human demonstrations are qualitatively different from an expert bot. Despite these differences, the evaluation CT metric on empty 4-goal tasks reaches a final value of $0.85$, indicating good generalisation and recall. In the 5-goal tasks, the evaluation CT metric is lower. As we show in Section \ref{sec:evaluation}, our agent is still capable of recall and generalisation with 5 goals, but performance drops off in larger worlds. The prominent `dip' in evaluation cultural transmission metrics after 100 training hours is due to ADR starting to expand randomisation ranges at around 100 training hours. This leads to a momentary drop in social learning ability, which is amplified by the difficulty of evaluation tasks. ADR pauses the expansion until the agent recovers at around 150 hours.

\subsection{Evaluation via ablations}
\label{sec:ablation-results}

Via ablations, we identify a minimal sufficient set of representational and experiential biases which give rise to cultural transmission. The five ingredients are memory (M), expert demonstrations (E), dropout (D), an attention loss (AL), and automatic (A) domain randomisation (DR). We refer to an agent trained with all of these ingredients as MEDAL-ADR. In this section, we ablate each component in turn, showing their impact on performance and cultural transmission capability in empty and complex probe tasks. For more details, see Appendix \ref{app:ablation}. 

\begin{figure}[H]
    \centering
    \begin{subfigure}[b]{0.27\textwidth}
        \centering
        \includegraphics[width=\textwidth]{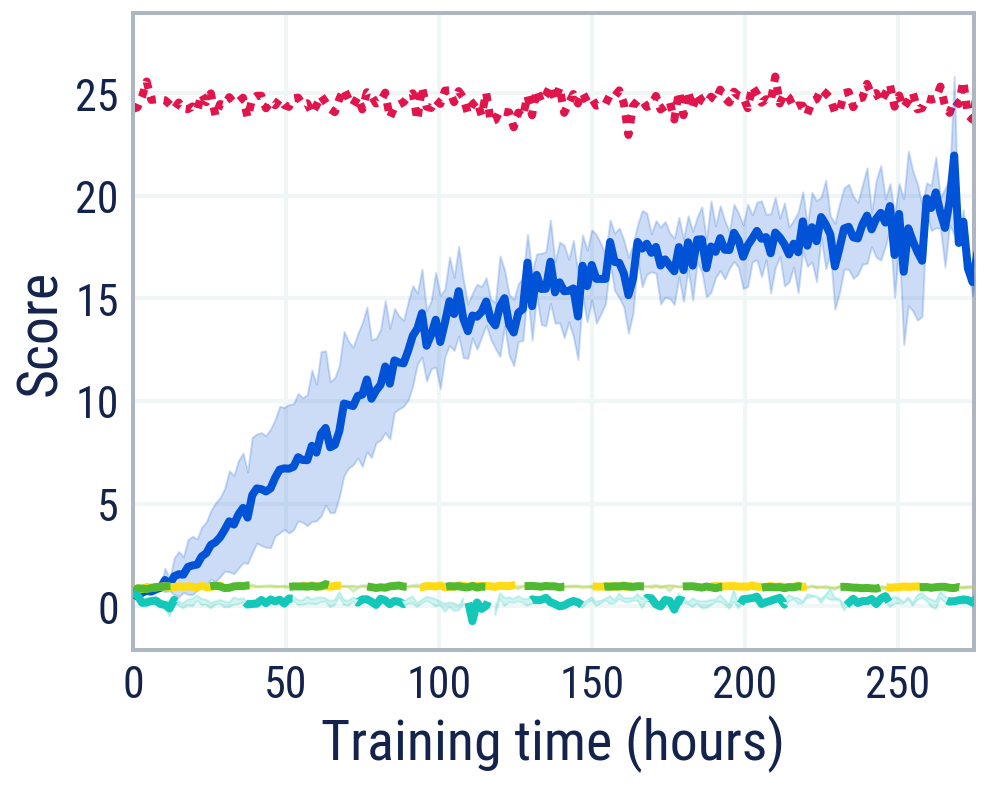}
        \label{subfig:ablate-0-agents-a}
    \end{subfigure}
    \hfill
    \begin{subfigure}[b]{0.27\textwidth}
        \centering
        \includegraphics[width=\textwidth]{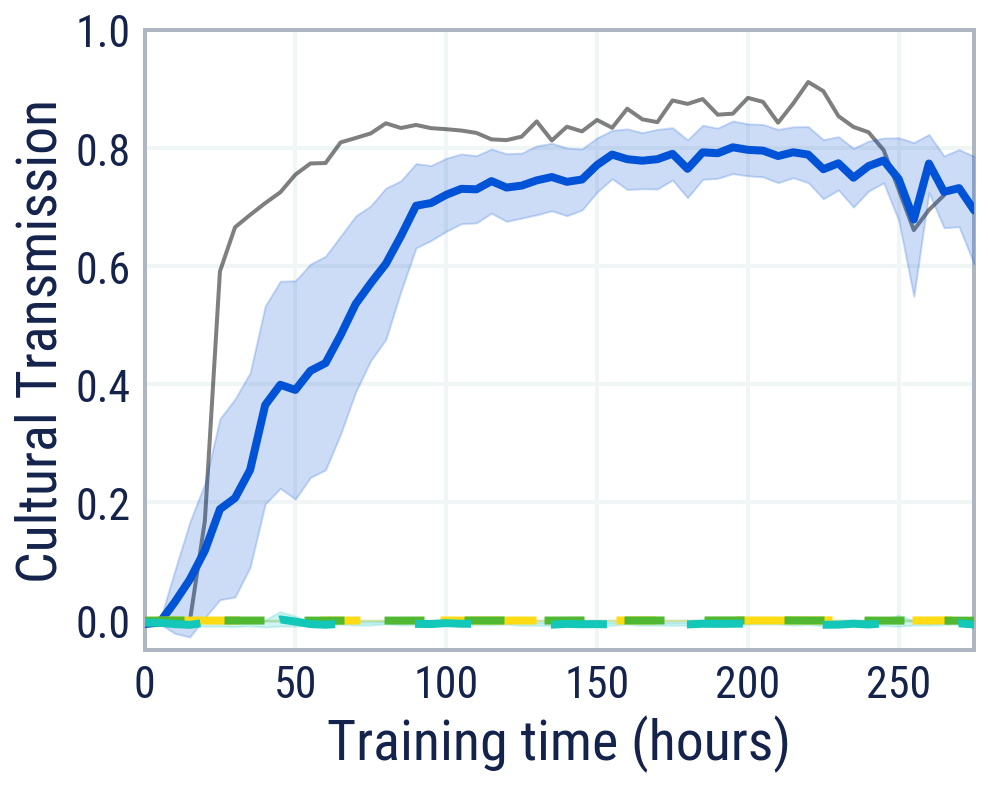}
        \label{subfig:ablate-0-agents-b}
    \end{subfigure}
    \hfill
    \begin{subfigure}[b]{0.27\textwidth}
        \centering
        \includegraphics[width=\textwidth]{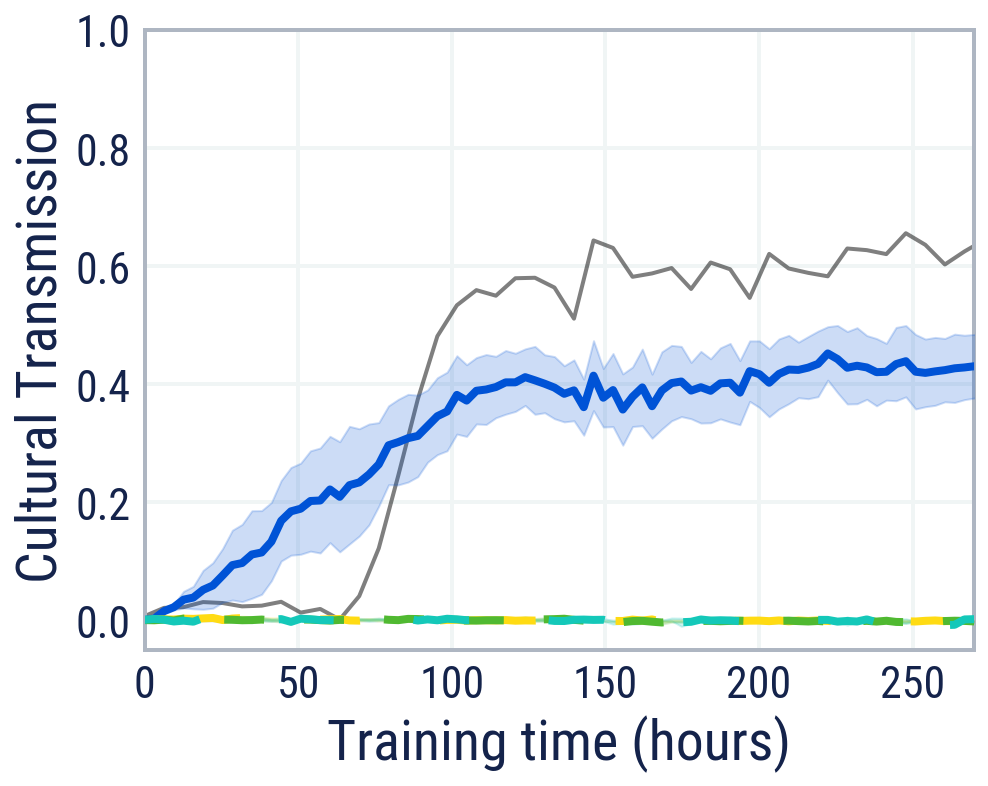}
        \label{subfig:ablate-0-agents-c}
    \end{subfigure}
    \hfill
    \begin{subfigure}[b]{0.15\textwidth}
        \centering
        \includegraphics[width=\textwidth]{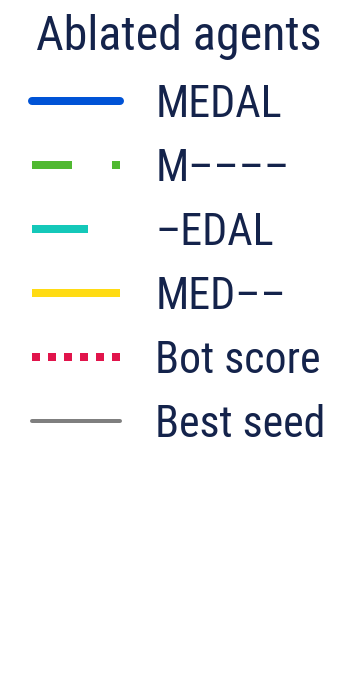}
    \end{subfigure}
    \caption{Three different ablated agents, each trained without one crucial ingredient: without an expert (M––––), memory (–EDAL), or attention loss (MED––). The evaluation CT is calculated on empty world $5$-goal probe tasks. All of these agents fail to even partially solve the task (left). They are unable to rely on social cues to acquire cultural information during training (centre) and, consequently, during testing (right). The result for M–––– shows that our task space presents a hard exploration challenge, since this is a pure RL baseline.}
    \label{fig:ablate-0-agents}
\end{figure}

\begin{figure}[H]
    \centering
    \begin{subfigure}[b]{0.27\textwidth}
        \centering
        \includegraphics[width=\textwidth]{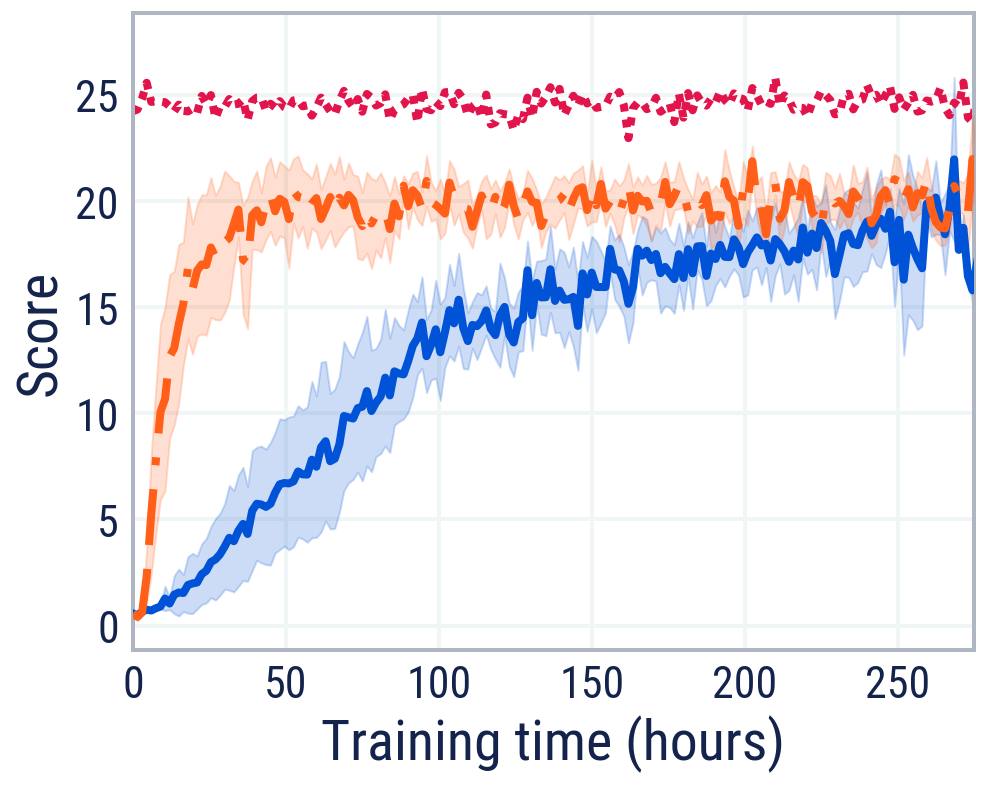}
        \label{subfig:ablate-dropout-a}
    \end{subfigure}
    \hfill
    \begin{subfigure}[b]{0.27\textwidth}
        \centering
        \includegraphics[width=\textwidth]{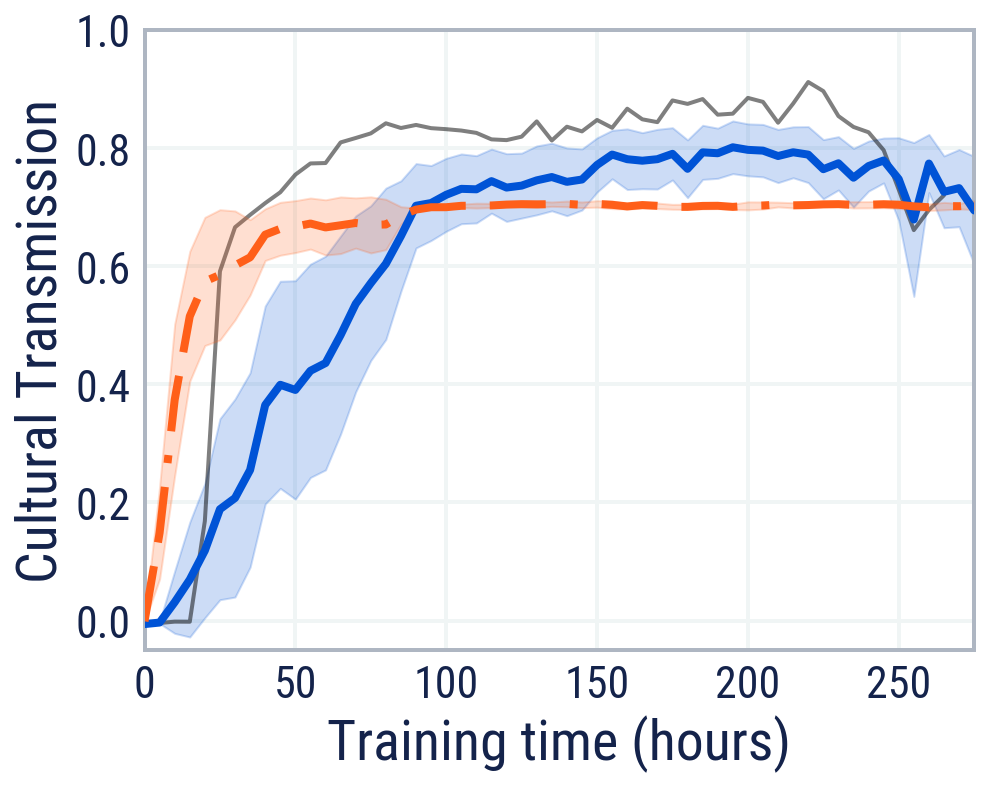}
        \label{subfig:ablate-dropout-b}
    \end{subfigure}
    \hfill
    \begin{subfigure}[b]{0.27\textwidth}
        \centering
        \includegraphics[width=\textwidth]{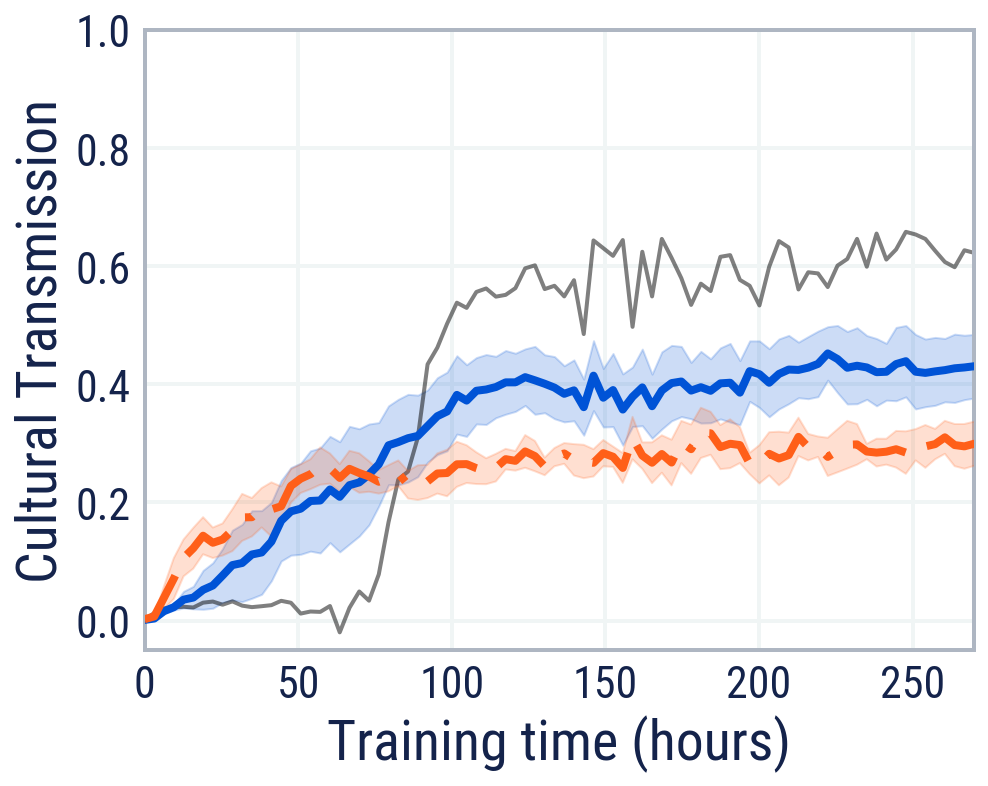}
        \label{subfig:ablate-dropout-c}
    \end{subfigure}
    \hfill
    \begin{subfigure}[b]{0.15\textwidth}
        \centering
        \includegraphics[width=\textwidth]{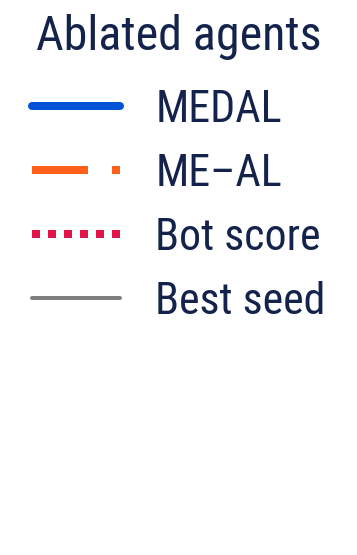}
    \end{subfigure}
    \caption{Isolating the importance of expert dropout, we compare our MEDAL agent with the ablated ME–AL agent. ME–AL is more sample efficient (left), but MEDAL achieves higher CT during training (centre) and when evaluated on empty world $5$-goal probe tasks (right). This is because MEDAL is capable of within-episode recall, whereas ME-AL is not. ME-AL represents the previous state-of-the-art method \citep{ndousse2020learning}.}
    \label{fig:ablate-dropout}
\end{figure}

\begin{figure}[H]
    \centering
    \begin{subfigure}[b]{0.27\textwidth}
        \centering
        \includegraphics[width=\textwidth]{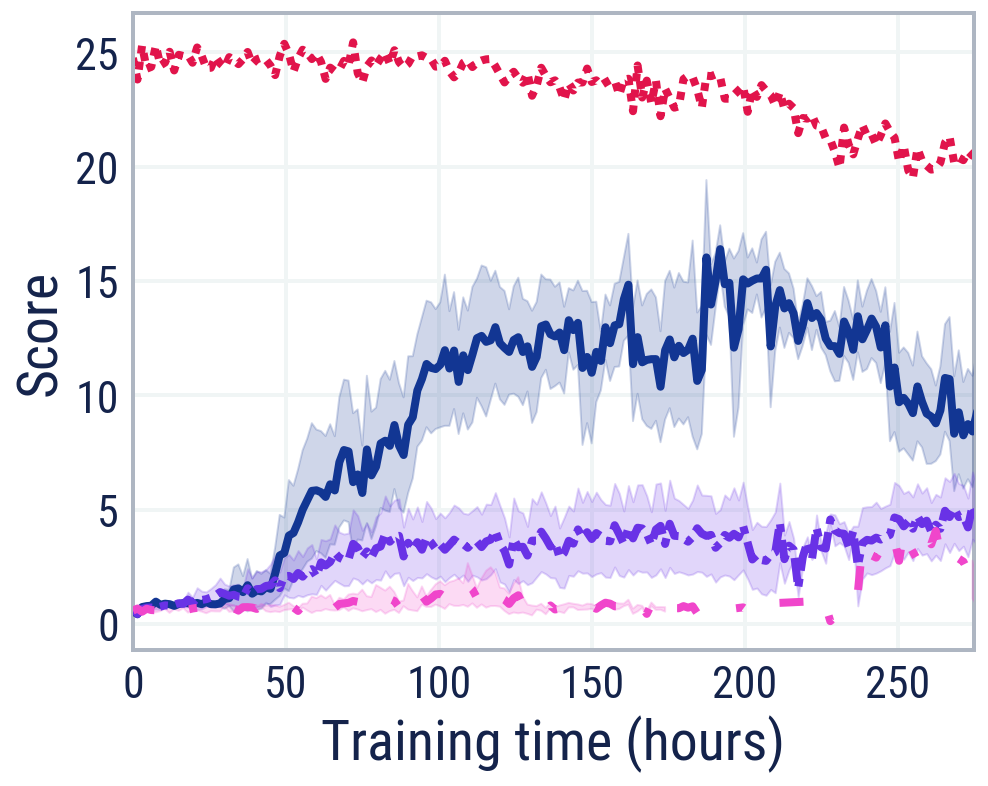}
        \label{subfig:ablate-adr-a}
    \end{subfigure}
    \hfill
    \begin{subfigure}[b]{0.27\textwidth}
        \centering
        \includegraphics[width=\textwidth]{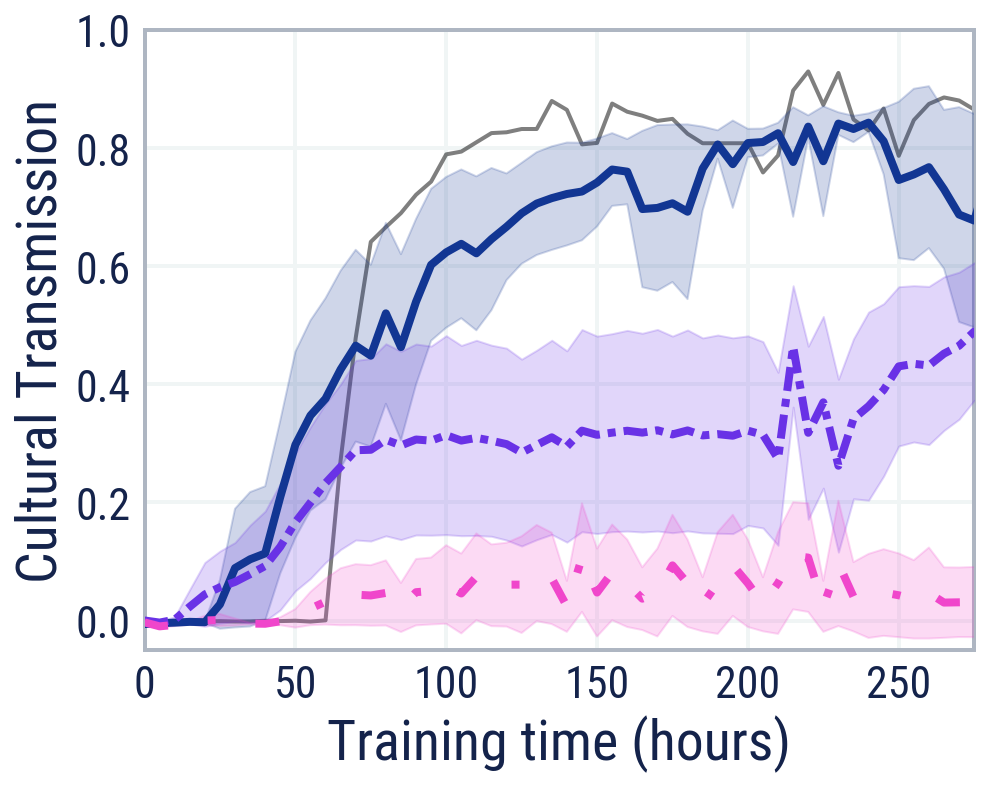}
        \label{subfig:ablate-adr-b}
    \end{subfigure}
    \hfill
    \begin{subfigure}[b]{0.27\textwidth}
        \centering
        \includegraphics[width=\textwidth]{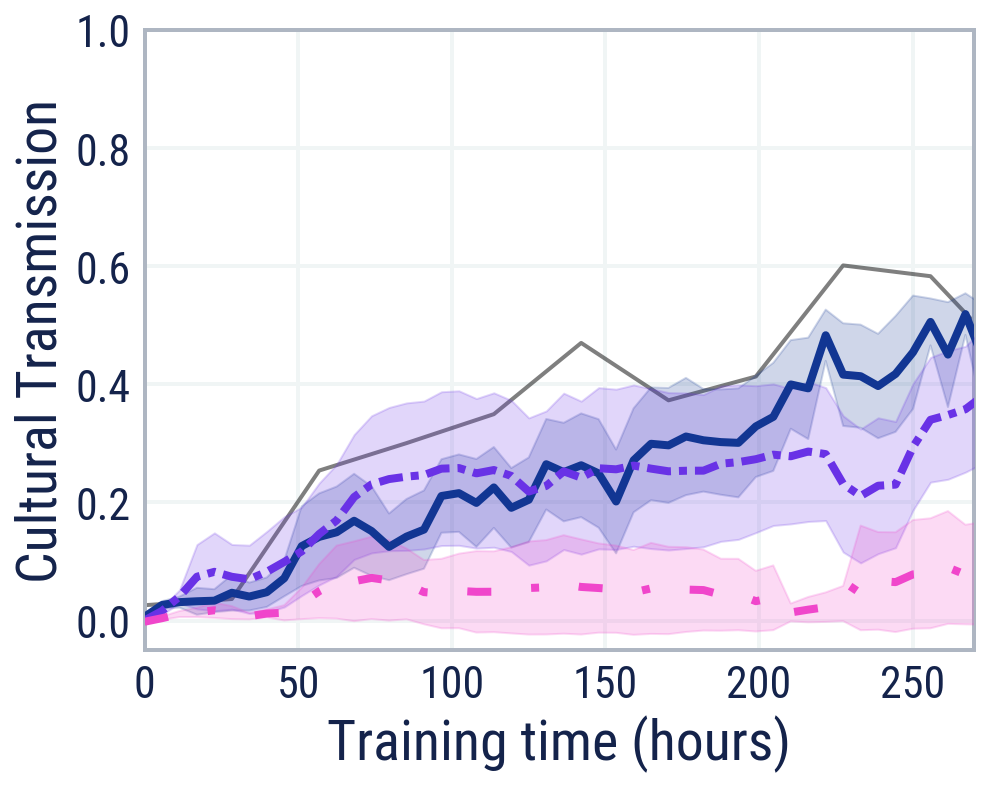}
        \label{subfig:ablate-adr-c}
    \end{subfigure}
    \hfill
    \begin{subfigure}[b]{0.15\textwidth}
        \centering
        \includegraphics[width=\textwidth]{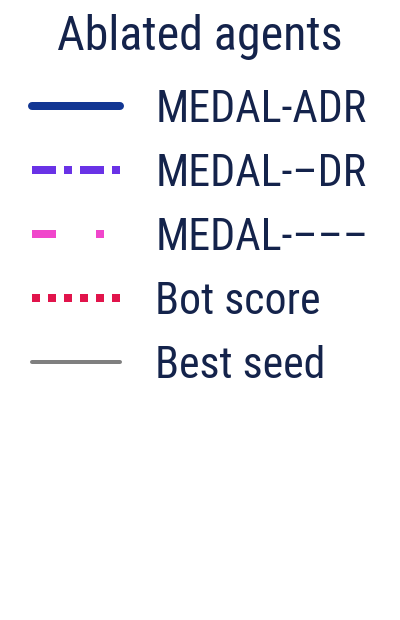}
    \end{subfigure}
    \caption{Isolating the automatic (A) and domain randomisation (DR) components, we show that our MEDAL-ADR agent outperforms the ablated MEDAL-–DR and MEDAL-––– agents. The evaluation CT metric is computed in complex world $5$-goal probe tasks (right).  MEDAL-–DR is competitive only on some tasks, while MEDAL-––– hardly learns. Thanks to its adaptive curriculum, MEDAL-ADR achieves the best score and the highest CT.
    }
    \label{fig:ablate-adr}
\end{figure}

For all our ablations, we plot the score achieved throughout training along with the training and evaluation cultural transmission metrics. We report the mean performance for each of these measures across $10$ initialisation seeds for agent parameters and task procedural generation. The shaded area on the graphs represents one standard deviation. We also report the expert's score and the best seed for scale and upper bound comparisons. Figures \ref{fig:ablate-0-agents}, \ref{fig:ablate-dropout} and \ref{fig:ablate-adr} show the results.

\clearpage

\section{Analysis}
\label{sec:evaluation}

In this section, we analyse our agent. There are three key strands of novelty compared with previous work: strong within-episode recall, generalisation over a diverse task space and rigorous measurement of high-fidelity knowledge transfer. To understand how our agent is capable of this, we ``neuroimage'' the memory activations, finding clearly interpretable individual neurons. We use held-out task seeds in all of our analyses. For full details, see Appendix \ref{app:analysis}.

\subsection{Recall}
To assess the recall capabilities of our agents, we quantify their performance across a set of tasks where the expert drops out. The intuition here is that if our agent is able to recall information well, then its score will remain high for many timesteps even after the expert has dropped out. However, if the agent is simply following the expert or has poor recall, then its score will instead drop greatly.

For each task, we evaluate the score of the agent across a number of contiguous $900$-step trials, comprising an episode of experience for the agent. In the first trial, the expert is present alongside the agent, and thus the agent can infer the optimal path from the expert. However, from the next trial onwards the expert is dropped out, so the agent must continue to solve the task alone. The world, agent, and game are not reset between trial boundaries; we use the term ``trial'' to refer to the bucketing of score accumulated by each player within the time window. We consider recall from two different experts, a scripted bot and a human player. For both, we use the worlds from the 4-goal probe tasks introduced in Section~\ref{sec:probe_tasks}.

\begin{figure}[htb]
    \centering
    \begin{subfigure}{0.48\textwidth}
         \includegraphics[width=\textwidth]{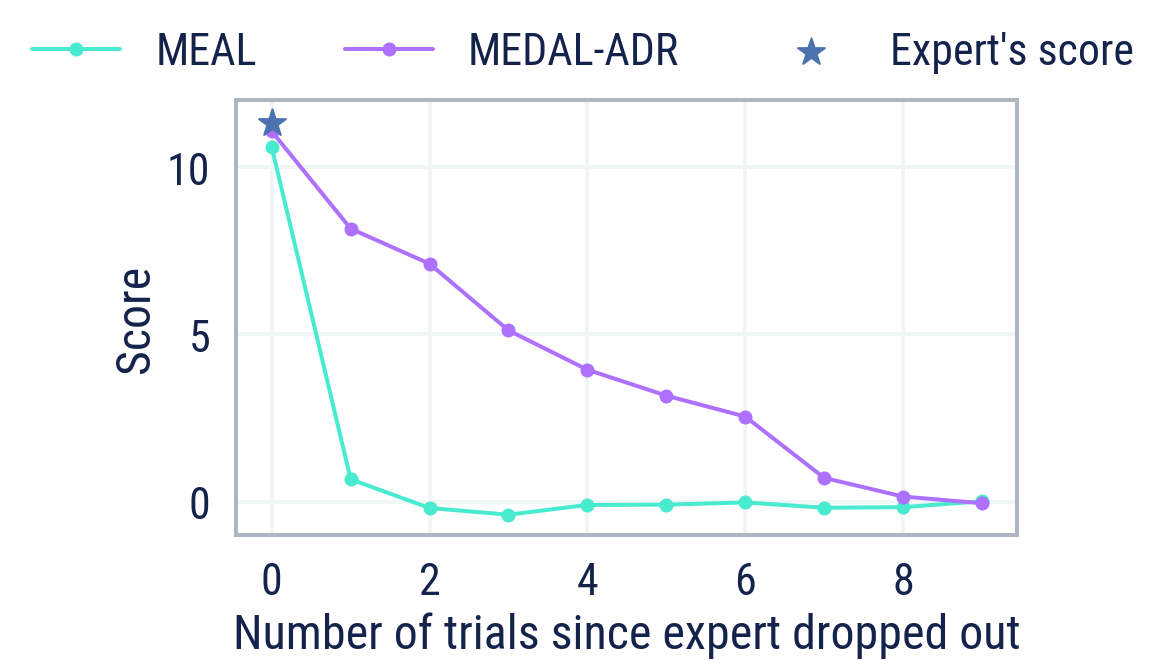}
        \label{fig:recall_analysis_bots}
    \end{subfigure}
    \hfill
    \begin{subfigure}{0.48\textwidth}
         \includegraphics[width=\textwidth]{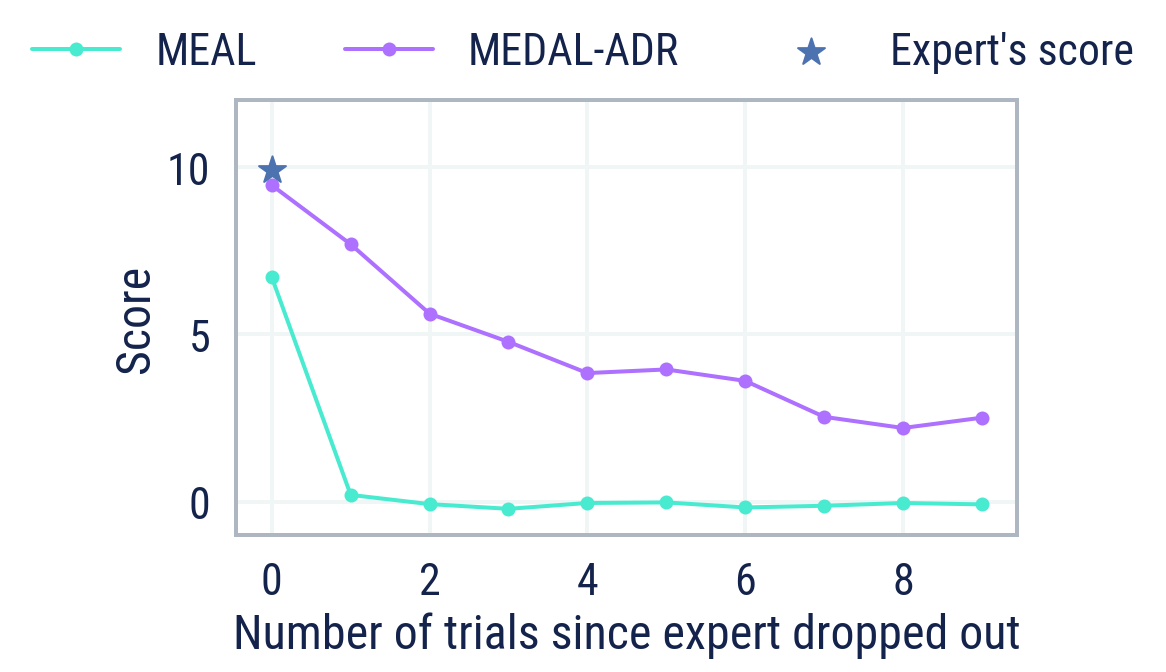}
        \label{fig:recall_analyis_humans}
    \end{subfigure}    
    \caption{Score of MEDAL-ADR and ME–AL agents across trials since the expert dropped out. (left) The experts are scripted bots. (right) The experts are human trajectories. MEDAL-ADR is able to recall significantly better than ME-AL. An accompanying \href{https://sites.google.com/view/dm-cgi\#h.gjdvzr76oj4x}{video} shows MEDAL-ADR's memorisation in a $3600$-step ($4$ trial) episode.}
    \label{fig:recall_analysis}
\end{figure}

Figure~\ref{fig:recall_analysis} compares the recall abilities of our agent trained with expert dropout (MEDAL-ADR) and without (ME–AL, closely related to the prior state of the art in \cite{ndousse2020learning}). Notably, after the expert has dropped out, we see that our MEDAL-ADR agent is able to continue solving the task for the first trial while the ablated ME–AL agent cannot. MEDAL-ADR maintains a good performance for several trials after the expert has dropped out, despite the fact that the agent only experienced $1800$-step episodes during training.

\subsection{Generalisation}
To assess the generalisation capabilities of our agents, we quantify their performance over a distribution of procedurally-generated tasks. We decompose our analysis of generalisation into world space, game space, and expert space. We analyse both ``in-distribution'' and ``out-of-distribution'' generalisation, with respect to the distribution of parameters seen in training (see Table~\ref{tab:adr-parameters-id-ood}). Out-of-distribution values are calculated as $\pm$20\% of the min/max in-distribution ADR values where possible, and indicated by cross-hatched bars in all figures. 

In every task, an expert bot is present for the first $900$ steps, and is dropped out for the remaining $900$ steps. We define the \textit{normalised score} as the agent's score in $1800$ steps divided by the expert's score in $900$ steps. An agent which can perfectly follow but cannot remember will score $1$. An agent which can perfectly follow and can perfectly remember will score $2$. Values in between correspond to increasing levels of cultural transmission.

\paragraph{World space} The space of worlds is parameterised by the size and bumpiness of the terrain (terrain complexity) and the density of obstacles (obstacle complexity). To quantify generalisation over this space, we generate tasks with worlds from the Cartesian product of these complexities, with zero-noise expert bots, and with games uniformly sampled across the possible number of crossings for 5 goals. 

\begin{figure}[htb]
    \centering
    \begin{subfigure}{0.48\textwidth}
         \includegraphics[width=\textwidth]{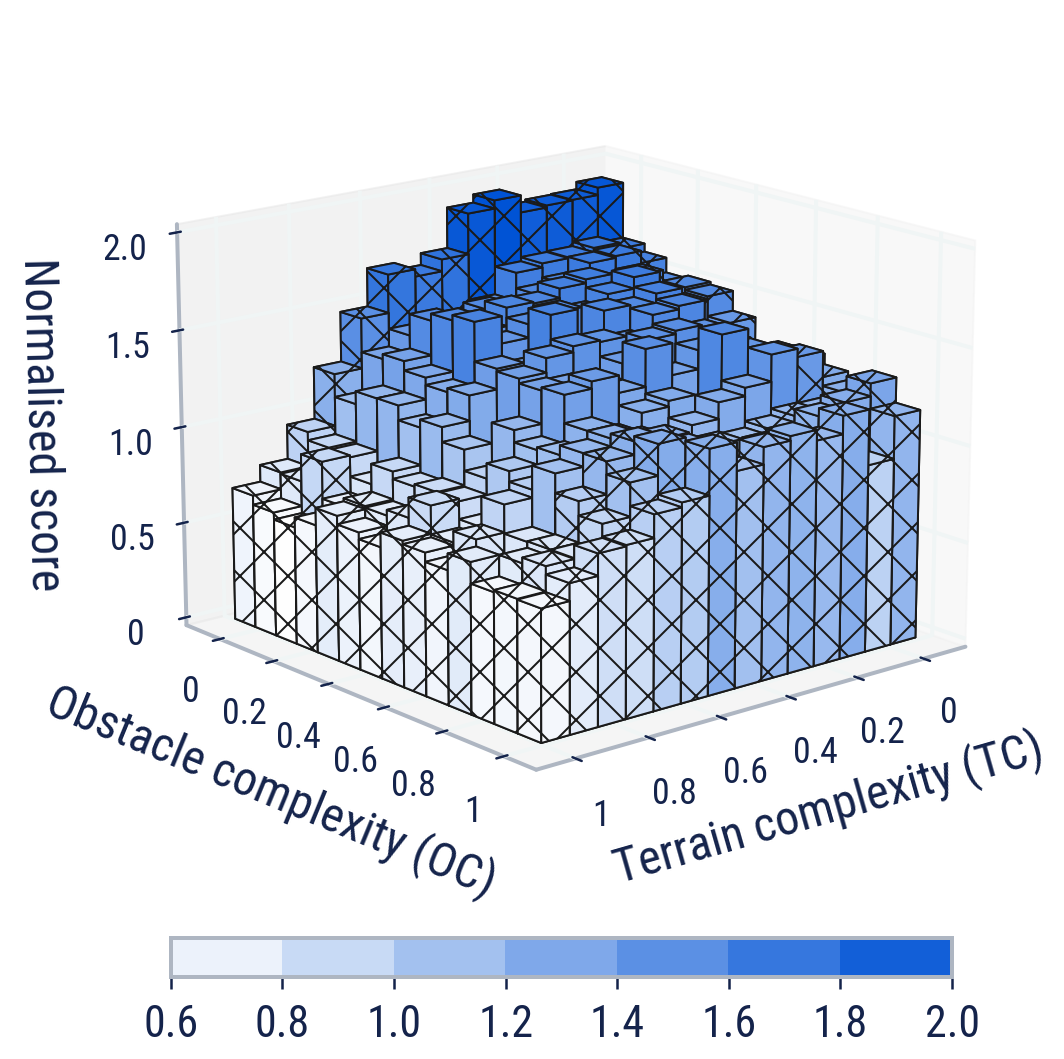}
        \caption{\centering Normalised score across complexities.}
        \label{fig:generalization-analysis-world-space-score}
    \end{subfigure}
    \hfill
    \begin{subfigure}{0.48\textwidth}
         \includegraphics[width=\textwidth]{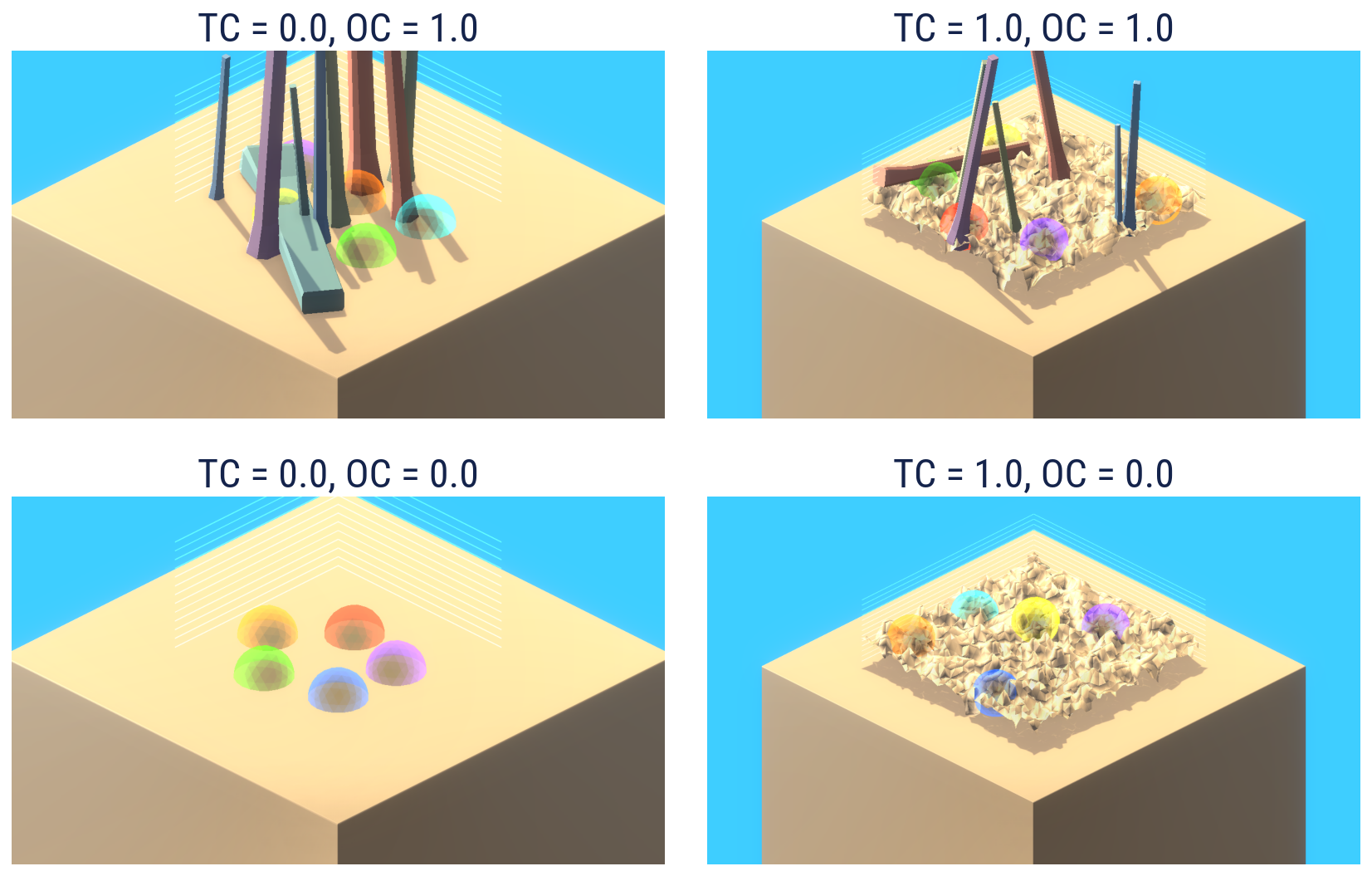}
        \caption{\centering Example worlds sampled from the extremal values of the distribution parameters.}
        \label{fig:generalization-analysis-world-space-example}
    \end{subfigure} 
    \caption{MEDAL-ADR generalises across world space parameters, demonstrating both following and remembering across much of the space. Obstacle complexity and terrain complexity are defined in Appendix \ref{app:analysis}. Hatched cells denote out-of-distribution values. Sample videos for the extremal values can be found on the  \href{https://sites.google.com/view/dm-cgi\#h.ms6jr5zahrco}{website}.}
    \label{fig:generalization-analysis-world-space}
\end{figure}

\begin{figure}[htb]
    \centering
     \includegraphics[scale=0.35]{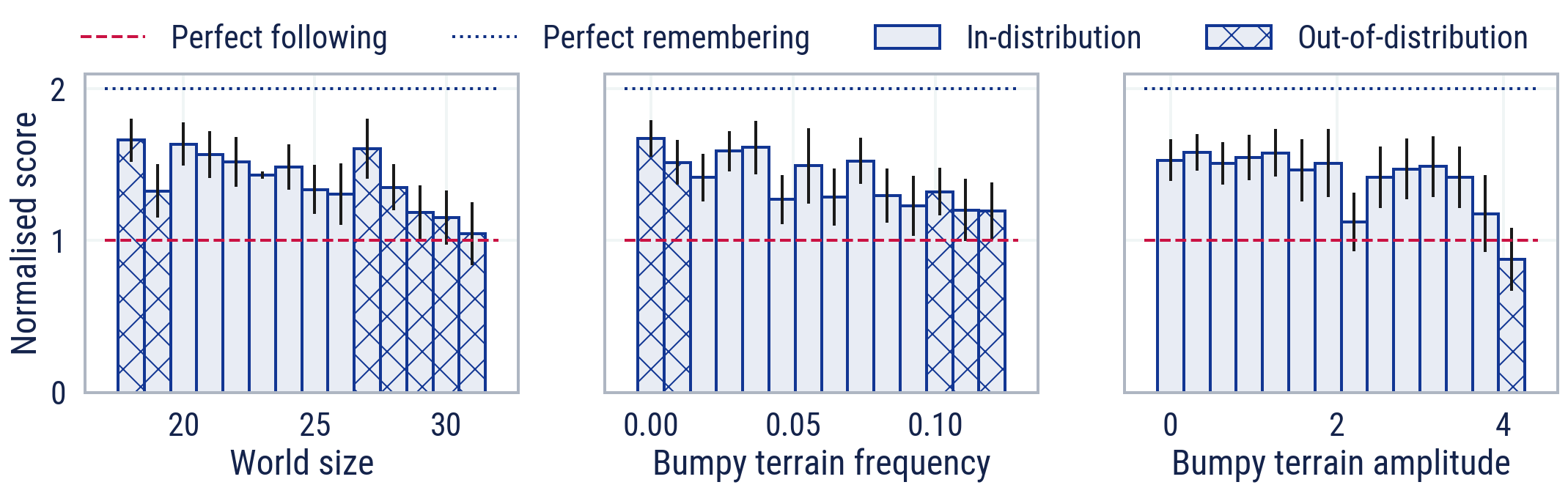}
     \includegraphics[scale=0.35]{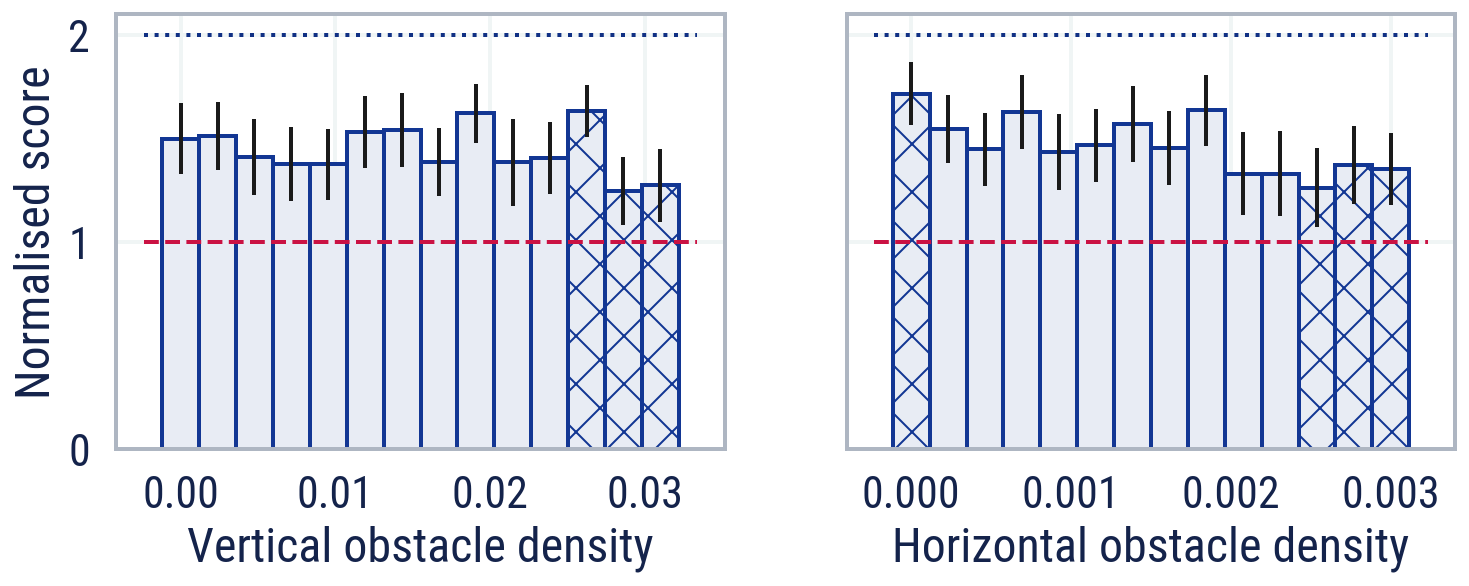}     
    \caption{A slice through the world space allows us to disentangle MEDAL-ADR's generalisation capability across different world space parameters. For the top parameters (terrain complexity), larger values lead to a gradual reduction in performance. For the bottom parameters (obstacle complexity), performance is stable until they reach their larger out-of-distribution values.}
    \label{fig:generalization-analysis-world-space-individual}
\end{figure}

From Figures \ref{fig:generalization-analysis-world-space} and \ref{fig:generalization-analysis-world-space-individual}, we conclude that MEDAL-ADR generalises well across the space of worlds, demonstrating both following and remembering across a majority of the space, including when the world is out-of-distribution. As both obstacle and terrain complexity increases, the performance of our agent drops. For example, as both complexities exceed 0.8 (and are therefore out-of-distribution), the agent's score drops below 1: while it can follow near-perfectly, it is unable to remember. Score gradually decreases as each of the world's individual parameters increases. This is most notable with the world size parameter. Comparing to Figure \ref{fig:adr-full-training}, this is unsurprising: it is likely that a combination of more training, a longer recurrent unroll, and a larger memory may be required to generate an agent capable of cultural transmission in larger worlds, where the number of timesteps for a complete goal cycle demonstration is larger.

\paragraph{Game space} The space of games is defined by the number of goals in the world as well as the number of crossings contained in the correct navigation path between them. To quantify generalisation over this space, we generate tasks across the range of feasible ``$N$-goal $M$-crossing'' games, each with a zero-noise expert bot in a flat empty world of size 23.

\begin{figure}[htb]
    \centering
    \includegraphics[width=0.8\textwidth]{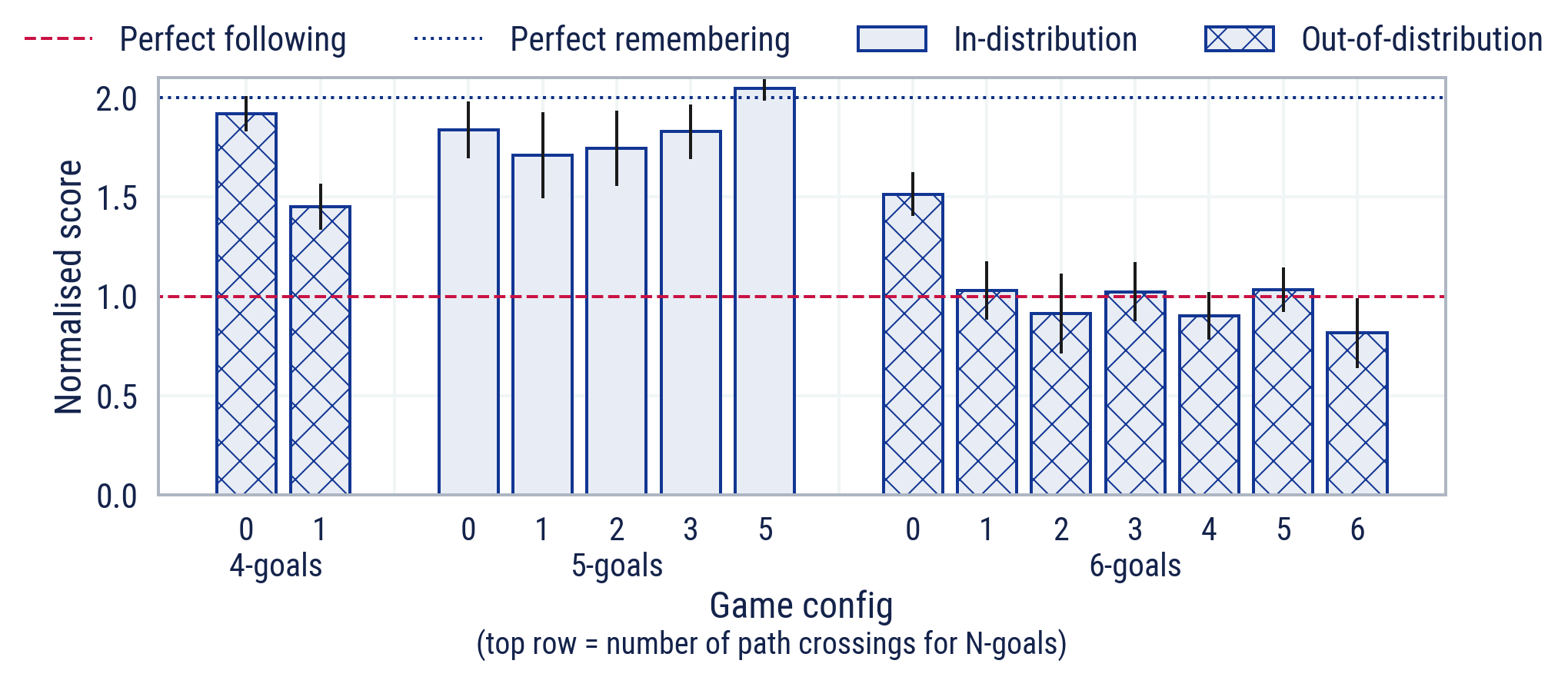}
    \caption{MEDAL-ADR generalises across games, demonstrating remembering capability both inside and outside the training distribution. Hatched bars denote out-of-distribution values. This can be seen in the \href{https://sites.google.com/view/dm-cgi\#h.xfaxoqy5f3ls}{videos} of generalisation over the game space.}
    \label{fig:generalization-analysis-task-space}
\end{figure}

Figure~\ref{fig:generalization-analysis-task-space} shows our agent's ability to generalise across games, including those outside of its training distribution. Notably, MEDAL-ADR can perfectly remember all numbers of crossings for the in-distribution 5-goal game. We also see impressive out-of-distribution generalisation, with our agent exhibiting strong remembering, both in $4$-goal and $0$-crossing $6$-goal games. Even in complex $6$-goal games with many crossings, our agent can still perfectly follow. Clearly we require more randomisation to get good generalisation of memorisation to higher numbers of goals.

\paragraph{Expert space} The space of experts is defined by the speed, dropout probability, and action distribution taken by the expert in the world. To quantify generalisation over this space, we generate tasks with expert bots from the Cartesian product of movement speed and dropout probability, flat empty worlds of size 23, and games uniformly sampled across the possible number of crossings for 5-goals.

\begin{figure}[htb]
    \centering
    \begin{subfigure}{0.48\textwidth}
        \includegraphics[width=\textwidth]{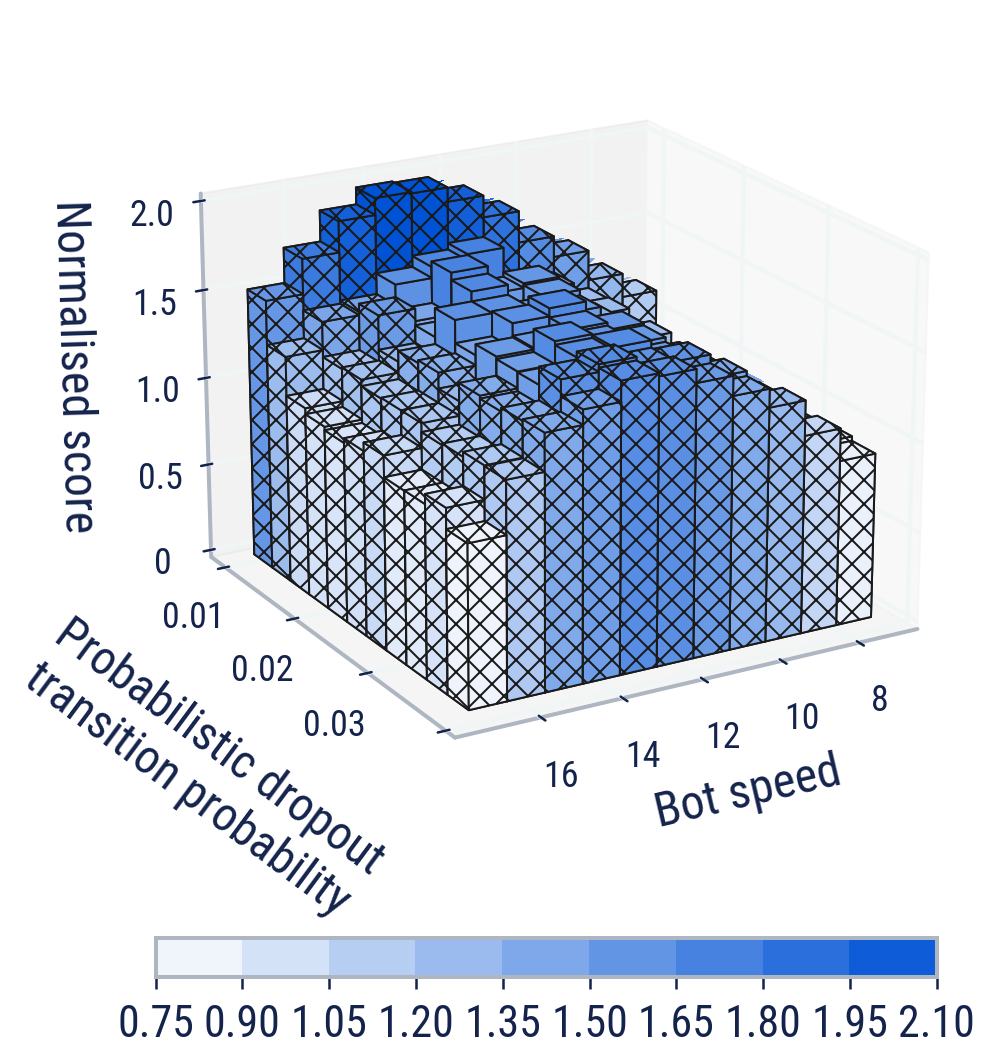}
        \label{fig:generalization-analysis-expert-space-3d}
    \end{subfigure}
    \hfill
    \begin{subfigure}{0.48\textwidth}
         \includegraphics[width=\textwidth]{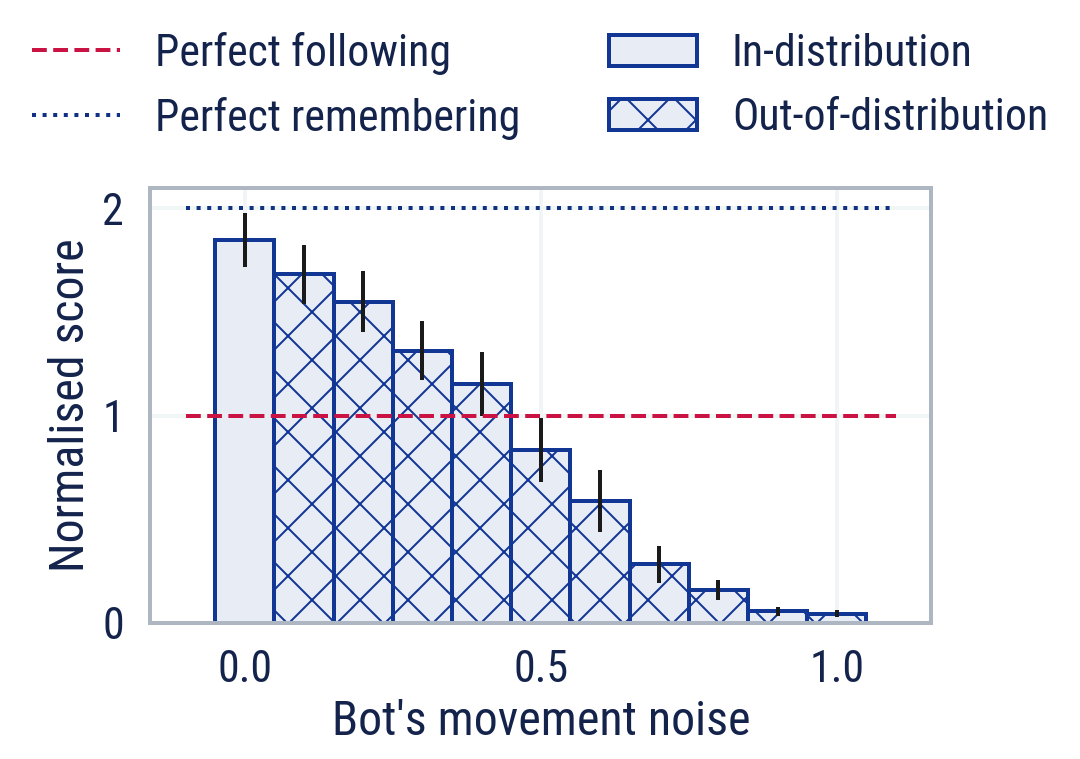}
        \label{fig:generalization-analysis-expert-space-noise}
    \end{subfigure} 
    \caption{Influence of variation in expert's parameters on MEDAL-ADR's performance. MEDAL-ADR generalises when paired with a variety of expert co-players (left), and is robust to significant levels of out-of-distribution noise in their movement (right). Normalisation in both plots is done with respect to the non-noisy expert bot with a speed of 13, for interpretability independent of the quality of the expert. In all episodes the bot drops out fully after $900$ steps. Videos of these episodes can be found on the \href{https://sites.google.com/view/dm-cgi\#h.ltw8sokd3hnk}{microsite}.}
    \label{fig:generalization-analysis-expert-space}
\end{figure}

Figure~\ref{fig:generalization-analysis-expert-space} demonstrates that MEDAL-ADR generalises well across a range of speeds and dropout probabilities, with consistent remembering of demonstrated trajectories. Furthermore, we can see that our agent is also robust to the introduction of noise in the expert's movement, exhibiting the ability to divine and recall the optimal path until the noise exceeds 30\% and succeeding at the task before dropout by following even with 50\% noise (despite never experiencing expert noise in training). Above 50\% noise, the expert's policy becomes degenerate and therefore unusable for solving the task.

\subsection{Fidelity}
\label{sec:fidelity}

\cite{Dawson1965ObservationalLI} introduced the ``two-action task'' as a means by which to study imitation of behaviours. In this experimental set-up, subjects are required to solve a task with two alternative solutions. Half of the subjects observe a demonstration of one solution, the others observe a demonstration of the other solution. If subjects disproportionately use the observed solution, this is evidence that supports imitation. This experimental approach is widely used in the field of social learning; we use it as a behavioural analysis tool for artificial agents for the first time, to our knowledge. Using the tasks from our game space analysis, we record the preference of the agent in pairs of episodes where the expert demonstrates the optimal cycles $\sigma$ and $\sigma^{-1}$. The preference is computed as the percentage of correct complete cycles that an agent completes which match the direction of the expert cycle. Evaluating this over 1000 trials, we find that the agent's preference matched the demonstrated option $100\%$ of the time, i.e. in every completed cycle of every one of the $1000$ trials.

\begin{figure}[htb]
    \centering
    \begin{subfigure}{\textwidth}
        \includegraphics[width=\textwidth]{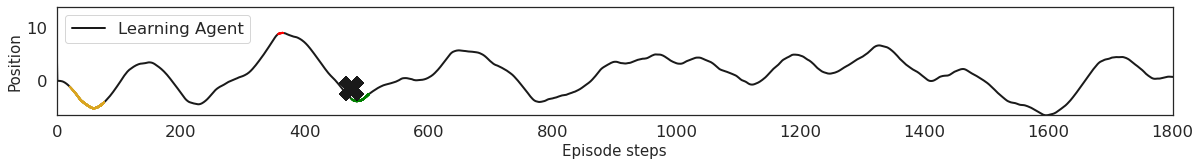}
        \caption{\centering The bot is absent for the whole episode.}
        \vspace{1cm}
        \label{fig:no_bot_trajectory}
    \end{subfigure}
    \begin{subfigure}{\textwidth}
        \includegraphics[width=\textwidth]{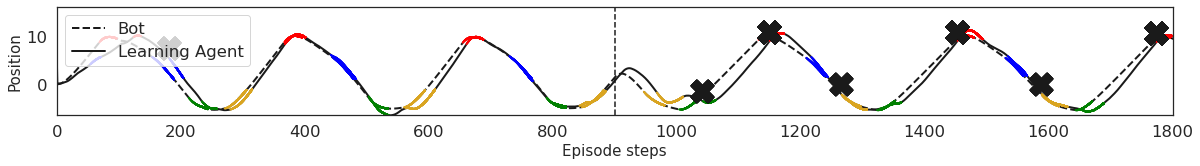}
        \caption{\centering The bot shows a correct trajectory for $900$ steps followed by an incorrect trajectory.}
        \vspace{1cm}
        \label{fig:wrong_bot_trajectory}
    \end{subfigure}
    \begin{subfigure}{\textwidth}
        \includegraphics[width=\textwidth]{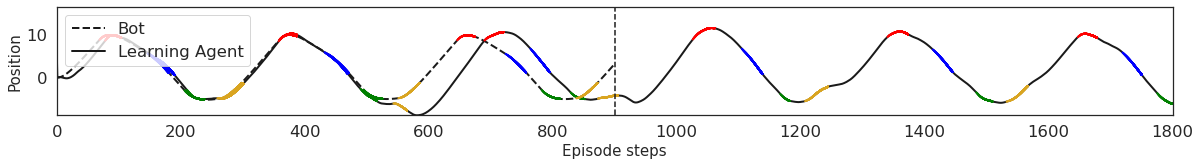}
        \caption{\centering The bot shows a correct trajectory in the first half of the episode and then drops out.}
        \vspace{1cm}
        \label{fig:dropout_trajectory}
    \end{subfigure}
    \begin{subfigure}{\textwidth}
        \includegraphics[width=\textwidth]{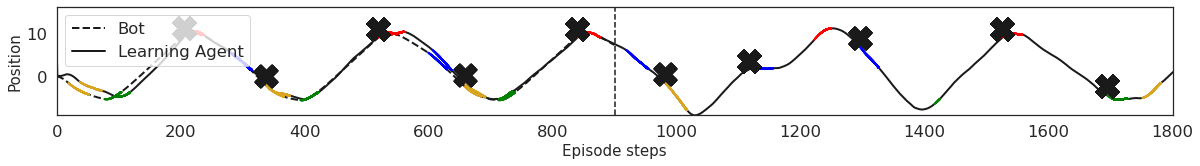}
        \caption{\centering The bot shows an incorrect trajectory in the first half of the episode and then drops out.}
        \label{fig:wrong_demo_trajectory}
    \end{subfigure}
    \caption{Trajectory plots for MEDAL-ADR agent for a single episode. The coloured parts of the lines correspond to the colour of the goal sphere the agent and expert have entered and the $\times$s correspond to when the agent entered the incorrect goal. Here, position refers to the agent's position along the z-axis. Four corresponding \href{https://sites.google.com/view/dm-cgi\#h.mbxtqto8j9b2}{videos} are available.}
    \label{fig:trajectory_plots}
\end{figure}

Trajectory plots reveal the correlation between expert and agent behaviour (see Figure \ref{fig:trajectory_plots}). By comparing trajectories under different conditions, we can argue that cultural transmission of information from expert to agent is causal. The agent cannot solve the task when the bot is not placed in the environment (Figure~\ref{fig:no_bot_trajectory}). When the bot is placed in the environment, the agent is able to successfully reach each goal (Figure \ref{fig:wrong_bot_trajectory}). Even when the bot begins performing incorrectly halfway through the episode, the agent continues to follow (Figure \ref{fig:wrong_bot_trajectory}). However, the agent is not dependent on the bot's continual presence. When the bot drops out, the agent continues to execute the demonstrated trajectory, whether right (Figure \ref{fig:dropout_trajectory}) or wrong (Figure \ref{fig:wrong_demo_trajectory}). 

\subsection{Neural activations}

Here, we ``neuroimage'' our agents to gain insights into the specific information represented by neurons in the belief state representation. We find two types of neurons with specialised roles: one, dubbed the \textit{social neuron}, encapsulates the notion of agency; the other, called a \textit{goal neuron}, captures the periodicity of the task. 

\begin{figure}[tbp]
    \centering
    \begin{subfigure}[b]{0.47\textwidth}
    \centering
    \includegraphics[width=\textwidth]{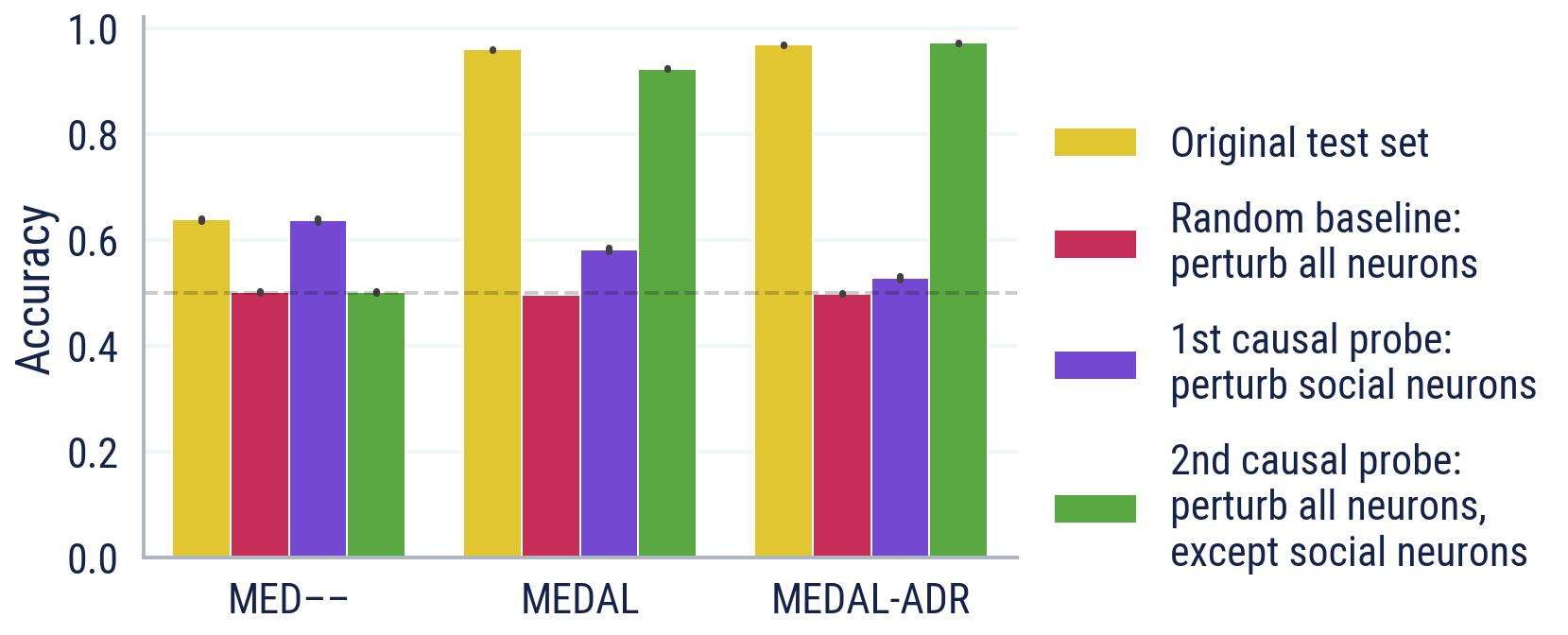}
    \end{subfigure}
    \begin{subfigure}[b]{0.47\textwidth}
        \centering
        \includegraphics[width=\textwidth]{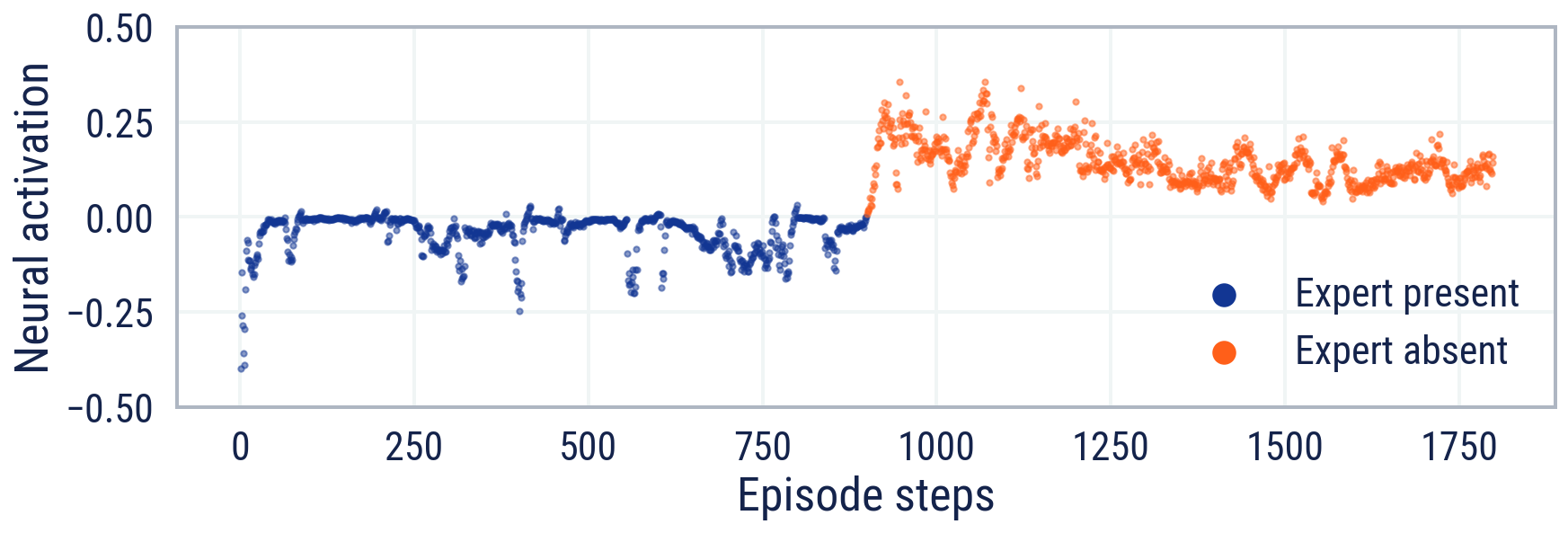}
        \label{subfig:medal-adr-activations}
    \end{subfigure}
    \caption{(left) We report the accuracy of three models for predicting the expert's presence based on the belief states of three trained agents (MED––, MEDAL, and MEDAL-ADR). We make two causal interventions (in green and purple) and a control check (in red) on the original test set (yellow). The small error bars, drawn from 10 different initialisation seeds, suggest a widespread consensus across the 10 runs on which neurons are social and how impactful they are on task performance. (right) Activations for one of MEDAL-ADR's social neurons.}
\label{fig:social-neuron-accuracy}
\end{figure}

A social neuron crisply encodes the presence or absence of the expert in the world. We detect such neurons using a linear probing method \citep{alain2016understanding, anand2019unsupervised} described in Appendix \ref{app:bonus-neural-activations}. This stark difference between the three agents probed in Figure \ref{fig:social-neuron-accuracy} suggests that the attention loss is (at least partly) responsible for incentivising the construction of ``socially-aware'' representations. Qualitatively, Figure \ref{fig:social-neuron-activations} shows that the activations of the maximally weighted neurons have meaningfully different magnitudes and opposing signs depending on whether the expert is present or not.

\begin{figure}[htb]
    \centering
    \begin{subfigure}[b]{0.49\textwidth}
        \includegraphics[width=\textwidth]{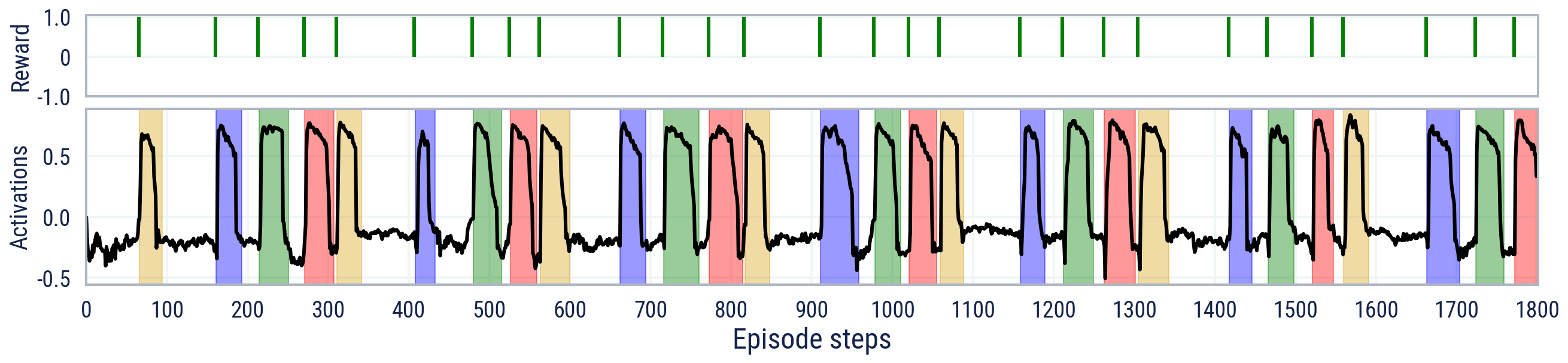}
        \caption{\centering Perfect bot for the entire episode.}
        \label{subfig:cyclic-neuron-full}
    \end{subfigure}
    \hfill
    \begin{subfigure}[b]{0.49\textwidth}
        \centering
        \includegraphics[width=\textwidth]{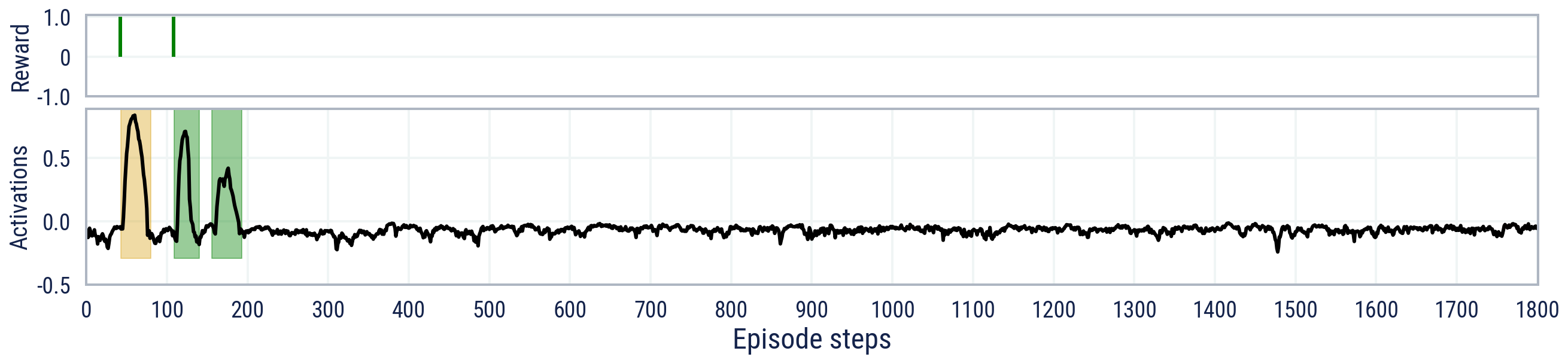}
        \caption{\centering No bot for the entire episode.}
        \label{subfig:cyclic-neuron-no-bot}
    \end{subfigure}
    
    \vspace{2mm}
    
    \begin{subfigure}[b]{0.49\textwidth}
        \centering
        \includegraphics[width=\textwidth]{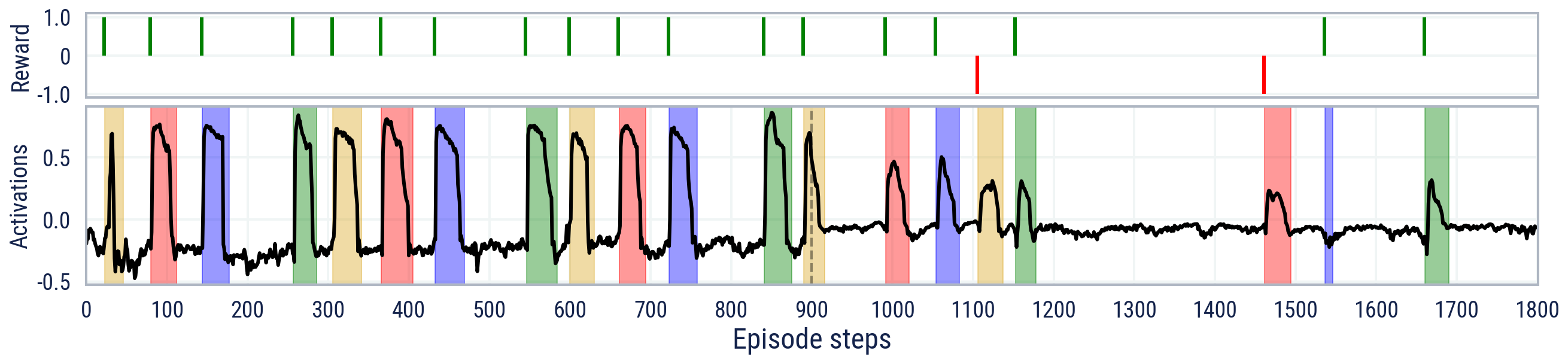}
        \caption{\centering Perfect bot drops out after half of the episode.}
        \label{subfig:cyclic-neuron-half}
    \end{subfigure}
    \hfill
    \begin{subfigure}[b]{0.49\textwidth}
        \centering
        \includegraphics[width=\textwidth]{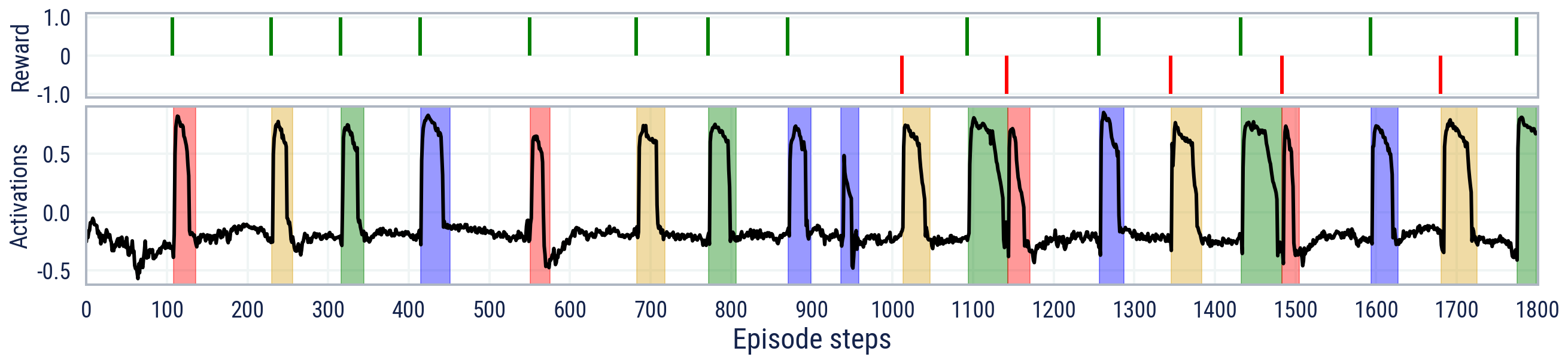}
        \caption{\centering Noisy bot for the entire episode.}
        \label{subfig:cyclic-neuron-noisy}
    \end{subfigure}
    \caption{Spikes in the goal neuron's activations correlate with the time the agent remains inside a goal (illustrated by coloured shading), irrespective of the reward received.}
\label{fig:cyclic-neuron-with-sphere-shading}
\end{figure}

The activation of a goal neuron is highly correlated with the entry of an agent into a goal sphere. We identified a goal neuron by inspecting the variance of belief state neural activations across an episode. Figure \ref{fig:cyclic-neuron-with-sphere-shading} shows that this neuron fires when the agent enters and remains within a goal sphere. For further empirical observations about the goal neuron, see Appendix \ref{app:bonus-attention-loss}.

\clearpage
\section{Related work}
\label{sec:related_work}

Our work continues a long line of research about how to generate artificial agents that imitate human behaviour online in rich 3D physical environments, most prominently in the robotics community (see \cite{BREAZEAL2002481} for a review). Much of this work has employed model-based systems to make online learning sample efficient (e.g. \cite{614389}), sometimes taking inspiration from prediction mechanisms in the human brain (e.g. \cite{doi:10.1080/09540090310001655129, 6789045}). Inspired by this, we ask whether it is possible to use model-free RL to learn an implicit model in the memory of a neural network, which at test-time is capable of online imitation. The answer is ``yes''. Our approach inherits the scaleability advantages of model-free methods \citep{DBLP:journals/corr/abs-1806-02308}, but retains the sample-efficiency of model-based methods at test time. Because our method makes heavy use of domain randomisation, it may even be amenable to use in sim-to-real contexts \citep{DBLP:journals/corr/abs-2010-11251}. 

Previous works have used model-free reinforcement learning to generate a fixed policy capable of test-time social learning and generalisation to held out tasks. \cite{borsa2017observational} demonstrated that A3C agents \citep{mnih2016asynchronous} were capable of learning to find and follow an expert co-player in a grid-world navigation task, purely on the basis of goal-location reward. Agents trained on a four-room maze generalised in zero-shot to follow in a nine-room maze. When the presence of the expert was dropped out across training, the agent learned to solve the nine-room maze fully independently. However, the authors did not produce a fixed policy capable of reproducing a demonstrated trajectory after expert dropout within a single evaluation episode.

\cite{ndousse2020learning} extend the work of \cite{borsa2017observational}, demonstrating more rigorously that learning social learning improves generalisation to held-out tasks and achieves better performance than solo baselines on hard exploration problems. Crucial to the success of their agents is a next-step prediction auxiliary loss \citep{shelhamer2016loss, jaderberg2016reinforcement}. The authors introduce the \textit{GoalCycle} environment where task knowledge comprises navigating goals in the correct order, in principle providing parameterised, open-ended navigational complexity. Similarly to previous work, \cite{ndousse2020learning} do not generate an agent capable of within-episode recall. 

There have been various works leveraging reinforcement learning to generalise to human co-play in coordination tasks, without using human data in the training pipeline. FCP \citep{strouse2021fcp} uses a population of self-play agents and their past checkpoints as a training distribution for a focal agent in a grid-world version of Overcooked \citep{carroll2019utility}, producing agents that generalise well to human co-play on the same task seen in training. LILA \citep{woodward2020learning} uses multi-agent deep RL to generate an agent capable of inferring a ``principal'' agent's goal, namely the preference for one of two object types, and using this information to assist the principal, achieve good performance with human co-players in simple grid-worlds. We take inspiration from these works, but tackle a different problem: high fidelity cultural transmission of long, strategic behaviours.

A line of work in the card game Hanabi \citep{bard2020hanabi} is also oriented in this direction, generating zero-shot coordination with human players via various explicit modelling techniques, including by exploiting symmetries \citep{hu2020other, bullard2021quasi}, learned belief search \citep{hu2021learned}, diversity algorithms \citep{canaan2020generating, lupu2021trajectory} and modelling theory of mind \citep{zhu2021few}. Hanabi is a turn-based game in which all actions are visible and each player has different legal actions, so it is not an appropriate arena in which to study the correspondence problem. 

We identify a minimal sufficient ``starter kit'' of ingredients (MEDAL-ADR) that give rise to a fixed neural network capable of cultural transmission. Many of these ingredients build off prior work. The importance of memory (M) was already identified via ablations in \citep{borsa2017observational}, and echoes the presence of recurrent state in meta-RL algorithms \cite{wang2016learning, duan2016rl}. Theoretically, our setting is automatically a POMDP, since the policy of the expert and the details of the task sampling are hidden information. Therefore, we would expect memory to be essential in de-aliasing states via observation of task dynamics and expert behaviour over time \citep{wierstra2007solving}. The presence of expert co-players (E) was already ablated in \citep{borsa2017observational}; we provide additional ablations that explore how the reliability of the expert affects the learning of cultural transmission, inspired by the competence dimension of interpersonal perception \citep{fiske2002model}. 

Dropout (D) is a well studied mechanism for reducing overfitting in many fields of machine learning \citep{labach2019survey}, and was used across training episodes in \cite{borsa2017observational, ndousse2020learning}. To our knowledge, we are the first to use co-player dropout \textit{within} the course of an episode. 

Unsupervised auxiliary losses in reinforcement learning were introduced in \cite{jaderberg2016reinforcement}, and are commonly used to shape representations, promoting more efficient learning, and helping to overcome hard exploration problems. \cite{ndousse2020learning} found that a next-step prediction loss was beneficial to promote social learning in a grid-world setting. Reconstruction losses in particular tend to have beneficial effects on generalisation \citep{le2018supervised}. We use a novel unsupervised auxiliary reconstruction loss focused on the relative position of co-players, which we refer to as an attention loss (AL).  

Automatic domain randomisation (ADR) is a technique within the broader field of co-adaptation of agents and environments to yield generalisation. Its precursor, domain randomisation (DR), used uniform sampling of a diverse set of environments and tasks to bridge the sim-to-real gap in robotics~\cite{peng2017domainrandomization, tobin2017domainrandomization}. ADR extended DR by automatically adapting the randomisation ranges to agent performance, leading to the zero-shot transfer of a policy for a complex, in-hand manipulation task from simulation to a real-world robot~\cite{openai2019}. Other forms of agent-environment co-adaptation include \cite{everett2019optimising, wang2019paired, DBLP:journals/corr/abs-2110-02439}. In the multi-agent setting, agent-agent interactions provide a rich source of autocurricula \citep{DBLP:journals/corr/abs-2012-02096, leibo2019autocurricula, DBLP:journals/corr/abs-1909-07528, oelt2021open} in which co-operation and competition can scale the task difficulty automatically as agents become more capable.

All of these ingredients have analogues in human and animal cognition. Better working memory (M) is known to correlate with improved fluid intelligence in humans \citep{duncan2012task} and the ability to solve novel reasoning problems \citep{cattell1963theory}, including by imitation \citep{vostroknutov2018role}. The quality of expert demonstrations (E) is a crucial determinant of the success of cultural transmission in humans, to the extent that humans possess psychological adaptations to enhance quality when multiple models are available, such as prestige bias \citep{henrich2001evolution}. 

The progressive increase in duration of expert dropout (D) mirrors the development of secure attachment in attachment theory \citep{bowlby1958nature, harlow1958nature}, where a human child learns to use their caregiver as a safe base for independent behaviour when that caregiver is absent. The success of expert dropout as a means of learning to recall demonstrations across contexts echoes the discovery that interleaving improves inductive learning \citep{kornell2008learning}. The importance of dropout for successful imitation has also been noted in animals \citep{galef1988imitation}: ``for the pattern of behaviour initiated by the leader to become part of the behavioural repertoire of the follower, independent of the leader, the pattern of behaviour must come under the control of stimuli not dependent on the presence of the leader''. Humans, along with many animals, have an in-built attentional bias towards biological motion \citep{bardi2011biological, bardi2014first}, mirroring our attention loss (AL). 

Among animals, social learning is known to be preferred over asocial learning in uncertain or varying ecological contexts \citep{aplin2016understanding, smolla2016copy}, environment properties we create via domain randomisation (DR). Learning cultural transmission, then, requires that the environment remains consistently in a ``Goldilocks zone'' of variability, as observed in autocatalytic models \citep{boyd1988culture} and empirical data on climate across human evolution \citep{richerson2005evolution, zachos2001trends}. We achieve this balance by varying the distributional parameters for domain randomisation automatically (A) as a function of the current cultural transmission capability of the agent, keeping the agent in the zone of proximal development \citep{vygotsky1980mind}.

There has been much laboratory work studying cultural transmission in humans, including in situations where humans learn a new behaviour and transmit this behaviour across generations (e.g. \cite{DBLP:journals/corr/abs-2107-13377, saldana2019cumulative, kirby2008cumulative}). Modelling in such settings is typically challenging, particularly when it comes to the intricate sensorimotor structure of behaviour which characterises much human tool use \citep{guerin2013}. On the other hand, there is a long history of interaction between cultural evolution and computational systems. In many such systems, the inheritance step is hard-coded directly (e.g. \citep{gabora1995, rendell2010copy, hart2022artificial, bredeche2022social}), pre-defining the abstraction of the behaviours that can be represented, and potentially limiting the open-endedness of the method. \cite{winfield2022experiments} propose a more ``data-driven'' inheritance system, in which imitation is carried out in third-person by an explicit online learning algorithm. Our work takes this one step further: the ability of cultural transmission is itself learned in a rich sensorimotor environment. Therefore, our methods head towards an even more general and scaleable approach to generating cultural transmission in artificial agents, and perhaps even offers inspiration for modelling; see Section \ref{sec:future_work}.

In RL algorithms, it is common to explore in the space of actions, such as in $\epsilon$-greedy and softmax policies, or when entropy regularisation is used. To overcome local optima in hard exploration problems, authors have proposed exploring in the space of policies via diversity objectives \citep{hong2018diversity}, in the space of states via intrinsic motivation \citep{vamshicuriosity, bellemare2016unifying, burda2018exploration}, in the space of value functions \citep{osband2019deep,badia2020up} and in the space of neural network weights \citep{plappert2018parameter,fortunato2019noisy}. Less common is work on exploration in the space of structured behaviours pertinent to the task at hand, for instance in hierarchical RL \citep{steccanella2020hierarchical}, the kinds of behaviours that humans naturally exchange during cultural transmission. This work helps to fill the gap, solving a hard exploration problem via within-episode cultural transmission, automatically amenable to human interaction. Since this capability is learned, it can be seen as novel form of ``meta-exploration'' \citep{gupta2018meta, liu2021decoupling}. 

We may characterise our setting as a form of meta-learning problem: learning to learn from other agents. The trained network must be capable of online adaptation \citep{Bottou98onlinelearning} with fixed weights, behaving in a way that is rational in hindsight \citep{morrill2020hindsight}. The reinforcement learning (RL) training can be thought of as embedding in the neural network's weights the logic for a state-machine capable of (approximately) Bayes-optimal cultural transmission at test time \citep{mikulik2020meta}. In our trained agent, we indeed found particular memory neurons that encode a subset of the sufficient statistics required for solving the task \citep{DBLP:journals/corr/abs-1905-03030}. Our chosen task has a periodic structure, generating reliable information that an agent can discover and exploit within an episode. Our agent possesses an LSTM memory \citep{hochreiter1997long} and observes its own reward, inspired by the setup in \citet{wang2016learning, duan2016rl}. Apart from these minimal affordances, we do not require any explicit meta-learning algorithms (e.g., \cite{mitchell2021offline}) to train our cultural transmission policy.  
\clearpage
\section{Discussion}
\label{discussion}

\subsection{Summary}

We train an artificial agent capable of robust real-time cultural transmission, and we do so without using human data in the training pipeline. Our work offers the following novel contributions:

\begin{enumerate}
    \item We present a trained neural network solving the full cultural transmission problem, not only inferring information about the task from an expert as in prior work, but also remembering that information within an episode after the expert has dropped out, and leveraging the information to solve hard exploration problems.
    \item We demonstrate that our cultural transmission agent can generalise in few-shot to a wider space of held-out tasks than previously considered, varying the positions of objects in the world, the structure of the game, and the behaviour of the expert player. Moreover, we show that our agent can generalise to cultural transmission from a human expert, a novel step.
    \item Via ablations we identify a minimal sufficient set of ingredients for the emergence of cultural transmission, including several elements which were not previously studied in this context, namely an other-agent attention auxiliary loss, within-episode expert dropout and automatic domain randomisation. 
\end{enumerate} 

\subsection{Commentary}

There are two dual perspectives on open-endedness in RL systems. On the one hand, one can try to construct an agent which has ever increasing reward in a single rich and complex task. On the other, one can try to construct an agent which achieves a reward threshold in an ever more complex space of tasks. In this work, we have found it convenient to take the latter view, because it is more pragmatic for curriculum design. Our agent algorithm itself is surprisingly simple. Perhaps fewer, stronger components is a natural direction of travel when generating more general, social, adaptive agents. We expect that representation learning, scaleable RL algorithms and co-adaptation of training experience will be key for future work in this space. 

We have characterised our approach to generating cultural transmission as memory-based meta-learning. This has several benefits. Imagine deploying a robot in a kitchen. One would hope that the robot would adapt quickly online if the spoons are moved, or if a new chef turns up with different skills. Moreover, one might have privacy concerns about relaying large quantities of data to a central server for training. Our agent adapts to human cultural data online, on-the-fly and within its local memory. It is both robust and privacy-preserving. 

We have shown that social learning research scales to complex $3$-dimensional task distributions. However, we emphasise that scale is not a pre-requisite for a research program on cultural evolution in AI. Indeed, one could implement the same algorithms on a smaller scale and still generate interesting recall and generalisation capabilities across a targeted range of tasks. The fact that we know that the approach scales is useful in motivating further research effort, both small-compute and large-compute. 

\subsection{Limitations}

From a safety perspective, our trained agent as an artefact has a case of objective misgeneralisation \citep{shah2022objective}, which we see in Figure \ref{fig:wrong_demo_trajectory}. The agent happily follows an incorrect demonstration and even reproduces that incorrect path once the expert has dropped out. This is unsurprising since the agent has only encountered near-perfect experts in training. To mitigate this, we hypothesise that an agent should be trained with a variety of experts, including noisy experts and experts with objectives that are different from the agent's. With such experience, an agent should meta-learn a selective social learning strategy \citep{doi:10.1177/0963721415613962, rendell10}, deciding what knowledge to infer from which experts, and when to rely on independent exploration. 

We identify three limitations with our evaluation approach. Firstly, we did not test cultural transmission from a range of humans, but rather with a single player from within the study team. Therefore we cannot make statistically significant claims about robustness across a human population. Secondly, the diversity of reasonable human behaviour is constrained by our navigation task. To gain more insight into generalisable cultural transmission, we need tasks with greater strategic breadth and depth. Lastly, we do not distinguish whether our trained agents memorise a geographical path, or whether they memorise an abstraction over the correct sphere ordering. To disambiguate this, one could change the geographical location of goal spheres at the moment of expert dropout but leave the ordering the same.

Reflecting on whether our method is constrained by our choice of the GoalCycle3D environment, we note that this is already a large, procedurally-generated task space. Moreover, it is the navigational representative for an even bigger class of tasks: those which require a repeated sequence of strategic choices, such as cooking, wayfinding and problem solving. It is reasonable to expect that similar methods would work well in other representative environments from this class of tasks. 

However, there are environmental affordances that we necessarily assume for our method, including expert visibility, dropout and procedural generation. If these are impossible to create or approximate in an environment, then our method cannot be applied. More subtly there are silent assumptions: that finding an initial reward is relatively easy, there is no fine-motor control necessary, the timescale for an episode is relatively short, there are no irreversibilities, the goals are all visible, the rewarding order remains constant throughout an episode. We encourage readers to relax each of these requirements in future work, creating new challenges for research.

Our method is limited in that it requires high-quality co-players for training. We were able to use a hand-coded bot, but this is not typically available in more challenging domains. One potential solution to this problem is to bootstrap agent capabilities across generations using generational training \citep{oelt2021open, vinyals2019grandmaster}. The resulting ratchet effect would lead to agents gradually becoming more capable at cultural transmission. This is a reasonable hypothesis, since cultural transmission can be viewed as amortised distillation, and distillation is already known to work well in generational training. 

\subsection{Future work} 
\label{sec:future_work}

We have focused on the imitation of behaviours. Humans also imitate over more abstract representations of the task at hand, including beliefs, desires and intentions. Whether co-adaptation of training experience can lead to such ``theory of mind'' in artificial agents remains an open question. This approach would complement explicit model-based methods adopted in prior work (e.g. \cite{bratman1987intention, DBLP:journals/corr/abs-1802-07740}).

We motivated this work by observing that cultural transmission is the inheritance process in cultural evolution. In the previous section, we argued that appropriate randomisation over experts may generate a selection process for knowledge. The final missing ingredient for the evolutionary ratchet is variation in behaviour space. Fortunately, there are a variety of off-the-shelf techniques for generating diverse policies \citep{rakicevic2021policy, DBLP:journals/corr/abs-1802-06070, NEURIPS2020_d1dc3a82}. Bringing all of these components together, it would be fascinating to validate or falsify the hypothesis that cultural evolution in a population of agents may yield progressively more generally capable artificial intelligence. 

We do not view our MEDAL-ADR method as a direct model for the development of cultural transmission during human ontogeny. However, the time is ripe for such research. The experimental (e.g. \cite{legare2017cumulative, saldana2019cumulative}) and theoretical (e.g. \cite{tomasello2019becoming, heyes2018cognitive}) fields are already well-developed, and this work provides plausible AI modelling techniques. Neuroscience and deep RL have already mutually benefited from collaborations with a modelling flavour \citep{hassabis2017}, and the precedent has been set for MARL as a modelling tool in social cognition \citep{DBLP:journals/corr/abs-2002-02325, DBLP:journals/corr/abs-2103-04982}. We look forward to fruitful interdisciplinary interaction between the fields of AI and cultural evolutionary psychology in the future. 
\clearpage

\section{Authors and contributions}
\label{authors}

Below is a list of authors and contributors listed alphabetically by last name. Each name is followed by the contributions of the individual.

\begin{itemize}
    \item \textbf{Avishkar Bhoopchand}: Learning process development, research investigations, infrastructure development, technical management, manuscript editing.
    \item \textbf{Bethanie Brownfield}: Quality assurance testing.
    \item \textbf{Adrian Collister}: Environment development.
    \item \textbf{Agustin Dal Lago}: Agent analysis, infrastructure development, code quality.
    \item \textbf{Ashley Edwards}: Research investigations, agent analysis. 
    \item \textbf{Richard Everett}: Cultural general intelligence concept, agent analysis, learning process development, research investigations, infrastructure development, additional environment design, technical management, team management.
    \item \textbf{Alexandre Fr\'echette}: Infrastructure development.
    \item \textbf{Yanko Gitahy Oliveira}: Environment development.
    \item \textbf{Edward Hughes}: Cultural general intelligence concept, research investigations, infrastructure development, team management, manuscript editing. 
    \item \textbf{Kory W. Mathewson}: Learning process development, research investigations, agent analysis, team management.
    \item \textbf{Piermaria Mendolicchio}: Quality assurance testing.
    \item \textbf{Julia Pawar}: Program management.
    \item \textbf{Miruna P\^{i}slar}: Learning process development, research investigations, agent analysis, infrastructure development.
    \item \textbf{Alex Platonov}: Environment visuals.
    \item \textbf{Evan Senter}: Infrastructure development, technical management, team management.
    \item \textbf{Sukhdeep Singh}: Program management. 
    \item \textbf{Alexander Zacherl}: Environment design, environment development, agent analysis.
    \item \textbf{Lei M.~Zhang}: Learning process development, research investigations, infrastructure development, technical management.
    
\end{itemize}

\paragraph{Corresponding authors:} Edward Hughes (\href{mailto:edwardhughes@deepmind.com}{\nolinkurl{edwardhughes@deepmind.com}}), Richard Everett (\href{mailto:reverett@deepmind.com}{\nolinkurl{reverett@deepmind.com}}).

\section*{Acknowledgements}
\label{acknowledgements}

We would like to thank many, both at DeepMind and in the wider community, for the conversations that have helped to shape this manuscript. We are particularly grateful to (in alphabetical order by last name): Lucy Aplin, Michael Azzam, Jakob Bauer, Satinder Baveja, Charles Blundell, Andrew Bolt, Kalesha Bullard, Max Cant, Nando de Freitas, Seb Flennerhag, Antonio Garcia Castañeda, Thore Graepel, Abhinav Gupta, Nik Hemmings, Max Jaderberg, Natasha Jaques, Andrei Kashin, Simon Kirby, Tom McGrath, Hamza Merzic, Alexandre Moufarek, Kamal Ndousse, Sherjil Ozair, Patrick Pilarski, Drew Purves, Thom Scott-Phillips, Ari Strandburg-Peshkin, DJ Strouse, Kate Tolstaya, Karl Tuyls, Marcus Wainwright, and Jane Wang.

\clearpage

\bibliographystyle{abbrvnat}
\nobibliography*
\bibliography{refs}

\clearpage
\appendix
\counterwithin{figure}{section}
\counterwithin{table}{section}

\section{Task space in detail}\label{app:task_space}

\subsection{World space}

The playable terrain is perfectly square and varies in size from $\SI{324}{\meter\squared}$ to $\SI{961}{\meter\squared}$. The terrain height field is generated using multi-octave coherent noise, using the Libnoise library \citep{bevinslibnoise} with Perlin noise as the underlying pseudorandom number generator. The frequency values of the lowest octave of noise varies between $\SI{0.01}{\per\meter}$ and $\SI{0.1}{\per\meter}$. The amplitude of the noise varies between $\SI{0}{\meter}$ and $\SI{5}{\meter}$.

\textit{Vertical obstacles} are perpendicular to the terrain and require the player to move around them. Their density can reach up to $\SI{0.05}{\per\meter\squared}$. In a flat terrain of size \SI{24}{\meter\squared}, the maximum value produces 29 vertical obstacles on average. \textit{Horizontal obstacles} lie flat on the terrain and require the player to jump over or (in rarer cases) crouch under them. Their density can reach up to $\SI{0.007}{\per\meter\squared}$. In a flat terrain of size \SI{24}{\meter\squared}, the maximum value produces 4 horizontal obstacles on average.

Obstacles are constrained to spawn only on surfaces with a absolute value of slope $< 10\degree$ for vertical obstacles and $< 30\degree$ for horizontal obstacles. Therefore, given a fixed obstacle density, there is an inverse relationship between the bumpiness of the terrain and the density of the obstacles. Each horizontal obstacle is of fixed length $\SI{16}{\meter}$ and random rotation in the horizontal plane. Each vertical obstacle is of diameter sampled uniformly at random between $\SI{0.75}{\meter}$ and $\SI{2.25}{\meter}$, and of sufficient height that no player can jump over it.

\begin{figure}[ht]
\centering
\includegraphics[width=1.0\linewidth]{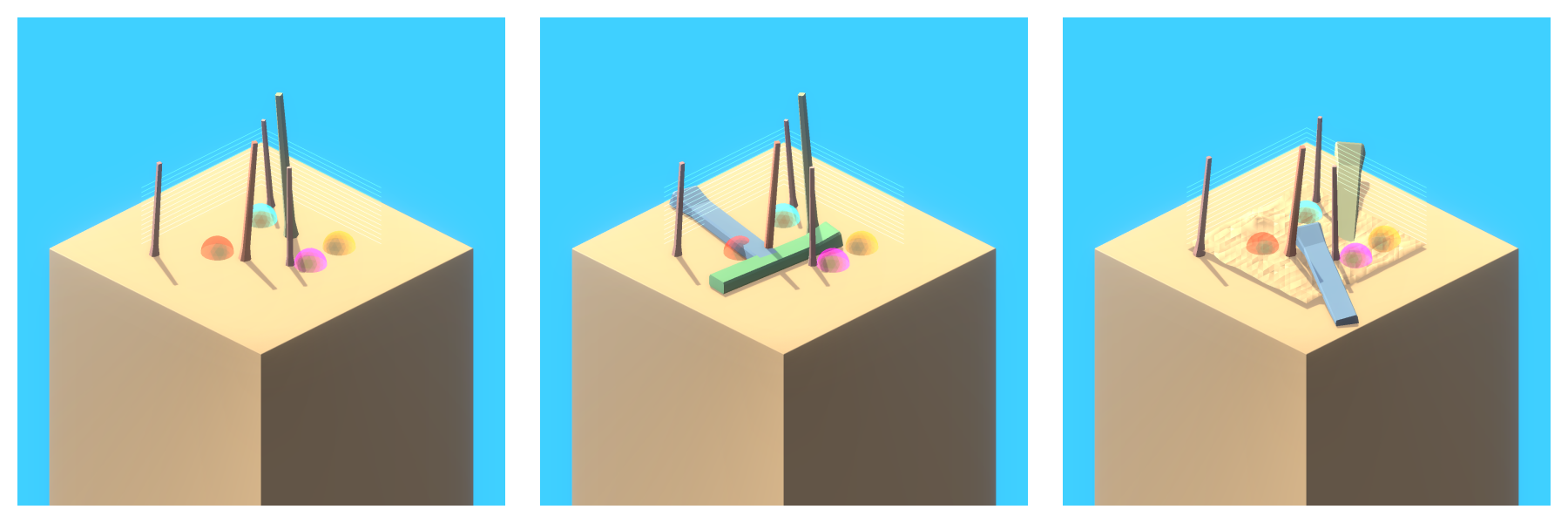}
\caption{Three worlds created from the same seed but with different procedural generation parameters.}
\label{fig:worlds-same-seed}
\end{figure}

\begin{figure}[ht]
\centering
\includegraphics[width=1.0\linewidth]{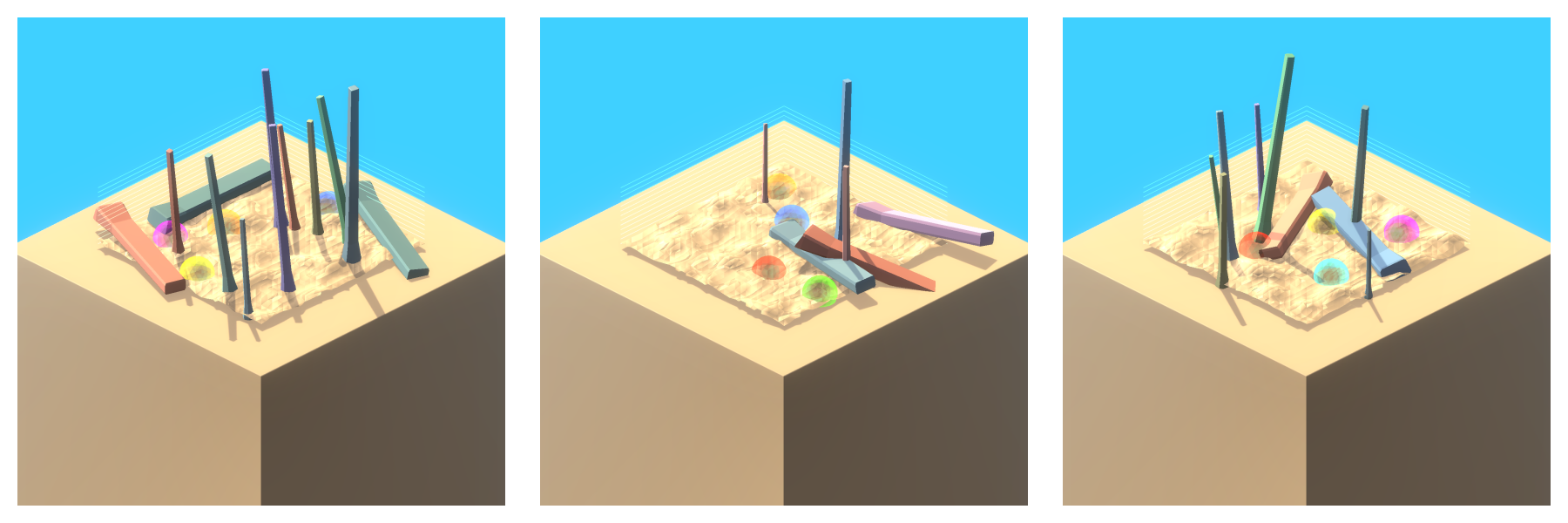}
\caption{Three worlds generated from the same procedural generation parameters but different random seeds.}
\label{fig:worlds-three-seeds}
\end{figure}

\subsection{Game space}

In mathematical terms, the distinct orders are Hamiltonian cycles on the complete graph $K_n$ where $n$ is the number of goals, the nodes of the graph are the goals, and the edges represent equivalence classes of paths between one goal and another, without passing through any intermediate goal. In $K_n$ there are $(n-1)!$ distinct Hamiltonian cycles, as may be readily verified by a counting argument. More formally, the order $\sigma^{-1}$ is the group inverse of $\sigma$ when viewed as an element of the symmetric group $S_n$.

In principle, subcycles may also be positively rewarding, but we show empirically that these are not optimal on average in Table \ref{tab:subcycles}. Goal sphere colours are sampled uniformly at random without replacement from a set of $8$ possibilities, depicted in Figure \ref{fig:env-at-a-glance} (bottom left). The diameter of each goal sphere is fixed as a function of the world size $w$ to be $\frac{w}{8} + 2$. Note that sampling the rewarding cycle and goal positions uniformly at random does not produce a uniform distribution over topologies.

\begin{figure}[htb]
\centering
\begin{subfigure}{.45\textwidth}
  \centering
  \includegraphics[width=0.7\linewidth]{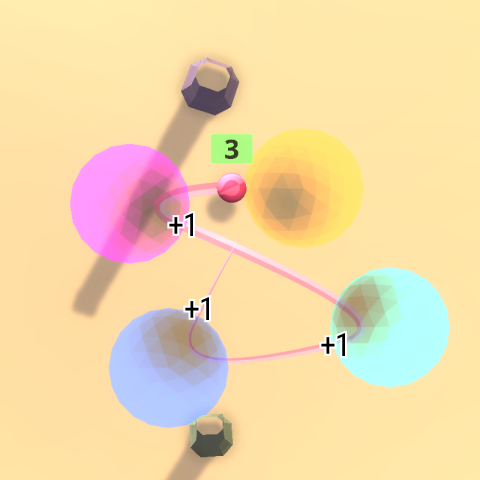}
\end{subfigure}
\hspace{1em}
\begin{subfigure}{.45\textwidth}
  \centering
  \includegraphics[width=0.7\linewidth]{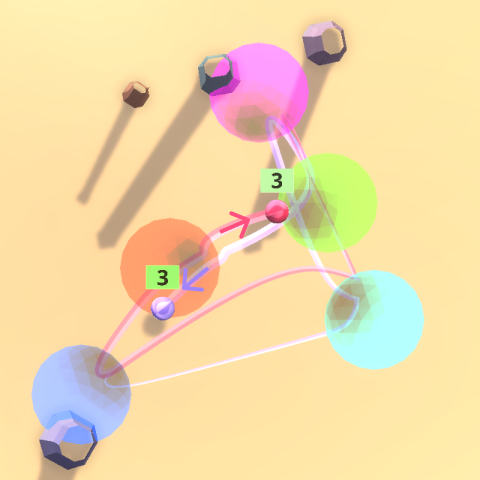}
\end{subfigure}
\caption{On the left, a single player navigates between four goals, with annotations showing when the score changes. On the right, two players demonstrate the two equally rewarding paths available between five goals.}
  \label{fig:goal-cycle-mechs}
\end{figure}

\begin{table}[h]
\centering
\subfloat[\centering]{
    \begin{tabular}{c|c}
         Length of subcycle & Average score \\ \hline
         $2$ & $1.2 \pm 1$ \\
         $3$ & $8.6 \pm 1$ \\
         $\mathbf{4}$ & $\mathbf{20.4 \pm 2}$ \\
         & \\
    \end{tabular}
    \label{tab:subcycles-a}
}
\subfloat[\centering]{
    \begin{tabular}{c|c}
         Length of subcycle & Average score \\ \hline
         $2$ & $1.2 \pm 1$ \\
         $3$ & $8.3 \pm 2$ \\
         $4$ & $12.6 \pm 2$ \\
         $\mathbf{5}$ & $\mathbf{23.5 \pm 3}$ \\
    \end{tabular}
    \label{tab:subcycles-b}
}
\caption{Average scores for subcycles of a representative (\subref{tab:subcycles-a}) 4-goal game and (\subref{tab:subcycles-b}) 5-goal game. The subcycles were sampled such that each goal sphere correctly follows the preceding sphere in one of the correct orders of the full cycle, with a disconnect only allowed at the beginning and end of the subcycle. Each row represents the average and one standard deviation of the score over 20 samples obtained by an expert bot following a subcycle of the given length. All the tasks used a $32\times32$ world with flat terrain and no obstacles.}
\label{tab:subcycles}
\end{table}

\subsection{Player interface}

Players are embodied as physical avatars. Avatars interact with the environment simulation on discrete timesteps. On each timestep an avatar perceives its surroundings using sensors and converts this to an observation. This observation is sent to the player who must return an action vector, which the avatar converts to actuation values passed to the environment simulator. The environment simulator takes actuation information from all avatars and performs a step of physics simulation to generate the next global state for the avatars to perceive. The avatar is surrounded by a collision mesh, which can be disabled to allow evaluation of agents alongside pre-recorded human trajectories.

Viewing distance for both humans and agents is fixed at $\SI{128}{\meter}$, sufficient to perceive to the edge of the world from anywhere in the world, for all world sizes used in this work. For humans, obstacles, terrain and avatars are opaque, goals are translucent and the boundary of the world is visible. Humans also perceive a box floating above each avatar displaying the avatar's current score, which is reset to $0$ at the beginning of each episode. LIDAR is a common and efficient observation modality for real-world robots (e.g. \cite{MALAVAZI201871}) and in prior multi-agent 3D physical simulated worlds (e.g. \cite{DBLP:journals/corr/abs-1909-07528}). For agent LIDAR, the zero direction for the azimuth is fixed to be the FORWARD direction for the avatar, so rotation of the avatar is meaningful perceptually. 

\begin{figure}[ht]
\centering
\begin{subfigure}{0.44\linewidth}
\includegraphics[width=\linewidth]{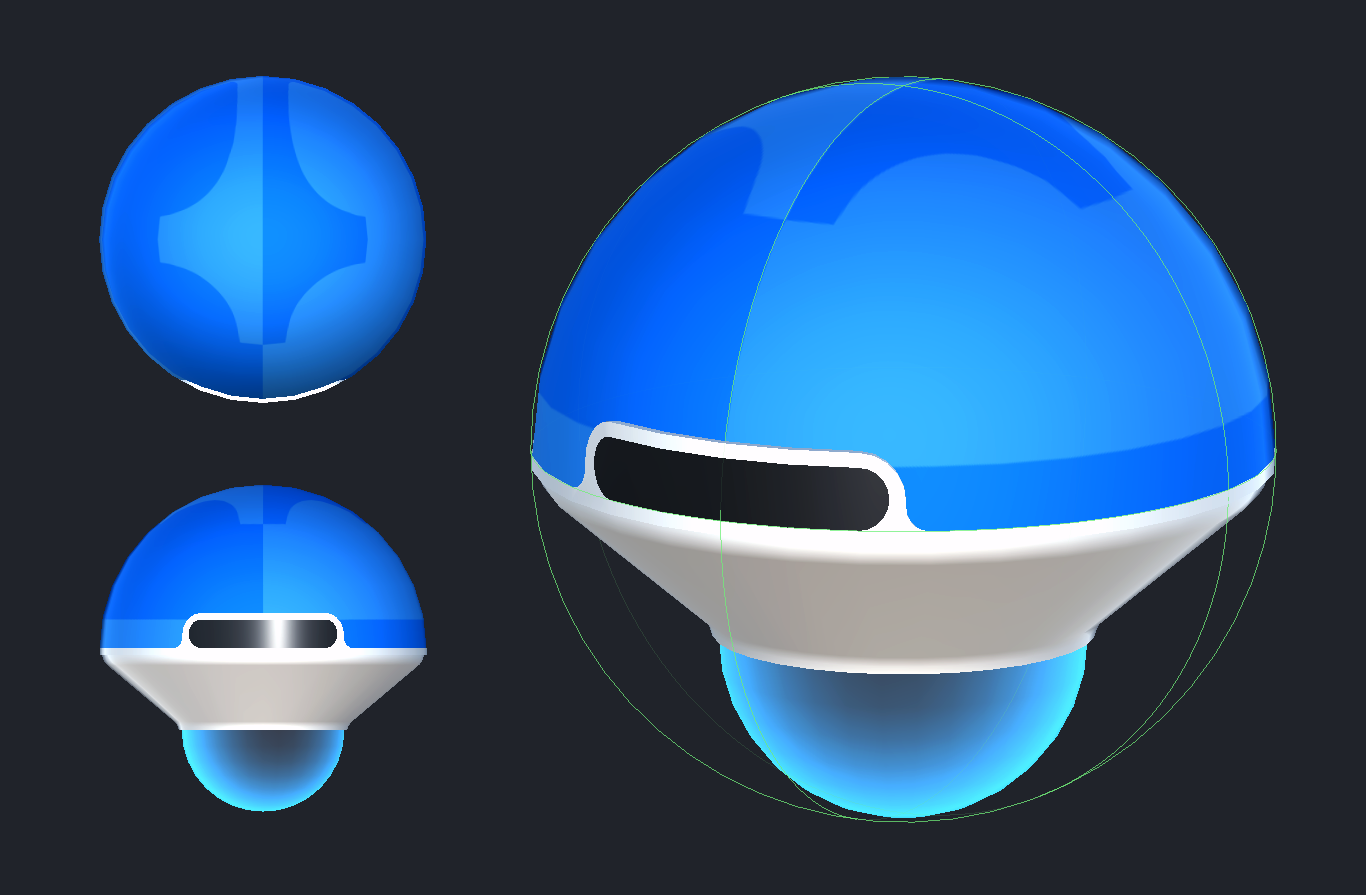}
\caption{Avatars are surrounded by a spherical collision mesh, shown on the right in green.}
\label{fig:avatar-closeup}
\end{subfigure}
\hfill
\begin{subfigure}{0.52\linewidth}
\includegraphics[width=\linewidth]{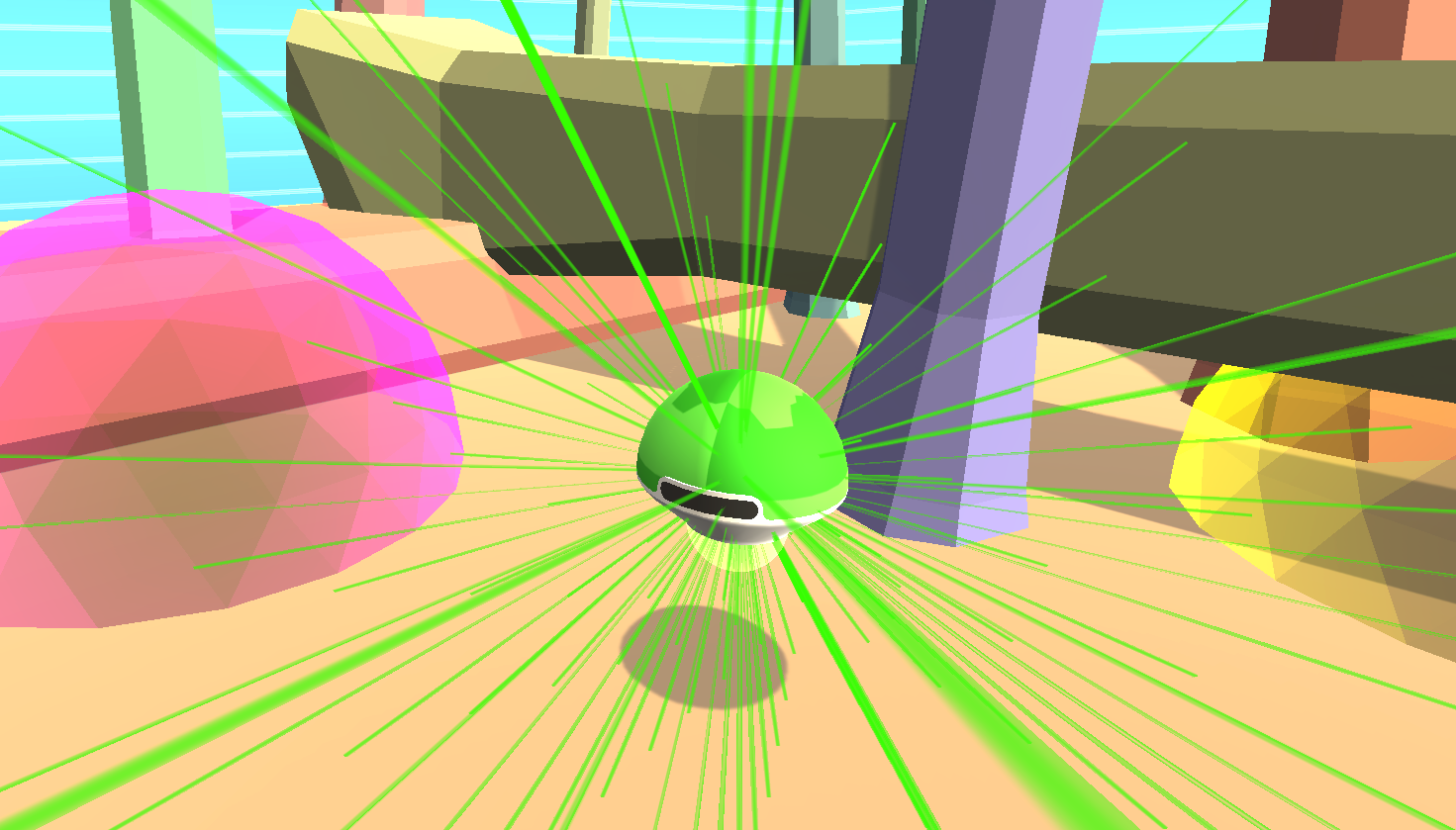}
\caption{LIDAR sensor rays emanate from the centre of an avatar. Here we show only rays that collide with an object in the world. An accompanying \href{https://sites.google.com/view/dm-cgi\#h.4kpbw1z7rww1}{video} is available.}
\label{fig:avatar-lidar}
\end{subfigure}
\caption{Players are embodied as physical avatars. Agents perceive the world through LIDAR.}
\end{figure}

\subsection{Probe tasks}

While our training and evaluation probe tasks are sampled from the same procedural generator, we use different random seeds for training and evaluation to ensure these probe tasks are held out, following \citep{zhang2018study, jaderberg2019human, mckee2021quantifying}. When using ADR, the vast majority of training tasks have different parameters to our probe tasks and are therefore different by definition. For identical task parameters, the hold-out is probabilistic in nature. Using a worst-case argument for 5-goal games (our best agents were never trained on 4-goal games), suppose we placed goals on a discrete grid with blocks of size $\SI{6}{\meter} \times \SI{6}{\meter}$ (corresponding to the goal sphere radius in $\SI{32}{\meter \times \SI{32}\meter}$ worlds) thereby ensuring that overlaps are impossible. This would result in $\binom{25}{5} = 53130$ possible goal placements. We also sample 5 out of 8 colours, giving $3 \times 10^6$ possible goal positions and colours. For flat, empty worlds of size $\SI{32}{\meter} \times \SI{32}{\meter}$ we only ever sample a maximum of 50 such goal placements and colours during training and 12 during evaluation using different seeds. It is therefore highly improbable that any single training task is the same as a probe task.

For the complex tasks with obstacles, consider a grid with blocks of size $\SI{1}{\meter} \times \SI{1}{\meter}$ and an average of 3 vertical obstacles and 3 horizontal obstacles. There are $\binom{1024}{3} = 1.8 \times 10^8$ possible placements for each of these obstacle types (note that obstacle overlaps are allowed by our procedural generator). Multiplying the number of goal placements and colours by the number of vertical obstacle placements and horizontal obstacle placements gives $9 \times 10^{22}$ possible tasks. Again in complex tasks we sample 50 possible world seeds which control the placement of these obstacles, which combined with 50 goal seeds gives only 2500 possible training tasks with the same task parameters used in our probe tasks. The likelihood of these training tasks overlapping the 12 probe tasks we sample is vanishingly small. Note that this already conservative argument ignores the random rotation of horizontal obstacles and also ignores the terrain bumpiness and therefore significantly undercounts the number of possible tasks in practice. 

\section{Learning algorithm in detail}\label{app:learning_algorithm}

\subsection{Reinforcement Learning (RL)}

Let $\mathcal{D}(\Omega)$ denote the space of distributions over the space $\Omega$. A Markov Decision Process (MDP) \citep{howard1960dynamic,sutton2018reinforcement} is a tuple $\langle \mathcal{S}, \mathcal{A}, T, r, \gamma \rangle$ where $\mathcal{S}$ is a set of states, $\mathcal{A}$ is a set of actions, $T: \mathcal{S} \times \mathcal{A} \to \mathcal{D}(\mathcal{S})$ is a transition function, $r: \mathcal{S} \times \mathcal{A} \to \mathbb{R}$ is a reward function, and $\gamma \in [0,1]$ is a discount factor. A mapping $\pi : \mathcal{S} \to \mathcal{D}(\mathcal{A})$ is called a stochastic policy. Given a policy $\pi$ and an initial state $s_0$, we define the value function $V_\pi(s_0) = \mathbb{E} [\sum_{t=0}^\infty\gamma^t\; r(s_t, \pi(s_t))]$ where $s_t$ is a random variable defined by the recurrence relation $s_t = T(s_{t-1}, \pi(s_{t-1}))$. Reinforcement learning seeks to find an optimal policy $\pi^*$ which maximises the value function from an initial state $s_0$. We assume that the agent experiences the world in episodes of finite length $T$. During training, our RL agent receives many episodes of experience, and updates its policy to become closer to the optimal policy, as measured on a held-out episode during testing.

A partially observable Markov Decision Process (POMDP) is defined by the tuple $\langle \mathcal{O}, \mathcal{A}, T, r, \gamma \rangle$, where each element of $\mathcal{O}$ is a partial observation of a true underlying state in $\mathcal{S}$. Typically, multi-agent settings are automatically POMDPs because each agent does not have access to the observations, actions, policies or rewards of their co-players \citep{littman1994markov}. This is especially true in our case, since co-players may also drop in and drop out of the environment within the course of an episode (see Section \ref{sec:expert_dropout}).

The optimal policy $\pi^*$ in an environment containing other agents is at least as good as the optimal policy in the same environment without other agents. This is true because the set of policies in the augmented world is a superset of the policies in the non-augmented world, by virtue of the richer state space in the augmented world. Practically speaking, an agent should be able to leverage behaviour of other agents when beneficial without regressing solo performance. 

\subsection{Maximum a Posteriori Policy Optimization (MPO)}

MPO is an actor-critic, model-free, continuous action-space reinforcement learning algorithm introduced by \cite{Abdolmaleki2018MaximumAP}. Instead of using gradients from the $Q$-function, it leverages samples to compare different actions in a particular state, updating the policy to ensure that better (according to the current $Q$-function) actions have a higher probability of being sampled. As with other actor-critic algorithms, MPO alternates between \emph{policy improvement} which updates the policy ($\pi$) using a fixed $Q$-function and \emph{policy evaluation} which updates the estimate of the $Q$-function.

Policy improvement uses Expectation Maximisation (EM) coordinate ascent, alternating between a non-parametric E-step and a parametric M-step. The E-step re-weights state-action samples, assigning them weights $q_{ij}$ according to Equation \ref{e-step}. Here $a_i$ refers to an action sampled under state $s_j$, $Q^{\pi}$ is the Q-function of the current policy $\pi^{(k)}$, $Z_j$ is a normalisation term and the temperature $\eta$ is a Lagrange multiplier. The M-step solves the primal optimisation problem in Equation \ref{m-step} by performing one iteration of optimising the Lagrange multiplier $\alpha$ holding the policy parameters $\theta$ constant, and then optimising $\theta$ holding $\alpha$ constant. The KL regularisation term prevents the policy from changing ``too quickly'' based on samples from a potentially inaccurate $Q$-function. This regularisation term is averaged over the $K$ states sampled and its strength is controlled by the hyperparameter $\epsilon$. Retrace (\cite{Munos2016SafeAE}) is used for policy estimation using off-policy trajectories. When used in continuous control, MPO fits the mean and covariance of a Gaussian distribution over the joint action space. 

\begin{gather}
    \label{e-step}
    q_{ij} = q(a_i, s_j) = Z_j^{-1}\exp\left(\frac{Q^{\pi}(s_j, a_i)}{\eta}\right) \,. \\
    \label{m-step}
    \max_{\theta} \min_{\alpha > 0} L(\mathbf{\theta}, \eta) = \max_{\theta} \min_{\alpha > 0} \left[\sum_{j} \sum_{i} q_{ij} \log \pi_{\mathbf{\theta}} (a_i | s_j) + \alpha\left(\epsilon - K^{-1}\sum_{j}^{K} \mathrm{KL}\left( \pi^{(k)}(a|s_j)\; || \;\pi_{\mathbf{\theta}}(a|s_j) \right) \right) \right]\,.
\end{gather}

\section{Methods in detail}

\subsection{Training framework}\label{sec:training-framework}

\begin{figure}[ht]
    \centering
    \includegraphics[width=0.9\linewidth]{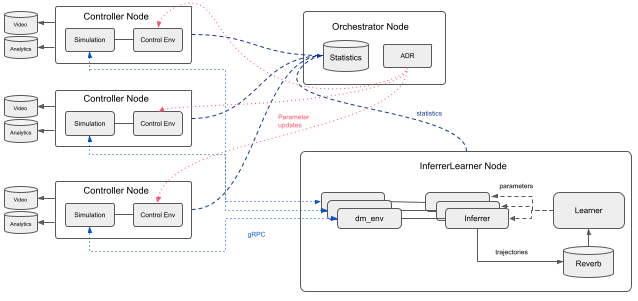}
    \caption{High level system architecture diagram, showing the relationship between \textit{InferrerLearner}, \textit{World} and \textit{Orchestrator} nodes.}
    \label{fig:system-architecture}
\end{figure}

Our distributed training architecture is depicted in Figure \ref{fig:system-architecture}. Each experiment has 192 independent \textit{GoalCycle3D} simulations, capable of accepting connections from any number of players (agent or human). These simulations run on Nvidia Tesla P4 and V100 GPUs, with 8 simulations per GPU. Connected to each environment is a \textit{(World) Controller} node which connects to the simulation using an environment interface. The environment interface collects analytical data from the simulation, including raw RGB observations from a third-person camera that are stitched together and periodically saved as videos. The \textit{Controller} node also manages episodes by keeping track of the (synchronised) steps taken by connected players and indicating the end of an episode to the simulation after $1800$ steps have occurred. The simulation handles synchronisation by blocking until all connected players have sent their action before stepping. It also signals the end of an episode to connected players by setting a ``last step'' flag in the connected player's observations. The \textit{Controller} also manages ``soft resetting'' of the simulation, where simulation parameters such as number and position of goals, vertical and horizontal and obstacle densities, terrain bumpiness, avatar spawn positions and world size are updated before re-rendering the simulation. These soft resets are aligned with the synchronised episode boundaries. 

Each agent in an experiment is controlled by an independent \textit{InferrerLearner} node. These nodes do not communicate with each other. Each \textit{InferrerLearner} node connects to all of the running simulations and communicates with these (sending actions and receiving observations) using a remote \texttt{dm\textunderscore env} (\cite{dm_env2019}) interface over gRPC. Thus, even though this is a multi-agent architecture, the problem looks like a local, single agent RL problem to an individual agent. Each \textit{InferrerLearner} node is divided into an \textit{Inferrer} and a \textit{Learner}, which both run continuously, sharing a single Google Cloud TPU v2 host. This enables high speed memory access between the \textit{Inferrer} and \textit{Learner} for parameter synchronisation. 

An \textit{Inferrer} maintains a thread for each environment it is connected to which runs the standard RL loop of receiving an observation, passing this through the agent network to obtain an action and returning this action to the environment. Trajectories of observations, rewards, actions and additional metadata computed by the agent in all threads are saved to a single FIFO experience replay buffer implemented using Reverb (\cite{cassirer2021reverb}). 

The learner continuously queries the replay buffer for a batch of trajectories, blocking if insufficient data is available in the buffer. A copy of the agent's parameters are updated using gradient descent on this batch of trajectories, before pushing the updated parameters to the policies running on the \textit{Inferrer} threads. 

A final node, called the \textit{Orchestrator} is responsible for coordinating the connection and disconnection of \textit{InferrerLearner} nodes to and from simulations. The \textit{Orchestrator} also keeps track of aggregate statistics from \textit{Controller} nodes including total step counts and average scores, which are used to calculate a training CT metric. It coordinates simulation parameter updates through the ADR mechanism (see section \ref{sec:adr}). These simulation parameter updates are pushed to the \textit{Controller} nodes which store them in a buffer, waiting to apply to them to the simulation when the next episode boundary is reached. 

The topology of the full distributed system, consisting of \textit{Orchestrator}, \textit{Controller}, \textit{InferrerLearner} nodes and Reverb replay buffers is configured and mapped to hardware using Launchpad (\cite{yang2021launchpad}).

\subsection{Training hyperparameters}\label{sec:hyperparameters}

\begin{figure}[htb]
    \centering
    \includegraphics[width=1.0\linewidth]{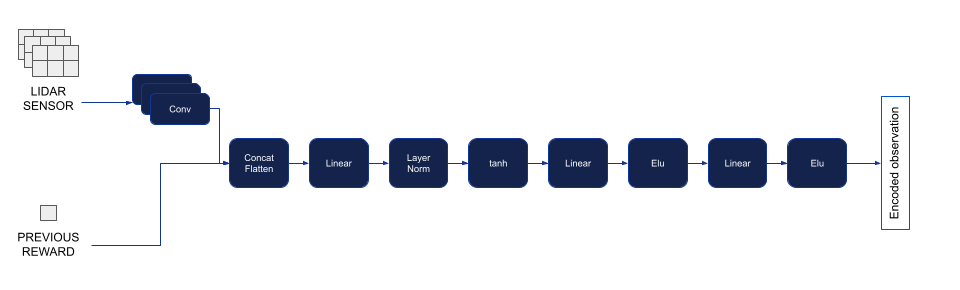}
    \caption{Encoder architecture.}
    \label{fig:encoder-architecture}
\end{figure}

The agent receives observations from the environment, computed by LIDAR and AVATAR sensors described in Section \ref{sec:player-interface}. The LIDAR sensor is used as an input to the agent's neural network, but the AVATAR sensor is not. We additionally feed the previous step's reward into the agent's neural network. The $14\times14\times9$ LIDAR sensor is passed through a convolutional layer with 24 square $2\times2$ kernels. The previous reward is concatenated to the output of this convolutional layer before being passed to a linear layer of size 128. Layer norm is applied before a \textit{tanh} activation followed by 2 linear layers of size 256 with \textit{elu} activation in between. The encoder is illustrated in figure \ref{fig:encoder-architecture}, resulting in an encoded observation. 

Table \ref{tab:hyperparams} presents the hyperparameters used during training of our agents . The values were either the defaults found in the DeepMind JAX Ecosystem \citep{deepmind2020jax} or were set using small-scale sweeps early in experimentation. 

\begin{table}[h]
\centering
\begin{tabular}{p{0.18\linewidth} | p{0.6\linewidth} | p{0.18\linewidth}}
     \textbf{Hyperparameter} & \textbf{Description} & \textbf{Value} \\ \hline
     \multicolumn{3}{l}{\textbf{MPO Hyperparameters} (See \cite{Abdolmaleki2018MaximumAP} for details)} \\ \hline
     Policy Samples & Number of actions to sample from the policy  & 100 \\
     Min stdev & A minimum (epsilon) standard deviation to add to the predicted value to prevent it from being 0. & 1e-4 \\
     Action embedding & Transformation function applied to action samples & tanh \\
     Alpha Init & Initial value of Lagrange multiplier for KL mean and covariance & 1 \\
     Epsilon Alpha KL mean & Epsilon for Lagrange multiplier for KL mean & 2.4e-2 \\
     Epsilon Alpha KL cov & Epsilon for Lagrange multiplier for KL covariance & 7e-5 \\
     Temperature Init & Initial value of temperature Lagrange multiplier & 0.325 \\
     Epsilon Temperature & Epsilon for temperature Lagrange multiplier & 2e-2 \\
     Gamma & Discount factor applied to returns & 0.99387 \\
     
     \multicolumn{3}{l}{\textbf{Training hyperparameters}} \\ \hline
     
     Critic learning rate & Learning rate of ADAM optimiser \citep{Kingma2015AdamAM} applied only to the critic neural network & 4.8e-4 \\
     Actor learning rate & Learning rate of ADAM optimiser applied to policy, encoder, RNN and attention loss networks & 2e-4 \\
     Network update period & Number of steps after which the online network is replaced with the target network & 100 \\
     Attention loss weight & Relative weight of the attention loss relative to the MPO loss & 10 \\
     Attention loss step & The (future) prediction step used by the attention loss & 0 \\
     Batch size & The batch size for learner steps & 32 \\
     Trajectory length & The number of steps over which the RNN is unrolled in each learner step & 800 \\
     Episode length & The number of steps making up an episode. The agent's state is reset and it respawns at a random location at the beginning of an episode. & 1800 \\

\end{tabular}

\label{tab:subcycles-a}
\caption{Hyperparameters.}
\label{tab:hyperparams}
\end{table}

\subsection{Attention Loss (AL)}\label{app:bonus-attention-loss}

Following up from section \ref{sec:auxloss}, we provide a discussion and some extra empirical findings that support the design choices in our attention loss. These include: (1) the time step at which to make the prediction; (2) the weight of the attention loss and (3) adding noise to the prediction targets. 

\begin{figure}[H]
    \centering
    \begin{subfigure}[b]{0.3\textwidth}
        \centering
        \includegraphics[width=\textwidth]{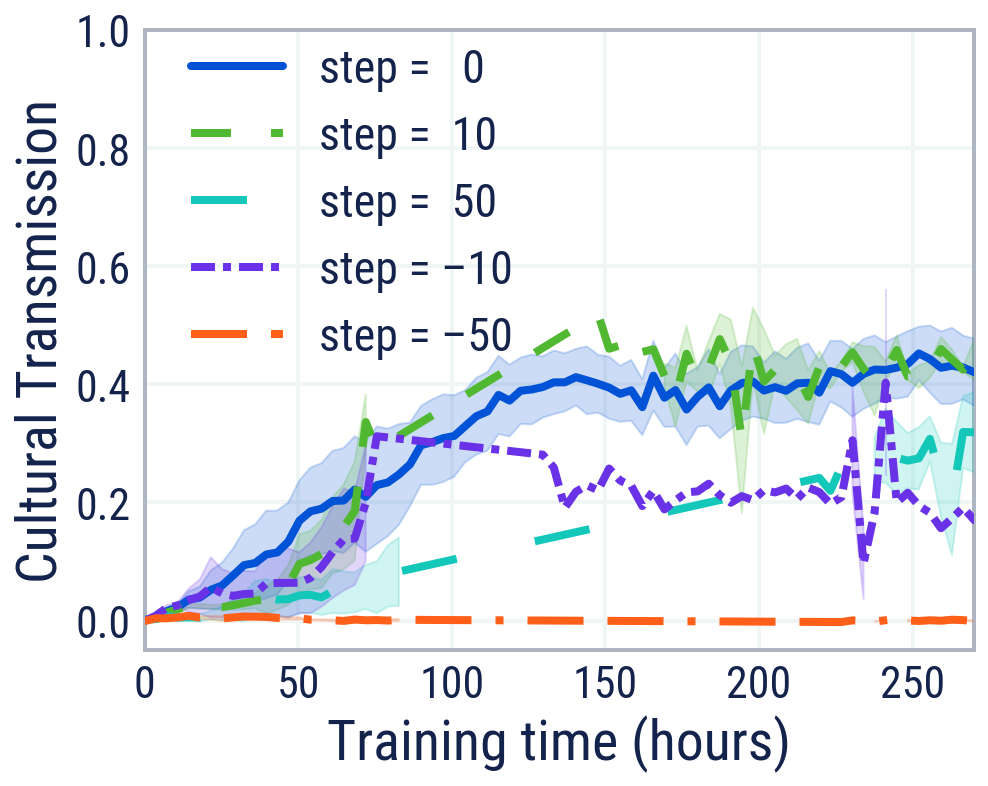}
        \caption{Prediction step\centering}
        \label{subfig:ablate-step}
    \end{subfigure}
    \hfill
    \begin{subfigure}[b]{0.3\textwidth}
        \centering
        \includegraphics[width=\textwidth]{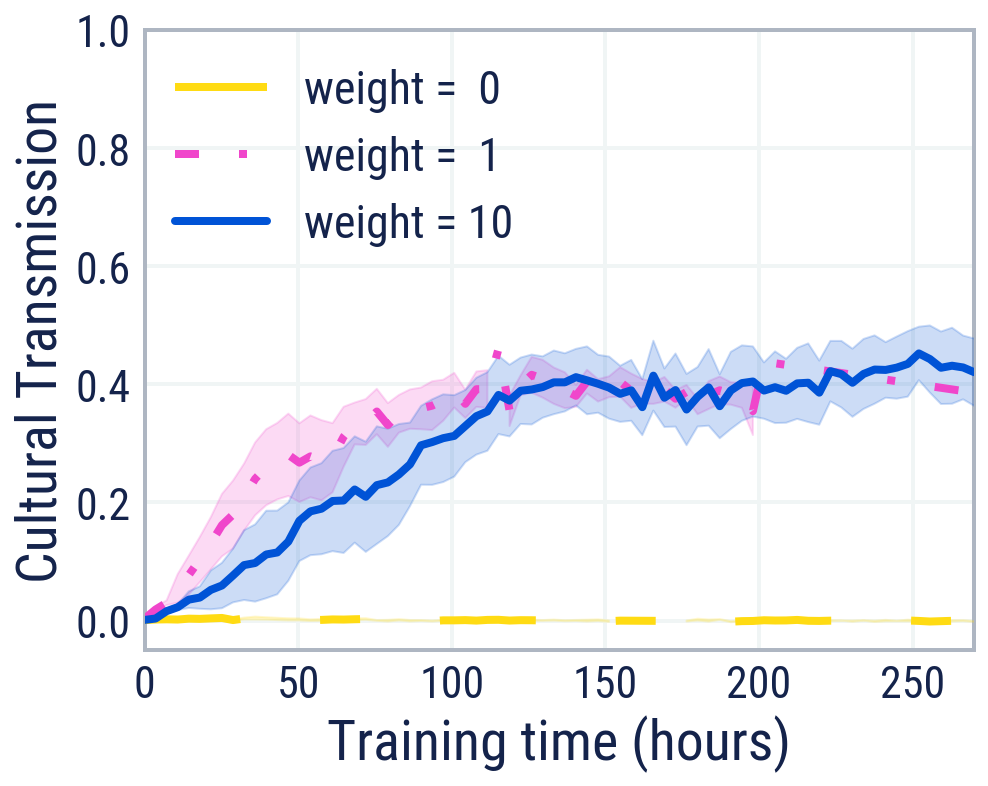}
        \caption{Loss weight\centering}
        \label{subfig:ablate-cost}
    \end{subfigure}
    \hfill
    \begin{subfigure}[b]{0.3\textwidth}
        \centering
        \includegraphics[width=\textwidth]{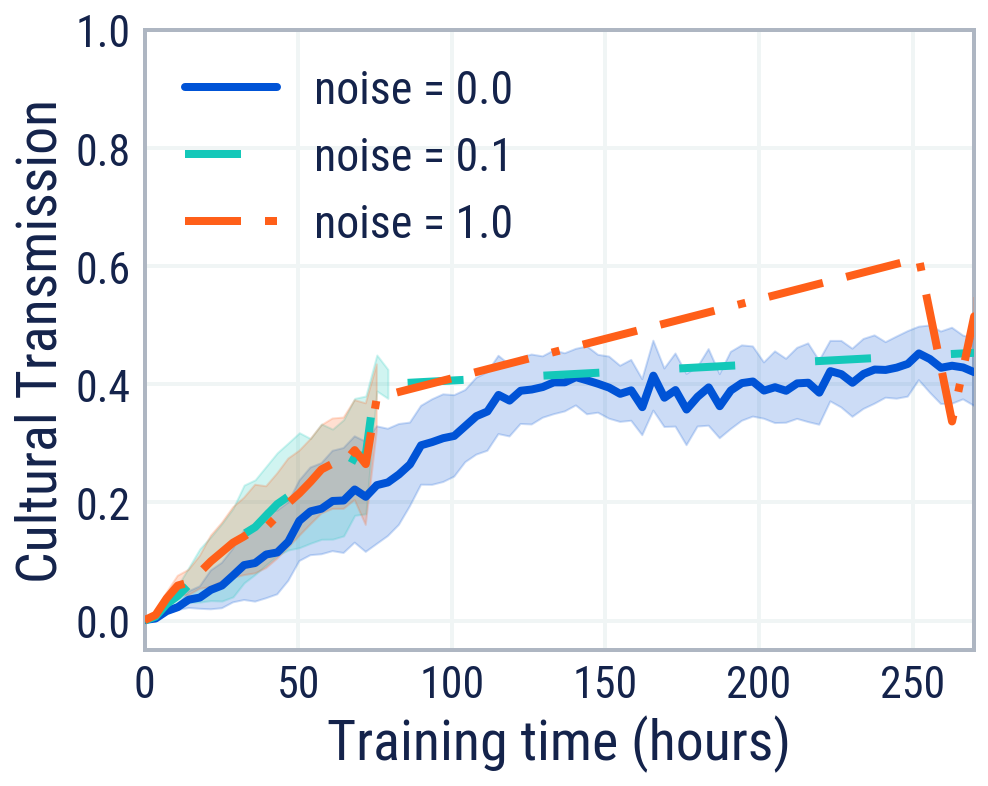}
        \caption{Target noise\centering}
        \label{subfig:ablate-noise}
    \end{subfigure}
    \caption{Extra empirical findings on the attention loss, ablating past, present, and future steps as alternatives for the prediction time (\subref{subfig:ablate-step}), the weight of the attention loss (\subref{subfig:ablate-cost}), and the amount of normal noise added on top of the prediction targets (\subref{subfig:ablate-noise}).}
    \label{fig:ablate-aux-loss-components}
\end{figure}

\begin{description}

\item[The prediction time step] is, by default, the current step for all our agents. However, predicting a past or a future step is also a sensible alternative. More formally, we can predict scaled coordinates $(\hat{x}_{t+n}, \hat{y}_{t+n}, \hat{z}_{t+n})$ for arbitrary choices of $n$, as in Equation \ref{eq:aux-loss-3} below. Positive values of $n$ correspond to a future step, potentially helping the agent build a model of the world and the other players. Negative values of $n$ target a past step, potentially enhancing the agent's memory. Having tried several values for $n$, we conclude that predicting the present (i.e., setting $n$ to 0) results in the best performance, as shown in Figure \ref{subfig:ablate-step}. This choice gives our attention loss the flavour of a reconstruction loss and resembles working memory. Note that we don't consider $n$ of 1 (as used in \cite{ndousse2020learning}) or -1 because the difference between single timesteps in our complex 3D simulation is small compared to a 2D grid-world.

\begin{equation}\label{eq:aux-loss-3}
(\hat{x}_{t+n}, \hat{y}_{t+n}, \hat{z}_{t+n}) = MLP([LSTM(s_t); a_t]) \in [0, 1] \, .
\end{equation}

\item[The attention loss weight] is the hyper-parameter deciding the importance of the attention loss relative to our agents' primary policy optimisation loss. Figure \ref{subfig:ablate-cost} shows that an agent becomes more socially adept and learns to pay more attention to the expert with a higher weight on the attention loss. Therefore, in all our experiments, we use a weight of $10.0$.

\item[Noisy targets] are a common technique for increasing the robustness to noise and regularise the label predictions during training \citep{szegedy2016rethinking}. However, in our case, noisy predictions did not make a significant difference (see Figure \ref{subfig:ablate-noise}).

\end{description}

\subsection{Automatic Domain Randomisation (ADR)}\label{sec:app_adr}

To train on a diverse set of tasks, the Orchestrator (see Appendix~\ref{sec:training-framework}) periodically re-samples all of the task parameters according to $\lambda \sim P_{\phi}$ and configures the environment simulations using the sampled parameters synchronously. Hence, between these resampling instances, the agent is trained on tasks distributed according to $P_{\phi}$. Note that the resampling interval is set to be $10$ minutes, much longer than the time it takes to complete the number of steps in an episode. In practice, $30$ to $50$ episodes are run using a task with the same parameters. 

To adapt the task distribution to the agent's Goldilocks zone, the distribution parameters $\phi$ are updated before sampling task parameters in the Orchestrator. To facilitate this update, we follow two techniques from the original ADR: boundary sampling and threshold updates.

When sampling a parameter $\lambda$, with probability $p_b \in [0, 1]$ (referred to as the boundary sampling probability) a task parameter $\lambda_{b}$ is chosen uniformly at random to be fixed to one of its `boundaries': $\{\phi_{b}^L\,, \phi_{b}^H\}$. In general, both boundaries are updated by ADR and the boundary value is randomly sampled from the two options. We also allow cases where only one of the boundaries is updated by ADR, while other is fixed (for example, if a parameter is non-negative). In these cases, the variable boundary is used as the sampled value. Once $\lambda_{b}$ is determined, the remaining components of $\lambda$ are sampled according to $P_{\phi}$. Finally, with probability $1 - p_b$, $\lambda$ is sampled directly from $P_{\phi}$.

Each task parameter that was boundary sampled contains one component at index $b$ that matches either $\phi_{b}^L$ or $\phi_{b}^H$. Let $b(\lambda) \in \{1, \dots, d\} \times \{L, H\}$ denote the mapping from a task parameter to the boundary to which $\lambda_{b}$ was fixed, specified by a tuple: $(b, L)$ or $(b, H)$. At the end of an Orchestrator update interval, each simulation sends its training cultural transmission metric (see Section~\ref{sec:ctmetric}) to the queue $q(b(\lambda))$. Note that up to $2d$ separate queues are required. Simulations where $\lambda$ was not boundary sampled do not push their cultural transmission metrics.

A training cultural transmission metric for a given set of task parameters is obtained by collecting agent scores in 3 special episodes with: no dropout, full dropout, and half dropout of the expert bot. These episodes are run every time the Orchestrator resamples task parameters. 

To update $\phi_i^L$ (and similarly $\phi_i^H$), we simply average $q(i, L)$ and compare the average training cultural transmission metric $\bar{c}(i, L)$ against fixed thresholds: $\text{th}_L$, $\text{th}_H$. $\phi_i^L$ is updated according to
\begin{equation}
\phi_i^L = 
\begin{cases}
    \phi_i^L - \Delta_i & \bar{c}(i, L) > \text{th}_H \\
    \phi_i^L + \Delta_i & \bar{c}(i, L) < \text{th}_L \\
    \phi_i^L & \text{otherwise or q(i, L) empty} 
\end{cases} \, ,
\end{equation}
and $\phi_i^H$ according to
\begin{equation}
\phi_i^H = 
\begin{cases}
    \phi_i^H + \Delta_i & \bar{c}(i, H) > \text{th}_H \\
    \phi_i^H - \Delta_i & \bar{c}(i, H) < \text{th}_L \\
    \phi_i^H & \text{otherwise or q(i, H) empty}
\end{cases} \, ,
\end{equation}
for $\Delta_i$ a fixed, positive step size. 

\section{Training and evaluation in detail}\label{app:training}

\subsection{Training without ADR}

Of the 4 phases of behaviours shown in Figure~\ref{fig:adr-none}, phases 2 and 3 demonstrate social learning abilities. Moreover, the follow and remember strategy learned in phase 3 is more rewarding than the pure following strategy learned in phase 2. We hypothesise that phases 1 and 2 are necessary for the training of agents that follow and remember. To avoid phase 4, we must control the training task complexity appropriately so that social learning is always better than independence. In other words, we are required to maintain an information asymmetry between the expert and our agent, so that social learning remains rewarding. We refer to the just-right task complexity as a ``Goldilocks zone'' from which cultural transmission strategies can emerge. We verify this hypothesis in the experiments in sections \ref{sec:adr-single-param} and \ref{sec:full-adr}. One might question why we deliberately avoid the interesting behaviour of phase 4 henceforth. The reason is quite simply expediency: in this work, we are particularly focused on generating the strong and useful prior provided by robust and generalisable cultural transmission. Phase 4 clearly points to a rich seam of future work by fine-tuning this prior. 

\subsection{Training with single-parameter ADR}

As the world size randomisation range expands, the evaluation cultural transmission metric increases correspondingly. This is expected since the held-out evaluation scenarios only contain worlds of size $32 \times 32$. Thus, ADR creates a curriculum that allows cultural transmission to emerge in simple $16 \times 16$ worlds, then gradually expand to larger worlds including $32 \times 32$. Note that even after exceeding the evaluation world size, the evaluation cultural transmission metric remains lower than training cultural transmission metric. This is likely due to other modifications in the evaluation scenarios from the training setup. For example, the expert bot is replaced by recorded human demonstrations. We highlight the advantage of using the training cultural transmission metric to adapt boundaries in ADR rather than the agent score used in~\cite{openai2019}. As a threshold-based method, ADR is sensitive to the scale and normalisation of the parameter used to adapt boundaries. As Figure~\ref{fig:adr-world-size} shows, the CT metric is insensitive to changes in the maximum achievable score as the set of training tasks changes.

\subsection{Training with multi-parameter ADR}

The process of emergent social learning with increasing task difficulty is gradual and the experiment run spanned more than 300 hours (1.5 weeks). To provide a sense of the experience consumed, the average rate of inference steps for the agent was approximately $4 \times 10^6$ per hour, and inference data was reused on average $55$ times for learning. We note that all parameters controlled by ADR expanded significantly over the experiment run. 

\begin{table}[h]
    \centering
    \begin{tabular}{l|cccc}
    Parameter & Min & Initial & Max & Step \\ \hline
    World size & 20 & 20 & 32 & 1 \\
    Horizontal obstacle density & 0.0001 & 0.0001 & 0.01 & 0.0001 \\
    Vertical obstacle density & 0.0 & 0.0 & 0.2 & 0.0005 \\
    Bumpy terrain max value & 10.0 & 10.0 & 15.0 & 0.1 \\
    Bumpy terrain frequency & 0.01 & 0.01 & 0.1 & 0.002 \\
    Bot speed & 7.0 & 11.0 & 14.0 & 0.1 \\
    Probabilistic dropout transition probability & 2/1800 & 20/1800 & 40/1800 & 2/1800 \\
    \end{tabular}
    \caption{Task parameters controlled by ADR.}
    \label{tab:adr-parameters}
\end{table}

\begin{figure}[h]
        \centering
        \includegraphics[width=\textwidth]{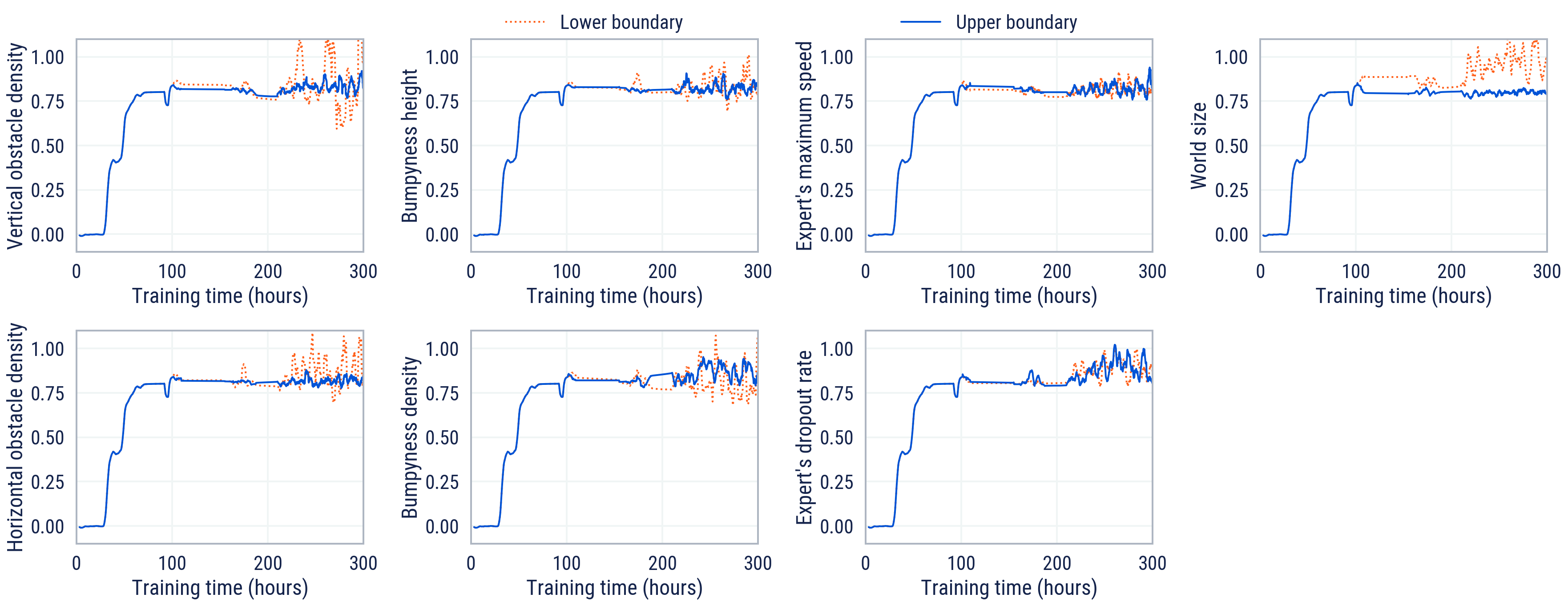}
        \caption{Training CT metrics. ADR adapts each randomisation range boundary by expanding when the metric is above an update threshold high of $0.85$ and contracting when the metric is below an update threshold low of $0.75$.}
        \label{fig:adr-full-training-local-ct}
    \end{figure}

\begin{figure}[h]
    \centering
    \includegraphics[width=0.7\textwidth]{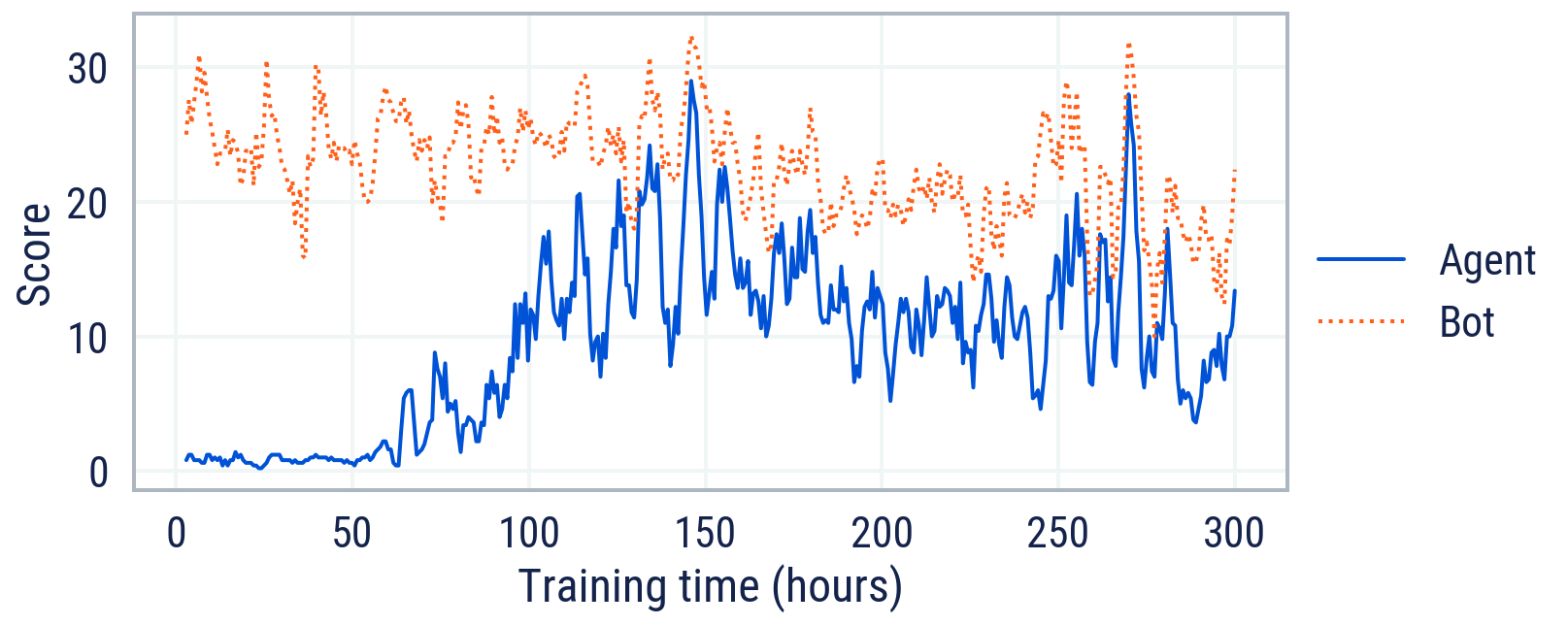}
    \caption{Agent and bot scores over ADR training. Smoothed using a simple moving average of length $5$.}
    \label{fig:adr-full-training-score}
\end{figure}

Figure~\ref{fig:adr-full-training-score} shows the agent and expert bot scores over the training run. As before, when ADR expanded the parameter boundaries, the maximum achievable score (approximately the bot score) decreased. The agent score also showed a corresponding decrease. The successful expansion of ADR reinforces our correct choice of using the cultural transmission metric to adapt ADR, in order to avoid scale and normalisation issues with score. Finally, we observe the ``spikes'' in both bot and agent scores over the training run, likely due to expansions and contractions of world parameter randomisation ranges by ADR.

\subsection{Evaluation via ablations}\label{app:ablation}

In our ablations, we take a top-down approach: starting from an agent of reference, we remove, in turn, only one single component at a time and train the ablated agent in the same manner as the reference. Across our ablations, we use two reference agents: 

\begin{enumerate}
    \item \textbf{MEDAL-ADR} is our highest performing agent, using full ADR, as described in Section \ref{sec:full-adr}. We use this agent as a reference only for ablating ADR.
    \item \textbf{MEDAL} is our best performing non-ADR agent, trained on a fixed $20\times 20$ world size with no vertical obstacles and few horizontal obstacles. The usual randomised procedural generation of the environment is, however, preserved. We use MEDAL as our reference agent when ablating all other components, except ADR. The choice of MEDAL over MEDAL-ADR provides a higher chance for the ablated variants to show their full potential in a simpler environment. Note that using MEDAL as a reference only permits sound evaluations in simple probe tasks because it never learns to navigate complex worlds without ADR.
\end{enumerate}

\textbf{Expert, Memory, and Attention} are crucial ingredients. Figure \ref{fig:ablate-0-agents} shows that the agent cannot solve the task without these (achieves $0$ score) and therefore also doesn't pick up any social cues from the expert (if present), accounting for a mean $0$ CT metric.

The presence of another agent (in our case, an expert) is necessary given the exploration difficulty of our task. By removing expert demonstrations -- and, consequently, all dependent components, the dropout (D) and attention loss (AL) -- the agent must learn to determine the correct goal ordering (also called solo training). Under the default exploration strategy built into MPO, the solitary agent is too risk-averse to solve the task. We noticed in \href{https://sites.google.com/view/dm-cgi\#h.xts8fmos08mg}{video recordings} that these solitary agents learn to avoid goal spheres altogether.

Without memory, our agent does not form connections to the previously seen cues, be they social, behavioural, or environmental. When replacing the LSTM with an equally sized MLP (keeping the same activation functions and biases, but removing any recurrent connections), our agent's ability to register and remember a solution is drastically reduced. 

Lastly, having an expert at hand is futile if the agent cannot recognise and pay attention to it. When we turn off the attention loss, the resulting agent, MED––, treats other agents as noisy background information, performing the task as if it were alone. Our agent needs to use social cues to bootstrap its knowledge about the task and the attention loss is an essential component that encourages it to do so.

\textbf{Dropout} is necessary to equip the agent with the ability to internalise new behaviours, rather than simply copy them live. Together with the attention loss, it strikes a balance between dependence on others and independent behaviour. Without expert dropout, the ablated agent, ME–AL, achieves more cumulative score early on in training because it can rely on the expert at all times. That comes, however, at the expense of lower cultural transmission. This is because the agent becomes dependent on the presence of the expert and cannot solve the task without it. However, our full MEDAL agent learns to be robust to different observations, recognises when it can bootstrap another agent's knowledge and when it needs to act on its own, which is a clear advantage.

\textbf{ADR} ensures adaptability to different levels of task difficulty, environmental complexity, and expert behaviour. In this ablation, we alter the training setup in two ways: 
(1) We set up the task to match the end-points of ADR, but remove any automatic adaptation. The agent needs to cope with the highest level of task complexity from the very beginning. We refer to this as the MEDAL-––– agent; 
(2) We sample uniformly the same values as in MEDAL-ADR; therefore, the agent sees an amalgam of settings that do not follow a curriculum (referred to as the MEDAL-–DR agent).
MEDAL-ADR achieves the highest cultural transmission in all probes, as shown in Figure \ref{fig:ablate-adr}. MEDAL-–DR, lacks robustness, achieving only half the CT of MEDAL-ADR. MEDAL-––– on the other hand, struggles to train at all: the endpoint for the ADR curriculum is too challenging as a starting point for RL. These results emphasise the importance of maintaining the task in the \textit{Goldilocks zone} for cultural transmission (with robust following and remembering) to be successful.

\section{Analysis in detail}\label{app:analysis}

Because the ADR parameter ranges can fluctuate over the duration of an ADR experiment, we use the following criteria to select the agent for analysis. First, we normalise the ADR parameter ranges at every training time step by the maximum range allowed according to Table~\ref{tab:adr-parameters}. The ranges are further averaged to obtain the mean ADR range. We also average the training cultural transmission metric of all ADR parameter boundaries at every training time step. Finally, we select a training time step that satisfies the following criteria:
\begin{itemize}
    \item Mean normalised ADR range $\geq$ 0.55
    \item Mean training cultural transmission metric $\geq$ 0.80
    \item Min normalised ADR range $\geq$ 0.10
\end{itemize}

\subsection{Generalisation}

\begin{table}[h]
    \centering
    \begin{tabular}{l|ccc}
    Parameter & In-distribution range & Out-of-distribution ranges \\ \hline
    World size & [20, 26] & [18, 19], [26, 31] \\
    Horizontal obstacle density & [0.0001, 0.0024] & 0.0, [0.0025, 0.003] \\
    Vertical obstacle density & [0.0, 0.0255] & [0.026, 0.031] \\
    Bumpy terrain max value & [0.0, 3.4] & [3.5, 4.1] \\
    Bumpy terrain frequency & [0.01, 0.1] & [0.0, 0.008], [0.102, 0.12] \\
    Bot speed & [9.0, 13.7] & [7.2, 8.6], [13.8, 16.4]\\
    Probabilistic dropout transition probability & [2/1800, 40/1800] & 0.0, [42/1800, 48/1800]  \\
    \end{tabular}
    \caption{In-distribution and out-of-distribution parameter values for the evaluated MEDAL-ADR agent.}
    \label{tab:adr-parameters-id-ood}
\end{table}

We define \textit{obstacle complexity} as the tuple (horizontal obstacle density, vertical obstacle density). Each component is normalised by its minimum and maximum values listed in Table~\ref{tab:adr-parameters-id-ood} to yield a value between 0 and 1. For example, an obstacle complexity of 0 corresponds to an empty world, while a value of 1 corresponds to a world with a density of 0.003 and 0.031 of the horizontal and vertical obstacles respectively. Similarly, we define \textit{terrain complexity} as the tuple (world size, bumpy terrain amplitude, bumpy terrain frequency), all normalised. 

Results from the evaluation over this space, alongside example worlds from its extremal values, are shown in Figure~\ref{fig:generalization-analysis-world-space}. In addition, results for each parameter are given in Figure~\ref{fig:generalization-analysis-world-space-individual} to disentangle MEDAL-ADR's generalisation capabilities across the different components. In the per-parameter plots, all other parameters are held fixed at the mean of their in-distribution value.

\subsection{Fidelity}

We plot trajectories of the MEDAL-ADR agent and expert bots to further investigate the correlation between their behaviours within a single episode (Figure~\ref{fig:trajectory_plots}). In particular, we first plot the trajectory of the agent and bot as the bot performs the task optimally, and then plot the trajectory after the bot either (1) drops out of the environment or (2) begins acting sub-optimally halfway through the episode. We additionally plot the agent's trajectory when the bot provides an incorrect demonstration and then drops out. Finally, we measure the agent's performance without ever placing the bot in the environment to qualitatively investigate whether the agent is capable of solving the task on its own. Together, these scenarios provide a visual indication of the fidelity with which an agent can imitate an expert's demonstration, and the fidelity with which an agent can reproduce that demonstration in the absence of the expert.

\subsection{Neural activations}\label{app:bonus-neural-activations}

The linear probing method for identifying a social neuron is as follows.

\begin{figure}[htb]
    \begin{subfigure}[b]{0.48\textwidth}
        \centering
        \includegraphics[width=\textwidth]{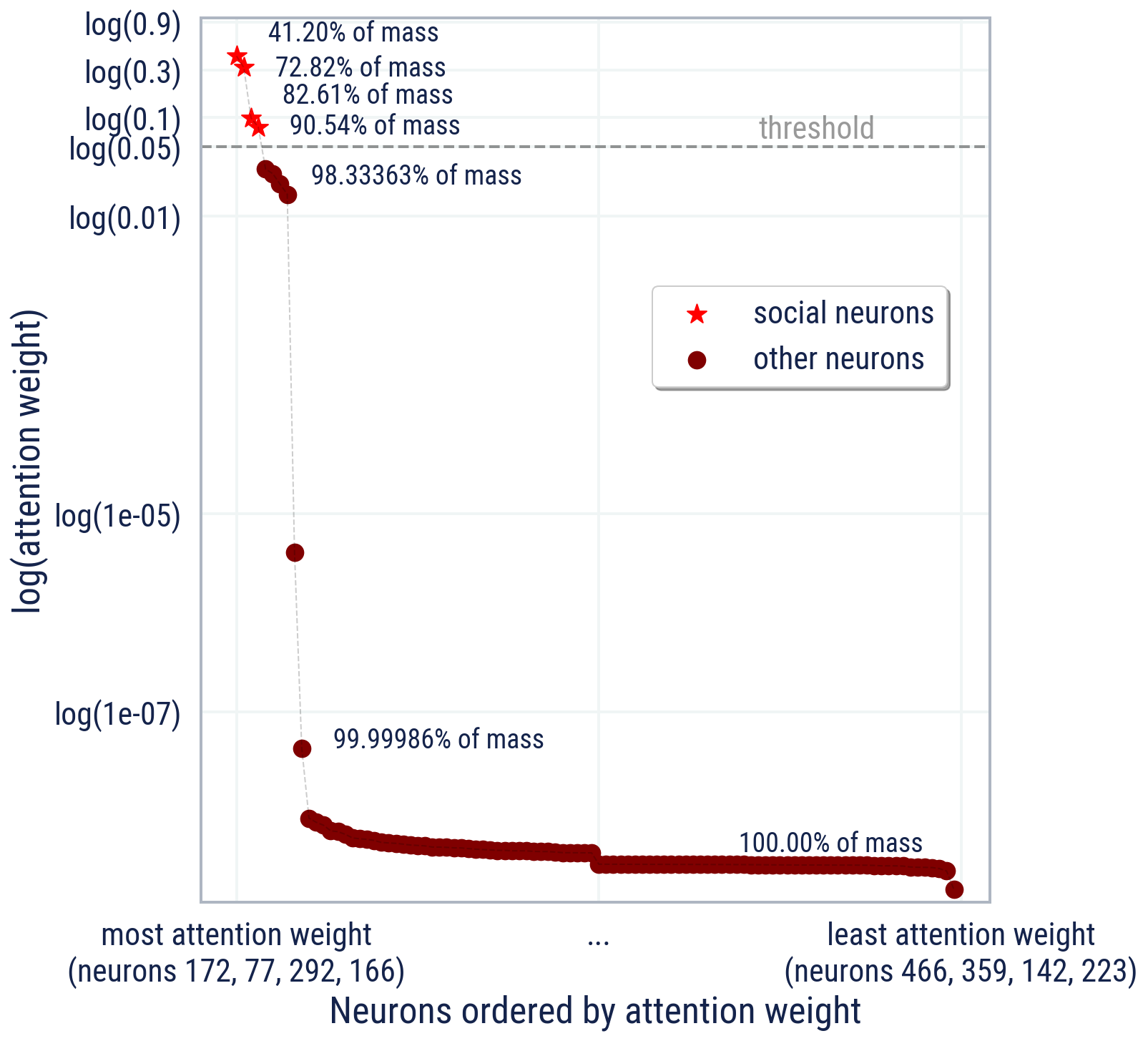}
        \caption{\centering MEDAL}
        \label{subfig:social-neuron-probability-mass-medal}
    \end{subfigure}
    \hfill
    \begin{subfigure}[b]{0.48\textwidth}
        \centering
        \includegraphics[width=\textwidth]{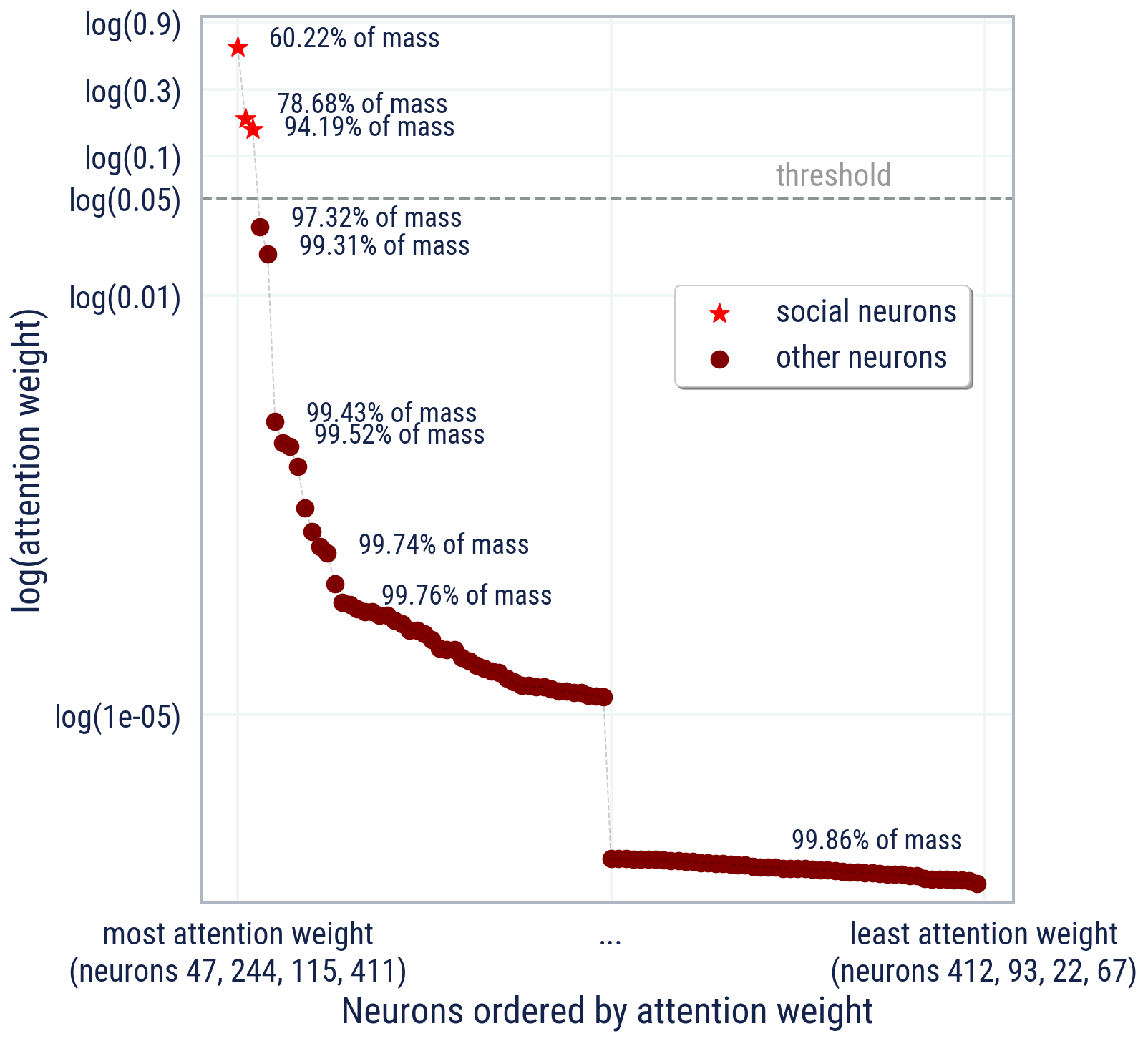}
        \caption{\centering MEDAL-ADR}
        \label{subfig:social-neuron-probability-mass-medal-adr}
    \end{subfigure}
    \caption{Ordering the neurons based on their corresponding (log) attention weight, we notice a considerable gap between the social neurons and the rest of the neurons. The identified social neurons account for more than 90\% of the total probability mass. This result is consistent across both MEDAL (\subref{subfig:social-neuron-probability-mass-medal}) and MEDAL-ADR (\subref{subfig:social-neuron-probability-mass-medal-adr}), and motivate the choice of our deliberately loose threshold (0.05). The other agent used for comparison, MED––, does not have any social neurons because none of the neurons surpasses the threshold.}
    \label{fig:social-neuron-probability-mass}
\end{figure}

\begin{enumerate}
    \item \textbf{Data collection\quad} We load a trained agent and run its policy (without further training) for 200 episodes, randomising the goal and the timestep at which dropout occurs. We collect pairs of features, namely the agent's belief state and a corresponding binary label indicating whether or not the expert is present for every timestep in each episode.
    \item \textbf{Training\quad} We randomise and split the previously collected dataset into a train (70\%) and test set (30\%). We train a new classification model that predicts whether the expert was present or not at each timestep based on the agent's associated belief state (already trained and frozen). The model is parameterised by an attention map of the same size as the belief (i.e., $512$), learning the relative importance of each of its neurons. The attention-weighted belief then passes through a linear layer and is projected onto a 2D space, corresponding to the task's classes. Under this training strategy, we identify social neurons as the maximally attended neurons, those for which the attention weight is higher than a threshold, which we set to $0.05$. Figure \ref{fig:social-neuron-probability-mass} proves that we picked this threshold sensibly: the social neurons selected under this criterion account for more than $90\%$ of the probability mass.
    \item \textbf{Evaluation\quad} We first check the accuracy of the newly trained classification model on the test set. To causally probe our earlier assumptions, we also make two interventions on the test set: one replaces the identified social neurons with random activations; the other replaces all neurons except the identified social neurons with random activations. Note that we have drawn these random values from a normal distribution resembling that neuron's original activation distribution in the training set. We also report the accuracy on the test set when setting all neurons to random activations drawn from their distributions (random baseline).
    \item \textbf{Analysis\quad} Finally, we look at the trained attention map and activations of the identified social neurons across episodes. 
\end{enumerate}

\begin{figure}[htb]
    \centering
    \begin{subfigure}[b]{0.46\textwidth}
        \centering
        \includegraphics[width=\textwidth]{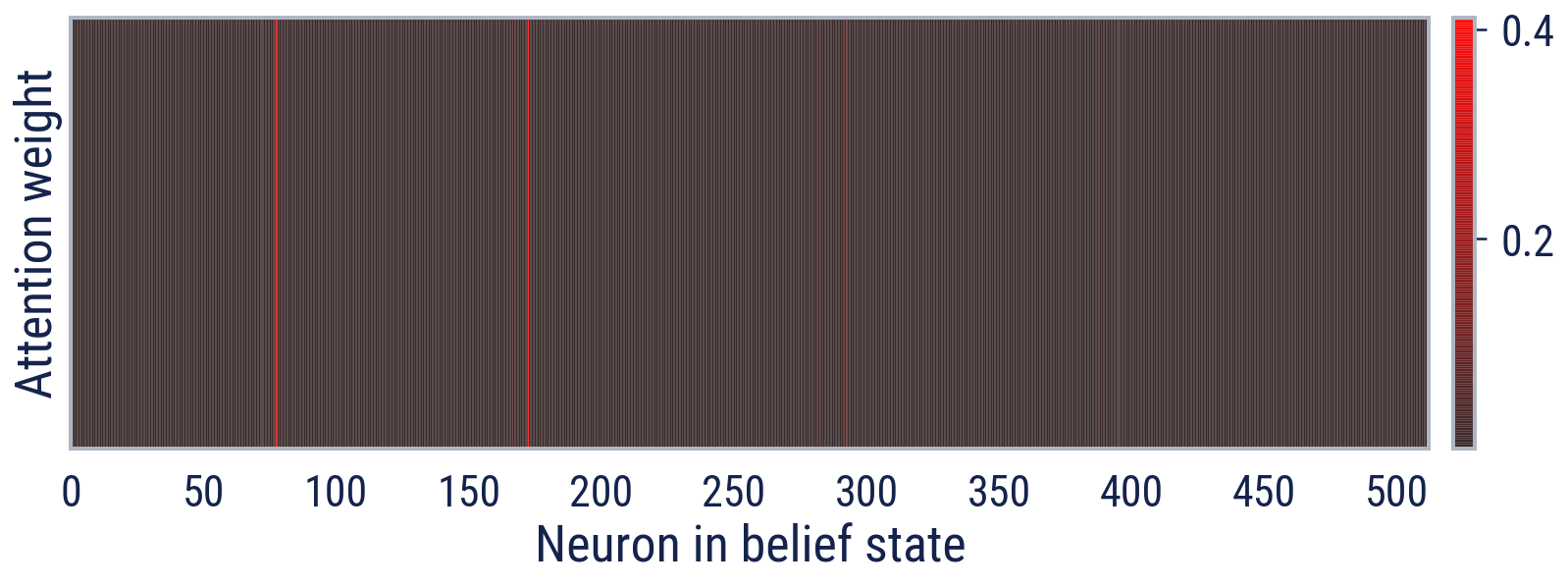}
        \caption{\centering MEDAL's learned attention map}
        \label{subfig:medal-attention-map}
    \end{subfigure}
    \hfill
    \begin{subfigure}[b]{0.47\textwidth}
        \centering
        \includegraphics[width=\textwidth]{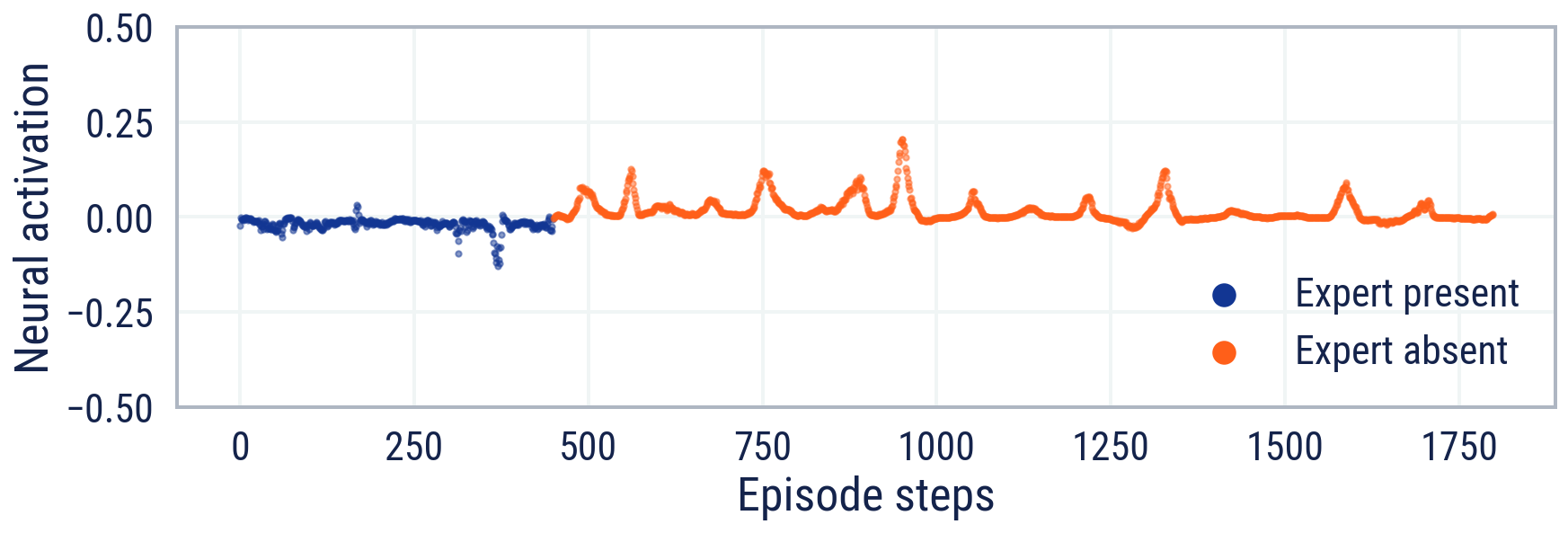}
        \caption{\centering One of MEDAL's social neuron activations}
        \label{subfig:medal-activations}
    \end{subfigure}

    \begin{subfigure}[b]{0.46\textwidth}
        \centering
        \includegraphics[width=\textwidth]{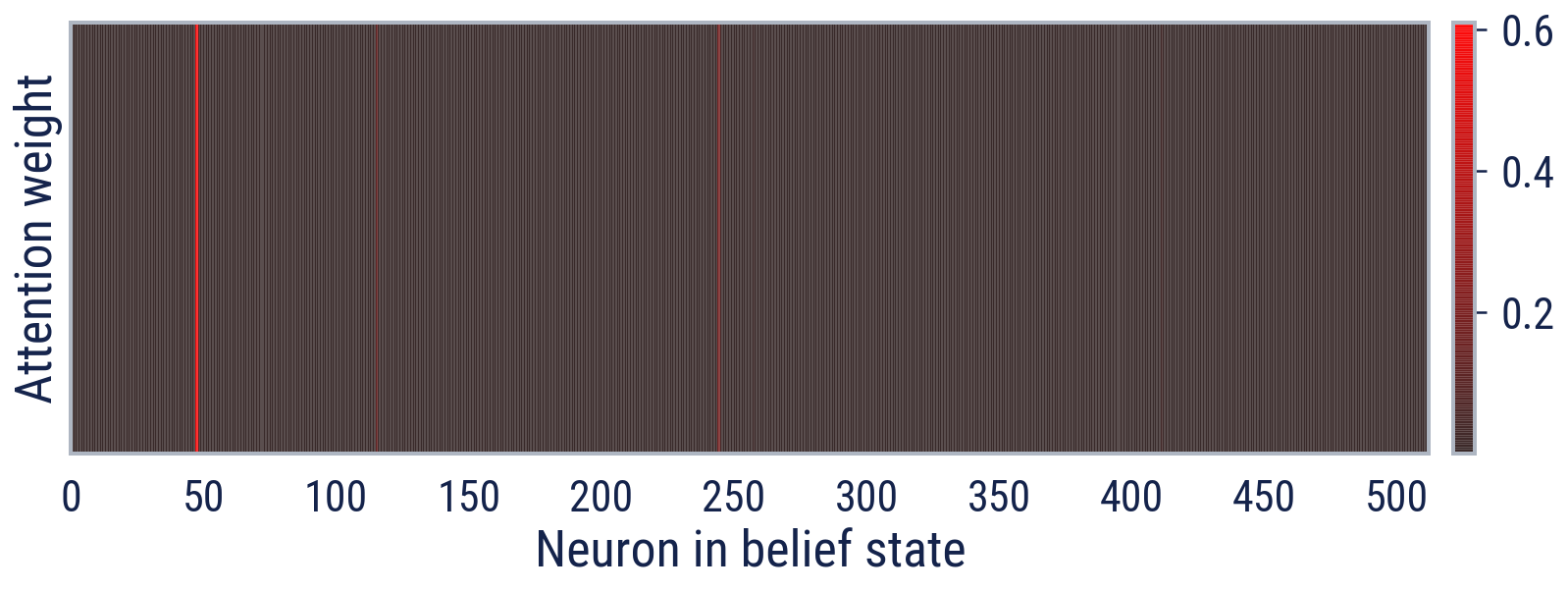}
        \caption{\centering MEDAL-ADR's learned attention map}
        \label{subfig:medal-adr-attention-map}
    \end{subfigure}
    \hfill
    \begin{subfigure}[b]{0.47\textwidth}
        \centering
        \includegraphics[width=\textwidth]{figures/activations/activations-social-neuron-47-medal-adr.png}
        \caption{\centering MEDAL-ADR's social neuron activations}
        \label{subfig:medal-adr-activations}
    \end{subfigure}

    \caption{Having identified social neurons as those maximally weighted in the attention map (\subref{subfig:medal-attention-map}, \subref{subfig:medal-adr-attention-map}), we observe a sharp sign or magnitude change in the activations of the social neurons when the expert drops out. As an example, we plot these across one randomly selected episode for MEDAL (\subref{subfig:medal-activations}) and MEDAL-ADR (\subref{subfig:medal-adr-activations}).}
    \label{fig:social-neuron-activations}
\end{figure}

We perform the steps above for three of our ablated agents: MED––, MEDAL, and MEDAL-ADR. Comparing their accuracies on the causal and control probes (see Figure \ref{fig:social-neuron-accuracy}), it is apparent that, on the one hand, the classifier trained on MED––'s representations achieves low accuracy, with no neurons surpassing the attention weight threshold; on the other hand, the classifiers trained on MEDAL and MEDAL-ADR's representations have high accuracies, and their identified social neurons play a role in maintaining and reliably predicting the expert's presence on their own. 

We make some empirical remarks about the nature and behaviour of the goal neuron found in our MEDAL-ADR agent. It suddenly fires when it enters a goal area and remains high as long as the agent remains inside. Interestingly, the neuron signals its anticipation of exiting the potentially rewarding zones with a slight decrease in the magnitude of its activation, followed by a sharp drop and levelling around zero. To see these, compare the activations of the goal neuron when the agent completes the task, aided by a bot (Figure \ref{subfig:cyclic-neuron-full}), and when the agent fails to do so because it is alone (Figure \ref{subfig:cyclic-neuron-no-bot}).

Note that it is not the presence or the following of an expert that determines the spikes. Figure \ref{subfig:cyclic-neuron-half}, probing the activations of the goal neuron in a dropout scenario, demonstrates this. After performing the task with the bot, the agent, when left alone, also tries to solve it independently. Despite the absence of an expert, its goal neuron spikes every time it enters an expected rewarding zone. Note that it is not the observation of a positive reward that causes the neuron's spiking. Figures \ref{subfig:cyclic-neuron-no-bot} and \ref{subfig:cyclic-neuron-noisy} (around steps 200 and 950, respectively) show that the agent can enter the same goal twice and spike each time, although it receives a reward only the first time.  

Not fulfilling the agent's expectation of a rewarding goal reduces the magnitude of the goal neuron's spike. That is manifested either by accumulating a negative point (Figure \ref{subfig:cyclic-neuron-half}) or by entering the same goal twice (Figure \ref{subfig:cyclic-neuron-no-bot}) in the absence of an expert. However, the expert's presence cancels out the effect of negative rewards on spikes, as shown in Figure \ref{subfig:cyclic-neuron-noisy} featuring a noisy scenario. We hypothesise that the identified goal neuron spikes in anticipation of reward and thus correlates with decisiveness and intention. It carries some resemblance to the analysis of persistent decision variables introduced by \cite{BARI2019stable}, where changes in firing rates occur around the moment of choices and outcomes.

\begin{figure}[htb]
    \centering
    \includegraphics[width=0.67\textwidth]{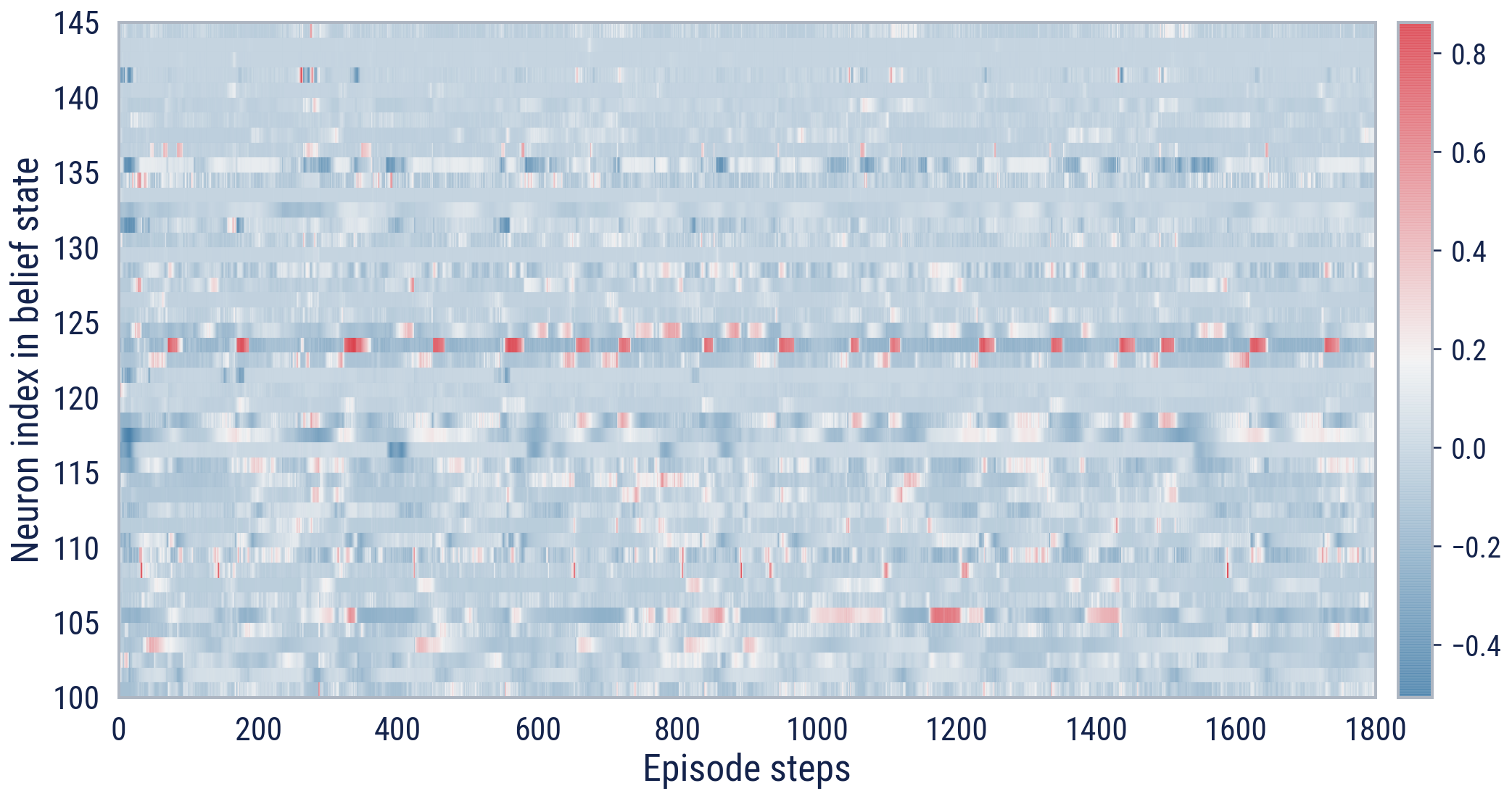}
    \caption{We provide a snapshot of the activations of 45 neurons in MEDAL-ADR's belief state, where we identify neuron $123$ as a ``goal neuron''. At first glance, and using the neighbouring neurons for contrast, the goal neuron is characterised by a distinctive periodic variation in its firing pattern across the episode.}
\label{fig:cyclic-neuron-activations-snapshot}
\end{figure}

\end{document}